\documentclass[a4paper,12pt,openany,oneside]{book}
\pdfoutput=1
\usepackage[inner=40mm,outer=20mm,top=30mm,bottom=20mm]{geometry}
\usepackage{setspace}
\usepackage{epsfig}
\usepackage{amssymb}
\usepackage{amsmath}
\usepackage{multirow}
\usepackage{pdfpages}
\usepackage[printonlyused,withpage]{acronym}
\usepackage{fancyhdr}
\usepackage{subfig}
\usepackage{comment}
\usepackage{xcolor}%
\usepackage{enumitem}
\usepackage{amsthm}
\usepackage{siunitx}
\usepackage{url}
\usepackage{etoolbox}
\newtheorem{theorem}{Theorem}[section]
\newtheorem{remark}{Remark}[section]
\newtheorem{assumption}{Assumption}[section]
\newtheorem*{prove}{\bf Proof}
\newtheorem{definition}{Definition}[section]
\AtEndEnvironment{prove}{\null\hfill\qedsymbol}

\begin{document}
\frontmatter

\begin{titlepage}
\centering
{	\bf \Large Decentralized Collision-Free Control of Multiple Robots in 2D and 3D Spaces}
\\
\vspace{4cm}
{by}
\\
\vspace{0.3cm}
{\bf \Large Xiaotian Yang}
\\
\vspace{14cm}
{2017}
\end{titlepage}
\chapter*{Abstract}
\addcontentsline{toc}{chapter}{Abstract}
Decentralized control of robots has attracted huge research interests. However, some of the research used unrealistic assumptions without collision avoidance. This report focuses on the collision-free control for multiple robots in both complete coverage and search tasks in 2D and 3D areas which are arbitrary unknown. All algorithms are decentralized as robots have limited abilities and they are mathematically proved.

The report starts with the grid selection in the two tasks. Grid patterns simplify the representation of the area and robots only need to move straightly between neighbor vertices. For the 100\% complete 2D coverage, the equilateral triangular grid is proposed. For the complete coverage ignoring the boundary effect, the grid with the fewest vertices is calculated in every situation for both 2D and 3D areas. 

The second part is for the complete coverage in 2D and 3D areas. A decentralized collision-free algorithm with the above selected grid is presented driving robots to sections which are furthest from the reference point. The area can be static or expanding, and the algorithm is simulated in MATLAB. 

Thirdly, three grid-based decentralized random algorithms with collision avoidance are provided to search targets in 2D or 3D areas. The number of targets can be known or unknown. In the first algorithm, robots choose vacant neighbors randomly with priorities on unvisited ones while the second one adds the repulsive force to disperse robots if they are close. In the third algorithm, if surrounded by visited vertices, the robot will use the breadth-first search algorithm to go to one of the nearest unvisited vertices via the grid. The second search algorithm is verified on Pioneer 3-DX robots. The general way to generate the formula to estimate the search time is demonstrated. Algorithms are compared with five other algorithms in MATLAB to show their effectiveness.

\tableofcontents

\listoffigures
\addcontentsline{toc}{chapter}{\listfigurename}

\listoftables
\addcontentsline{toc}{chapter}{\listtablename}

\mainmatter
\pagestyle{fancy}
\renewcommand{\chaptermark}[1]{\markboth{#1}{}}
\renewcommand{\sectionmark}[1]{\markright{\thesection\ #1}}
\fancyhead{}
\fancyhead[LO]{\bfseries\rightmark}
\fancyhead[RE]{\bfseries\leftmark}
\setlength{\headheight}{15pt} 

\setcounter{secnumdepth}{3}

\chapter{Introduction}
\chaptermark{Introduction}%
The recent improvements of technologies and algorithms in robotics enable robots to be widely used in various applications. The civilian usages include the vacuum cleaning at home \cite{RN173,RN174,RN179,RN180}, manufacturing and material handling in factories \cite{RN175,RN176}, fire detection in the forest \cite{RN86} and warehouse \cite{RN263}, pollution estimation in the sea \cite{RN264,RN273} and the air \cite{RN274},  and data harvest \cite{RN131}. There are also some military applications such as the border patrol \cite{RN98,RN127,RN50}, the deployment of robots for intruder detection \cite{RN169,RN118,RN171,RN172,RN181,RN184,RN182,RN262}, intelligence, surveillance and reconnaissance missions \cite{RN130,RN148}, mine clearance \cite{RN49,RN177,RN249,RN251,RN253,RN255,RN256,RN124}, boundary monitoring \cite{RN270}, and the terrorist threat detection in ports \cite{RN272}. Most of the reference above is about search, coverage and path planning tasks. In the employed robots, the autonomous mobile robots, including \acp{UUV}, \acp{UGV}, and \acp{UAV}, have attracted the most research interests as they can be programmed to execute certain tasks without further control from the human. Sometimes, a team of multiple robots is used in a task as it has certain advantages compared to a single robot. When a team of autonomous robots is employed, the ways that robots coordinate to perform cooperative tasks become the main research topic. The algorithms can be divided into decentralized methods and centralized methods. Another issue raised by a team of robots is collision avoidance. However, many algorithms ignored this problem or use unrealistic assumptions thus cannot be applied to real robots. Thus, to ensure a safe navigation through the task, standard robot models or actual robots need to be considered as the base of the assumptions. The task area can be \ac{2D} or \ac{3D}, so the suitable method to represent the 2D or 3D environments, as well as obstacles inside should be carefully considered.

This report studied the problems of complete coverage and search using multiple robots. The task area is arbitrary unknown, and robots move without collisions based on a grid pattern. The best grids under different situations are discussed. One decentralized algorithm for the complete coverage and three decentralized random algorithms for search tasks are provided. The algorithms are initially designed in 2D area and can be used in 3D area with easy modification. The convergence of each algorithm is proved, and all algorithms are verified by extensive simulations and comparisons with other methods. Algorithms are designed based on datasheets of real robots, and one search algorithm is applied to Pioneer 3-DX robots to show the effectiveness.

Sections below are the literature review on the related topics in this report.   
\section{Multiple Robots}
When finishing some tasks using a single robot, the robot usually needs to have strong abilities with advanced sensors and complex algorithm. One kind of popular ground robots in research is the quadruped robot such as BigDog designed by Boston Dynamics \cite{RN187} and Mountainous Bionic Quadruped Robot from China North Industries Corporation \cite{RN186}. These robots are designed for rough-terrain with different obstacles which may be too steep, muddy and snowy to use common wheeled robots. They are also stable with heavy loads and external interference such as pushes from different directions. Correspondingly, those robots need various sensors for the environments, complicated computing and power systems, advanced actuators and sophisticated dynamic controllers. For UAVs, the Parrot AR. Drone has been widely used in research and education \cite{RN191}. In \cite{RN190}, a trajectory tracking and 3D positioning controller was designed for this drone using its onboard video camera to building the map of the task area. A single robot path planning algorithm for complete coverage using this robot is developed in \cite{RN192}. A famous UUV used in search of Malaysia Airline MF370 is the Bluefin-21 robot \cite{RN124}. Some other robots in research can be found in the survey paper \cite{RN194}.

However, there are many tasks which cannot be done by a single robot. For the time-sensitive works, such as search and rescue after a disaster or the detection of dangerous leaking biological, chemical, and radioactive materials, a single robot may be unable to finish the task on time. For the coverage or data collection task in a vast area, only one robot would not be able to have such a large sensing range, and a larger sensing range means more expensive equipment should be used. Sometimes, a task is designed for multiple members such as the soccer game and the targets tracking mission with multiple aims moving in different directions where one robot is not enough to follow each of them. Thus, the cooperation of multiple robots is required. The recent developments in Micro-electro-mechanical systems, integrated circuits, wireless communication, as well as the swift progress in algorithm innovation also attracts researchers to focus on the control problem of the robots network. As cheaper robots \cite{RN257,RN258,RN201,RN180} with simple functions are available \cite{RN249} to finish a task by cooperation, an expensive robot with advanced capability may be unnecessary.

A system with multiple robots sometimes may be called as a \ac{WMSN} in some papers such as \cite{RN138,RN84,RN10,RN6,RN94,RN44,RN145,RN2,RN143,RN83}. In this report, both of them refer to a group of autonomous mobile robots which carry sensors to detect targets or cover an area. The energy of the robots comes from its battery. Robots can also build the local network to communicate with neighbors to share information. Based on those detected or received information, a robot moves dynamically to finish a task based on the control algorithm. As a considerable number of robots can be used, each of them should have a reasonable price, a carry-on battery, limited sensing ranges for obstacles and targets, and a limited communication range. 

The robots in a team do not always need to be homogeneous. They can be heterogeneous as in the algorithms and experiments in \cite{RN170,RN183,RN225,RN115}. Due to the complexity of the tasks, especially search and rescue, and the limited abilities of each robot, members of a multi-robot system can have different functions. For example, the HELIOS team with five robots were proposed by Guarnieri et al. for the search and rescue task \cite{RN216}. In the system, only three robots have laser range finders and cameras to create the virtual 3D map of the task area. The other two robots with manipulators are used to execute the special tasks including handling objects and opening doors. Sugiyama et al. \cite{RN218,RN215} developed the chain network in a rescue system with multiple robots. The system consists of a base station, some search robots and some relay robots. Different robots have their behavior algorithms to form the chain network. The method of the classification and the behavior algorithm is based on the forwarding table of each robot constructed for ad-hoc networking. Then the robots recognized as relay robots will act as information bridges, and frontier robots will search the area. Luo et al. \cite{RN217} exploited a team with both ground and aerial vehicles for search and rescue tasks. The environment mapping was executed by the ground vehicle. The search and localization were finished simultaneously by the micro aerial vehicle with a vertical camera and a horizontal camera. Another two micro ground vehicles equipped with color sensors, sonar, and compasses were used as the backup team. \cite{RN225} proposed a hierarchical system consisting of managers with advantages in computation and worker with limited ability. This system allows the processing resources to be shared and global environment to be divided into local sections for the workers. Worker robots can be separated into planners, explorers or robots with both jobs based on their ability to only deal with their assigned tasks. Algorithms in this report only require the robots to satisfy certain assumptions about their sensing ranges, the communication range, and the maximum swing radius but the robots can be different. In search tasks, all robots are doing the same task without further actions on the found targets so there is no need to separate robots to distinct roles.

Multiple robots have many advantages comparing to a single robot \cite{RN195,RN196,RN200,RN201,RN199,RN1,RN101}. Firstly, a system with multiple robots is more efficient as discussed above because they can share information with neighbors so that they can cooperate to know the environment faster and better. Multiple cooperative robots can be separated spatially and temporally. Therefore, they can fulfill the complex, decomposable task efficiently such as cleaning floor and surveillance. Correspondingly, it may save the total consumed energy. Secondly, it is reliable as more robots provide more redundancy than a single robot. If one robot is not working, other robots could finish the task if they use an independent algorithm. Through communication, the robot failure may even be detected. Thirdly, the system is robust to errors because the information of one robot can be judged or rectified using information from other robots. Also, the system can be flexible to different tasks with different areas maybe by reassigning the task for each robot. Finally, a robot team can be scalable with various numbers of members to satisfy the task requirement, but the expense of robots need to be considered. 

While using a group of robots, many algorithms for a single robot may not be available, and it is challenging to design the algorithm in different aspects. The later sections will discuss them one by one. 
\section{Centralized vs. Decentralized Algorithms}
In the robot teamwork, communication and control methods are significant and complex topics. The algorithms for multiple robots can be divided into centralized algorithms and decentralized algorithms. In centralized control tasks such as \cite{RN212,RN226,RN227}, the communication center or the leader robot collects and distributes the information to all the others, which is relatively easy to control and to design algorithms. Thus, the leader robot has a larger memory and stronger communication ability which will be expensive. However, on this occasion, global information such as the \ac{GPS} data may be needed for synchronization and localization. This method also requires all robots to have good communication ability. However, those demands may be unrealistic in some occasions such as in caves, under water and in areas without control such as drones in enemy’s airspace. In another type of centralized control method, the leader-follower algorithm \cite{RN213,RN204,RN214,RN125}, robots need to keep a very restrictive type of communication topology \cite{RN206,RN207}, namely, they need to stay in the communication range of each other. In the rescue network of \cite{RN218}, communication chains must be continuously existing from the base station through the relay path to the distant rescue robots to reconnoiter disaster areas, and those chains must be transformed to a suitable topology if the target area of exploration is changed. Some other papers assumed that the communication graph is minimally rigid \cite{RN208,RN209} or time-invariant and connected \cite{RN210}. Thus, they are used in formation control or achieving some consensus variables such as the same speed, the same direction, but they cannot work independently in a large area. A deadly disadvantage of the centralized scheme is that if the leader robot is broken, the whole system will stop working. Therefore, the centralized control and communication algorithms have limited applications and lead researcher to the decentralized methods. 

Decentralized control of multiple robots is an attractive and challenging area \cite{RN84,RN19,RN22,RN153,RN26,RN17}, however, it provides the system with benefits in robustness, flexibility, reliability, and scalability according to Zhu \cite{RN13}. Specifically speaking, this problem describes a group of dynamically decoupled robots, namely, each robot is independent and will not be affected by others directly according to Savkin \cite{RN317}. Research in this area is largely inspired by the animal aggregation problem which belongs to ecology and evolution branches of biology. It is believed by Shaw \cite{RN24}, Okubo \cite{RN28} and Flierl \cite{RN27} et al., that some simple local coordination and control laws were exploited to achieve the highly intelligent aggregation behavior by a group of bees, a school of fish and a swarm of bacteria. Those laws may be used in formation control. Described by \cite{RN102}, efficient foraging strategy of animal can be used in a coverage and search problem for animals. The evolutionary process through the natural selection led to highly efficient and optimal foraging strategies even with physical and biological constraints. It can be seen in the method that lobsters used to localize and track odor plumes or in the bacterial chemotaxis mechanism used by e-coli to response the nutrition concentration gradient. Similarly, robots only need a local communication range and small sensing ranges which are achievable for many cheap robots. Also, less communication is necessary compared to the centralized method in which gathering information from and sending tasks to each terminal need much time. Moreover, in some steps, using a decentralized algorithm, different members of the team can execute their missions in parallel to improve the efficiency of the work. Both centralized and decentralized communications are used in this report to improve the efficiency of the search or coverage task as well as to save the total energy. The centralized communication is only used to deliver the significant information such as the positions of targets or the positions of the failed robots, which is like the usage of the flare gun for delivering distress signal in the real rescue scenarios. Decentralized control and communication is the main part of the algorithms in this report to save time and money and thus, making the algorithms more applicable. Initially, the information of the environment is unknown. Then a robot will gradually record its sensed map and exchange the map with neighbor robots that it can communicate with. Thus, the robots only use local information to make the decision and will have different neighbors in later steps \cite{RN153,RN17,RN1,RN101,RN94,RN2,RN11,RN7}.

In literature from the perspective of control engineering, ideal communication situations were usually considered. However, research in the area of telecommunication discussed the non-ideal communication including the transmission delay due to the limited bandwidth, the propagation delay because of the long transmission distance or the information loss caused by inference especially in a battlefield where severe interference is generated by the enemy. \cite{RN65} considered the multi-agent system with communication time-delays. It used switching topology and undirected networks. Then a direct connection between the robustness margin to time-delays and the maximum eigenvalue of the network topology is provided. In \cite{RN125} with the leader-follower structure, the state of the considered leader not only kept changing but also might not be measured. Thus, a state estimation method for getting the leader's velocity is proposed. In systems with communication constraints such as limited capacity in communication channels, estimation of the necessary information are needed using methods of Kalman state estimation in \cite{RN128,RN129,RN219,RN318,RN221,RN222,RN341,RN342,RN343}. Other hardware limitations may include the insufficient memory for a large global map. \cite{RN224} solved this problem by describing the global map by local maps which was explored by a two-tiered A* algorithm. This algorithm can be executed entirely on the robots with limited memory. As the energy of the mobile robots is limited \cite{RN267}, \cite{RN10} researched the area covered during the cover procedure instead of the final result. It considered coverage rate at a specific time instant or during a time interval. It also showed how long a location is covered and uncovered. Then a relation between detection time for a randomly located target and parameters of the hardware were given to help a user to decide the number of robots to use. Ideal communication is assumed in this work and propagation delay would not happen as local communication is fast enough to ignore that delay. The problem caused by the limited bandwidth is also solved by designing a suitable mapping method to minimize the information needed for the area. The memory is considered as sufficient and the search time with different parameters is discussed. 

\section{Complete Coverage and Search}
Popular tasks for cooperation of multiple robots include coverage control, search and rescue, formation control \cite{RN11,RN90,RN62,RN19,RN61}, flocking \cite{RN26,RN68,RN67,RN64}, and consensus \cite{RN184,RN25,RN65}. This report focuses on the complete coverage tasks, and search and rescue tasks. Coverage problems were defined differently by various researchers \cite{RN32,RN34,RN36,RN40} from their research interests which are lack of generality. It could be concluded as going through a target area by utilizing the sensors on the robots and achieving the requirements as consuming the least time, going through the shortest path or having the least uncovered area. As defined by Gage \cite{RN43}, there are three types of coverage problem namely, barrier coverage \cite{RN226,RN227,RN6,RN83}, sweep coverage \cite{RN163,RN83,RN5} and blanket coverage \cite{RN2,RN94,RN155,RN171,RN172}. Another two coverage types are dynamic coverage \cite{RN10} and search. Barrier coverage means the robots are arranged as a barrier, and it will detect the invader when they pass through the barrier. Sweep coverage can be thought as a moving barrier to cover a particular area which maximizes the detections in a time slot as well as minimizes the missed detections per area. Blanket coverage is the most difficult one. It aims at deploying robots in the whole area so that each point of the area is covered by at least one robot at any time \cite{RN2} in which the selected deployment locations are designed to minimize the uncovered area and the deployment procedure should be as short as possible to reach the blanket coverage early. In the \acl{3D} area, the term $ blanket $ is not suitable for the same coverage problem and is called complete coverage or complete sensing coverage in general for both types of areas.

For complete coverage, there are two types of self-deployment strategies based on the survey \cite{RN139} which are the static deployment and the dynamic deployment. In the static deployment, robots move before the network start to operate. However, the deployment result such as the coverage rate or the connectivity of the network is assessed under the fixed network structure, or the assessment is irrelevant to the state of the network. For the dynamic one, the condition of the network may change such as the load balance of the nodes leading to the repositioning of the mobile robotic sensors \cite{RN245}. The complete coverage in this report firstly considered 100\% coverage between nodes for a fixed area as the static deployment. Then the dynamic self-deployment in an extending area, such as the demining area in the battlefield which could be enlarged, is considered by a simple modification of the static method. Some deployment methods deployed all robotic sensors manually \cite{RN165}, however, the cost is high if the area is large, so the applications are usually in an indoor area as in \cite{RN307,RN306,RN305}. Moreover, in some applications in a dangerous or unknown environment, such as demining, human actions are impossible. Thus, in some papers, mobile sensors were randomly scattered from helicopters, clustered bombs or grenade launchers with assumptions of the distribution pattern \cite{RN139} such as the uniform distribution in \cite{RN309}, the simple diffusion, and the R random in \cite{RN308}.

Dynamic coverage in \cite{RN10} is similar to a search task. They both describe the coverage problem with continuously moving sensors to find targets. Movements of sensors are required because, in a vast area, blanket coverage is not realistic as it needs a significant number of robots which is expensive. Therefore an algorithm should be designed which uses only a selected number of robots to cover the unknown area completely dynamically. Thus, a section of the area will be covered sometimes and be uncovered in other time slots. However, the dynamic coverage ensures that each point is visited at least once through the whole search procedure. Note that in dynamic coverage or search, mobile sensors do not need to move together in a formation like in the sweep coverage. Note that, a search task may also be named as an area exploration task \cite{RN160,RN215,RN32,RN12,RN225,RN183,RN86} in the mapping related topics or be called as a foraging strategy for animals \cite{RN248,RN247}.

In both complete coverage and search problems, some papers used the assumption about the boundary effect \cite{RN147,RN171,RN122,RN279}. The assumption claims that the length, width (and height if it is a 3D area) of the search area is much larger than the sensing range so that the gaps between the sensed area and the boundary are small enough to be ignored so targets would not appear in those gaps \cite{RN122}. However, some other papers ignored this assumption but still claimed they reached the complete coverage or a complete search \cite{RN134,RN118,RN119,RN138,RN229,RN234,RN230,RN101,RN232,RN233,RN1}. Those papers could only be correct by having assumptions about the shape of the boundary and passage width. For example, the boundary is simple and smooth to allow the equilateral triangle grid to stay inside the border. In this report, complete coverage without exceeding the board is considered by either providing some assumptions about the shape of the area \cite{RN133,RN132,RN172} or employing the assumption for the boundary effect \cite{RN147,RN171}. 

Current research about coverage and search were mainly in 2D areas which had a relatively easy geographical feature and can be applied to the plane ground and the surface of the water. However, there are a considerable number of applications in 3D terrains, such as detection of the location of the forest fire \cite{RN263}, underwater search and rescue \cite{RN115}, oceanographic data collection \cite{RN271} ocean sampling, ocean resource exploration, tactical surveillance \cite{RN261,RN269,RN148} and hurricane tracking and warning \cite{RN265,RN148}. The problems in the 3D area bring challenges in many aspects \cite{RN235}. The area of interest in 3D spaces, such as underwater area or airspace, is usually huge and the telecommunication and sensing ranges need to be improved from circular ranges to spherical ranges which may be solved by attaching more sensors on robots or using better sensors. The mobility of robots needs to have more \ac{DOF}, which requires entirely different mechanical structure as well as advanced controllers and sensors. Thus, tasks in 3D spaces require more expensive hardware on autonomous robots with higher mobility \cite{RN260}. For the connectivity and coverage in algorithms, some comprehensively researched tasks in 2D areas are still waiting to be proved in 3D areas which will be discussed in detail in the next section. The complex system and task in 3D search and complete coverage lead to extra computational complexity and larger memory space needed. Some algorithms in 2D may not be applied to 3D tasks with simple extensions or generalizations \cite{RN259}. However, several work researched the coverage \cite{RN116,RN117,RN121,RN122,RN138,RN171,RN228,RN229,RN230,RN235,RN259,RN269}, search \cite{RN118,RN119} and trajectory tracking \cite{RN190} problems in 3D areas \cite{RN234}. Those algorithms can be used in the flight in a 3D terrain \cite{RN190,RN192,RN141,RN158} and the exploration in underwater environment for rescue tasks \cite{RN115,RN227,RN15,RN252,RN121,RN122}. Inspired by the search for MH370 in the underwater environment, this report develops algorithms in a 2D area first and modifies them to be suitable to be applied on a 3D area.

In complete coverage and search, the ranges of sensors are critical parameters. There are various assumptions for the sensing ability of sensors \cite{RN246}. Most papers \cite{RN134,RN118,RN119,RN138,RN229,RN234,RN230,RN101,RN232,RN233,RN1,RN147,RN171,RN147,RN171} assumed a fixed circular sensing range in 2D tasks and a spherical range in 3D tasks. However, some papers used a directional sensing range especially for some surveillance tasks using cameras \cite{RN235, RN240,RN241,RN236}.  Reference \cite{RN243,RN244} considered the sensor with a flexible sensing range so that the combination of different ranges may reach the maximum coverage rate and save the energy. \cite{RN242} examined the adjustment for both the direction and the range. \cite{RN238} discussed the situation that sensors are positioned with only approximate accuracy because of practical difficulties and provided the method to set (near-)optimal radius or errors allowed in the placement. Boukerche and Fei \cite{RN237} used irregular simple polygon assumption for the sensing range and solved the problem in recognizing the completely covered sensors and finding holes. In this report, all the sensors for both targets and obstacles are ideal and omnidirectional.

\section{Mapping}
Geometric maps \cite{RN140} and topological maps \cite{RN34,RN109} are two kinds of methods for robots to describe the environment. Although the geometric map is more accurate containing detailed information of every point of the area, it needs a significant amount of memory and considerable time for data processing, especially for data fusion which deals with a huge amount of data collected from other robots. This kind of work can be seen in navigation using \ac{SLAM} such as \cite{RN140,RN275,RN276}. In contrast, the topological map only needs certain points of the area which is more efficient and applicable. To reduce the memory load, simplify the control law and save energy in calculations, a topological map should be used. 

Grid map is a kind of topological map. There are three common ways to generate the grid map from literature. The first way to get the topological map is generating a Voronoi tessellation \cite{RN121,RN145,RN161,RN162,RN266}. According to Aurenhammer \cite{RN277} and Wang \cite{RN246}, a Voronoi diagram with $ N $ sensors form $ s_{1} $ to $ s_{N} $ in an area can be defined as a separation of the plane into $ N $ subsection with one sensor in each subarea.  In a subarea, the distance from the sensor to a point should be smaller than the distance from any other sensor to that point. If the subsections of two sensors have a common border, these two sensors are called neighbor sensors. Voronoi cells are usually used in sensor deployment to reach a complete coverage. However, with initial random deployment, the mobile sensors may not be able to communicate with others due to a limited communication range. Another cell division can be seen in \cite{RN278} which used the square grid map. The robots go from the center of one cell to another cell and visited cells are remembered. This square grid representation is usually used in the graph search and the tree search algorithms see, e.g., \cite{RN160}. The third way to generate the coverage grid is based on uniform tessellation in 2D area and uniform honeycomb in 3D space. 

The third method is used in this report thus it is discussed in detail. Using this approach, robots will move between the vertices of the 2D grid or the center of the 3D cells in each step to search the area or reach the complete coverage. Therefore, only the coordinates and visitation states of the vertices need to be included in the map which is simpler than the grid generated by Voronoi partition. The uniform division is used as each robot has the same ability so each cell a robot occupied in complete coverage or visited in search needs to be the same to maximize the available sensing range. It also simplifies the algorithm design as each step has a similar pattern. To cover a 2D area with the grid, a topological map can be made using an \ac{T} grid, a \ac{S} grid or a \ac{H} grid. \cite{RN45} proposed that the T grid pattern is asymptotically optimal as it needs the fewest sensors to achieve the complete coverage of an arbitrary unknown area. With collisions allowed and the assumption that robots have no volume, Baranzadeh \cite{RN1, RN232,RN233,RN101} used the T grid which contains the fewest vertices as the communication range equals $\sqrt{3}$ times the sensing range. However, in the simulation of those papers, that relation was not always used. In fact, when considering collision avoidance, the volume of the robot and the obstacle sensing range, the T grid could not always contain the fewest vertices in different conditions and could only guarantee the complete coverage with the assumption of the boundary effect \cite{RN132,RN133}. 

Unlike the proved result in 2D coverage, there is no proven conclusion to shown which grid pattern could use the smallest number of sensors to cover a 3D space completely. Coverage related works in 3D include Kelvin's conjecture \cite{RN312} for finding the space-filling polyhedron which has the largest isoperimetric quotient and Kepler's conjecture \cite{RN313} for sphere packing in cuboid area \cite{RN121}. Based on proposed cells in these two conjecture and considering uniform honeycomb, \ac{TO} cells, \ac{RD} cells, \ac{HP} cells, and \ac{C} cells could be used, and the honeycomb with the minimum number of cells under different relations between the communication range and the target sensing range is given in \cite{RN122,RN279}. Then \cite{RN119,RN118,RN138,RN229,RN234,RN230} chose the TO grid based on their common assumption of the large ratio of the communication range to the target sensing range. However, \cite{RN229} did not provide the ratio and all these paper ignore collisions. In this report, to guarantee a strict 100\% 2D complete coverage of the task area, only the T grid could be used with assumptions about the passage, the relations between different parameters and the curvature of the boundaries based on \cite{RN132,RN133}. When a complete coverage is not required and the assumption of the boundary effect is used, the way to choose the suitable grid under different situations for both 2D and 3D areas is discussed. In simulations, the grid pattern is chosen based on the parameters of the potential test robots so that the simulation results can be applied and verified directly in experiments. 

When designing the grid map, there are some papers considering the connectivity of the network \cite{RN4,RN121,RN122,RN279,RN126,RN282}. A mobile sensor may need to have more than one communication neighbors in case some sensors fail. Thus, the extra communication neighbors could provide redundancy of the network and improve the reliability. However, some papers only considered the general space-filling problem with the assumption of boundary effect but with no obstacles \cite{RN122,RN279,RN126}. The area could be a convex polygon in 2D areas or a convex polyhedron in 3D spaces \cite{RN4,RN282} as the assumption of passage width and the volume of robots are not given. \cite{RN121} limited the shape of the area further to a cuboid based on the sphere packing problem. So, in a real task area with irregular shape with obstacles inside, their methods for multi-connectivity is invalid. In this report, a realistic unknown area is considered, one sensor only needs to have at least one communication neighbor. 

There are also some algorithms without map especially the algorithm inspired by animals \cite{RN287,RN293,RN295,RN281,RN283}. Animal foragers need to explore the area by only detecting the randomly located objects in the limited vicinity using the random walk. The best statistical strategy of optimizing the random search depends on the probability distribution of the flight length given by \cite{RN103,RN105,RN247,RN281,RN283}. When the targets are sparsely and randomly distributed, the optimum strategy is the \ac{LF} random walk which can be seen in marine predators, fruit flies, and honey bees \cite{RN106}. It uses the L\'evy distribution to generate the step length and the normal distribution to generate the rotation angle so that robots have fewer chances to go back to previously visited sites \cite{RN102,RN107,RN291,RN286,RN290,RN278,RN284}. Thus, robots could move freely and detect all points in the task area. This scheme means the \acl{LF} algorithm does not need a map to guarantee a complete search of the area so that exchanging information on the visited areas is also unnecessary and only the positions of visited targets need to be communicated. This algorithm, on the one hand, largely decreased the demand for memory and on the other hand, is a kind of a waste of the communication ability. If the number of targets is unknown, the \acl{LF} algorithm could not be used as robots will not know when the area is completed searched as no map are recorded. Reference \cite{RN118,RN119} considered a combination of the \acl{LF} algorithm and the TO grid for both static target and moving target. In those two papers, the step lengths were some fixed distances from one vertex to any other vertex, and the directions were the corresponding direction from current vertex to the destination vertex. The combined algorithm decreased the search time comparing to each single algorithm. However, the algorithm only considered the situation that the number of targets is known but did not consider obstacle avoidance, so the search area was simple. The algorithm also did not claim the sensing range for the boundary thus how to reject the long step which would exceed the boundary is unknown. This report will compare the proposed algorithm with animal inspired algorithms in \cite{RN102,RN292} and always consider collision avoidance. The algorithm is available for both when the number of targets is given and when it is unknown.   

Some current algorithms in coverage problems were heuristic \cite{RN278,RN242,RN280} but there were still several papers with grid maps having rigorous proofs \cite{RN5,RN44}. \cite{RN2} achieved 100\% coverage of an unknown 2D area with a mathematical proof, but it exceeded the boundary of the area which only has limited applications such as using the UAV in open airspaces to cover the ground or the sea surface. If there are solid walls or coastlines and ground robots are used, this algorithm could not work. If the boundary of the area is a borderline in a military application, exceeding the boundary is also forbidden. Although \cite{RN1} used a T grid inside the boundary, the routes of robots still exceeded the boundary which can be seen in its simulation figures. The reason for this problem is that the author thought the assumption that all accessible vertices are connected equals to the assumption that accessible vertices can be reached by moving along the shortest path, however, there may be no vertices to go through in that path. Even if there were vertices in the shortest passage, they might be too close to the boundary considering the volume of the robots. Thus, those vertices should be thought as blocked so that robots should choose a longer path. If two vertices are reached from two unconnected edges, the section between the two vertices may still be undetectable because of the short sensing range which is only set to be greater than zero in \cite{RN1}. This report proves the convergence of all the proposed algorithms without asking robots to go beyond the boundary.

\section{Collision-Free Navigation}
For a multi-robot system,  the increased number of robots leads to a higher possibility for robots to encounter each other also rises.  However, some algorithms for single robot navigation are no longer available. Therefore, collision avoidance is a significant and attractive research topic in multi-robot systems. 

Navigation strategies can be categorized into global methods and local methods. The global one is off-line path planning. Given the global information about the environment and obstacles as a priori, the robots will build a model of the scenario \cite{RN51} and find an optimal traceable path to solve the problem and avoid collisions \cite{RN52}. If a path is known, an appropriate velocity profile will be planned for collision avoidance \cite{RN53}. This problem can be thought as resource allocation thus corresponding algorithms can be utilized. Although this method is robust to deadlock situation, the disadvantages of it are clear that the environment of the system should be known and be structured for the algorithm to segment it to separated pieces \cite{RN7}. These disadvantages lead to a centralized thought of the problem and cannot be conveniently and easily modified to a decentralized approach. \cite{RN296} proposed an algorithm in a known area. It used ring-shaped local potential fields to have a path-following control of a nonholonomic robot. In some situations, a part of the information is given. For instance, the edge of the area in \cite{RN294} was known and in \cite{RN138,RN230}, the equation to describe the covered area was given. \cite{RN298} provided semantic information to assist the exploration. In many path planning and navigation problems like \cite{RN299}, the limited knowledge was given which was the direction from the starting point to the target. This is possible in a workspace or factory floor by installing radio beacons at the target so that the robot can track the beacon to plan the path. 

For the local or reactive method, algorithms are based on sensed and shared local information. The drawback of some algorithms with this approach is that they are not based on the kinematic model with realistic constraints of the robot. Instead, they used heuristic methods. Without justification and mathematical proof, some methods may fail in special circumstance. For example, the control algorithm in \cite{RN54} would work for two or more robots under the conditional that the robot had a constant speed. If the acceleration was limited for a holonomic robot, a potential field approach could support the control of three robots at most based on \cite{RN55}. To control an unlimited number of robots with no collision, \cite{RN56} used robots with unbounded velocity and \cite{RN57} assigned the velocity of the robot as the control input. In reactive methods, the artificial potential field algorithm is widely used \cite{RN262,RN15,RN300,RN301,RN302,RN296,RN55} as it generates a safe path to the target with little computational effort. It generates a gradient field between robots and targets to direct the movement of robots \cite{RN108}. In that field, goals like the targets generate the attractive force and act like valleys in the field. Obstacles and boundaries generate the repulsive force, acting like peaks in the field. So, the superposition of the two fields will drive robots to go downhill towards targets while moving away from obstacles. One disadvantage of the algorithm is that robots can be trapped in local minima which results in the failure of the navigation task. Therefore, \cite{RN102,RN292} only used the repulsive force between robots to disperse them to move in different directions, which is also utilized in this report based on \cite{RN132,RN133,RN147}.

Examples in the above two paragraphs were all considering steady obstacles. More methods in this category include dynamic window \cite{RN58}, the lane curvature \cite{RN59} and the tangent graph based navigation \cite{RN51}. For dynamic obstacles including moving and deforming obstacles, it is tough to solve. Thus, many methods put constraints on the moving velocity of obstacles or the changing speed of shapes. The obstacles were usually explained as rigid bodies with a selected shape. Moreover, in \cite{RN51}, the characteristic point of the robots in concern with its global geometry is calculated, and velocity of the obstacles is achieved. A local method with range-only measurements in \cite{RN98} allowed the shape of obstacles to be a time-vary variable, but it had a fierce limitation on the robots' velocity. Inspired by binary interaction model, a reactive approach with integrated sensed environment presentation was illustrated in \cite{RN51}. This method did not require approximation of the shapes of obstacles and velocity information of obstacles. It would find a path between obstacles without separating obstacles. In the deployment and complete coverage part of this report, local navigation is needed. With a grid pattern, collision avoidance is easier. In the beginning, robots only know their initial positions are within the search area without other knowledge about the area. However, while moving, the robots are generating their maps about the visited areas so that the global method could also be used.

A large percentage of research in coverage, search, formation, and consensus did not consider the problem of collision avoidance such as \cite{RN1,RN2,RN118,RN119,RN138,RN229,RN234,RN230}. Most of them used simulations to verify the algorithms, and a robot is always assumed as a particle point without the volume. \cite{RN2,RN1} assumed that if more than one robots chose the same vertex to go, those robots would stay around the vertex and the vertex was thought as a parking lot. However, it did not show how to choose a parking space. Even with a parking space choosing mechanism, when the number of robots increased, there would be more robots choosing the same vertex so that the car park model is not available. Therefore, in this report, robots would not choose together, and the communication range should be designed large enough so the choice can be known by all robots which may select the same vertex in this step. Although there will be more communication cost, the robots can avoid collisions while choosing the next step instead of during moving. In \cite{RN232,RN233}, robots could move with different distance in one step which resulted in collisions during moving. However, the velocity of the movement was not stated, and the strategy for collision avoidance was not stated clearly. Also, the proposed algorithm in \cite{RN233} could not match its result in its experiment. Therefore, in this report, robots always move with the unit step length to avoid this problem. 

\section{Contribution Highlights}
The main contributions of the report can be described as follows:
\begin{enumerate}
    \item One decentralized algorithm for the complete coverage is proposed for multiple mobile robotic sensors to deploy themselves in both 2D and 3D areas. The detailed descriptions of the coverage algorithms are presented.
    \item The report proposes three decentralized random algorithms for the search task in 2D or 3D areas. They are improved step by step. 
    \item The task areas for the four algorithms are arbitrary unknown with obstacles inside. For complete coverage in a 2D area, assumptions about a large target sensing range and the curvature of boundaries are given. For other problems, if the boundary effect is ignorable, these two assumptions are not needed.
    \item All algorithms utilize grid patterns to cover the area and robots move between vertices of a grid. The methods to choose the best grid, namely, the grid with the fewest number of vertices in a 2D area and the fewest cells in a 3D area, are proposed considering the relation between different parameters. In some previous algorithms, the sensing range for obstacles was not considered. 
    \item All algorithms consider collision avoidance between robots and between robots and boundaries. Also, realistic constraints of robots including the volume are set based on potential test robots. Robots choose the next step in a sequence, and they are always synchronized.
    \item Simulations in MATLAB are used to demonstrate the performance of the algorithms. Another complete coverage algorithm and other five algorithms for search tasks are simulated in the same area to show the effectiveness of the proposed algorithms. 
    \item The convergence of all algorithms are mathematically proved which is different from many heuristic algorithms.
    \item The algorithms are suitable for both 2D areas and 3D areas. Only slight modification based on the grid structures are needed. So it is easy to adjust them to apply to the other dimension. 
    \item For the three search algorithms, the number of targets can be known or unknown.
    \item Using the simulation result of the second search algorithm, the way to generate the mathematical relation between the search time, the size of the area and the number of robots is given. The effects of other parameters are also discussed.
    \item Experiments for the second search algorithm is conducted on Pioneer 3-DX robots to verify its performance in real situations and to help design errors of the algorithm.      
\end{enumerate}

\section{Report Outline}
The remainder of the report is organized as follows:

Chapter 2 describes the method to choose the best grid pattern to cover the area for the four algorithms considering related parameters of robots. It introduces the basic definitions about the task area with assumptions for complete coverage and for coverage ignoring the boundary effect. The grid pattern for potential test robots which are the Pioneer 3-DX in 2D and the Bluefin-21 in 3D are chosen. Chapter 3 provides a decentralized complete coverage algorithm for the self-deployment of multiple robots in both 2D and 3D areas without collisions. The convergence of the algorithms is proved with simulations to show its efficiency comparing to another algorithm. Then three distributed search algorithms in 2D and 3D areas are presented in Chapter 4, 5 and 6 respectively. Rigorous mathematical proofs of the convergence of these algorithms are provided, and they are all simulated in MATLAB with comparisons to other five algorithms to show the effectiveness. In Chapter 4, a decentralized random algorithm for search task is provided by using the visitation state of the neighbor vertices. The second search algorithm in Chapter 5 adds potential field algorithm on the first one to disperse robots and decrease the repeated routes. In 2D area, some factors affecting the search time are analyzed and the way to find the relation between search time and other parameters are given. Also, experiment on Pioneer 3-DX robots is used to check the performance of the algorithm. Chapter 6 adds breadth-first search algorithm on the algorithm in Chapter 4. It helps robots to jump out of the visited area by looking for the nearest unvisited vertex in their local maps. The three proposed search algorithms are also compared with each other to analyze their own merits and demerits. Chapter 7 concludes the report and provides some possible future research topics to improve or to extend the presented work.
\chapter{Grid Selection for Coverage in 2D and 3D Areas}
\chaptermark{Grid Selection for Coverage in 2D and 3D Areas}
For complete coverage of an unknown area, mobile sensors are put into the area and need to move to some places which form a grid to maximize the coverage of the team. For search problems, robots move around the area until all targets are found, or the entire area is detected. To ensure the whole area is sensed, robots also need to have a map. Moreover, to cover or search the area easily, the grid pattern could be used as a topological representation of the area to decrease the information needed during movement of the two tasks. Then in a 2D space, the robot can move from a vertex of the grid to one of the nearest neighbor vertices straightly, and in a 3D area, robots can move straightly from the center of a polyhedron to one of the closest centers of another polyhedron. As the robot has a sensing range for the target, after visiting all the vertices, the whole area can be searched if the distance between two cells of the grid is set reasonably. Similarly, in the complete coverage task, after mobile sensors are deployed to those nodes, the whole area is covered. With a grid pattern, robots only need to record the positions of the vertices in a 2D area or the centers of polyhedrons in a 3D area to build their maps. Thus, this topological method also decreases the memory and calculation load comparing to the geometrical method.

Looking for a grid to cover an area is the same as the tessellation problem for a plain and the honeycomb for a 3D space. As robots in the system have the same sensing ranges and the same communication range, regular tessellation and space-filling could be considered to simplify the design of the control rule further. Different grids have different numbers of vertices. If the same control algorithm is applied, it is intuitive that the search time will be affected by the number of vertices. When robots select the next vertex, they need to ensure that there are no obstacles on the path to that vertex, so the obstacle sensing range is considered in grid selection. In the selected grid, robots also need to form the sensing coverage of the whole area and communicate with neighbor robots to exchange information so the target sensing range and the communication range should be considered. 

This chapter finds the grid with the minimum number of vertices to cover the area considering different coverage requirements and all kinds of relations between parameters about ranges of the robots. The selection criteria are applicable for both complete coverage and search tasks. Section 2.1 considers the problem in a 2D area while Section 2.2 considers the selection criteria a 3D area. A summary of the grid selection is shown in Section 2.3.
\section{Grids for 2D Areas}
This section illustrates that only the T grid can provide a 100\% complete coverage.  It also proposes the general method to select the grid with the least number of vertices in a 2D area if areas near boundary can be ignored by considering the communication range, the target sensing range and the obstacle sensing range of robots. Then the rule is applied to a Pioneer 3-DX robot to see which grid is suitable for it in a search task.

\begin{figure}
    \centering
    \subfloat[An \acl{T} grid]{
        \begin{minipage}[t]{0.31\textwidth}
            \centering
            \includegraphics[width=1\linewidth]{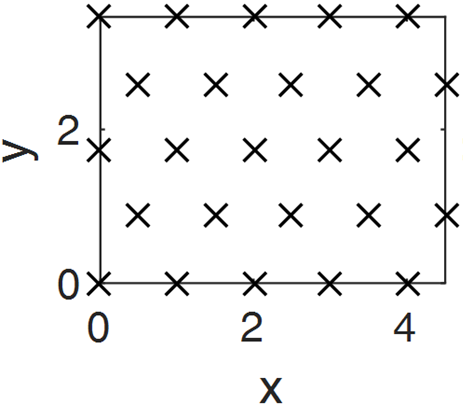}
        \end{minipage}
    }
    \subfloat[A \acl{S} grid]{
        \begin{minipage}[t]{0.31\textwidth}
            \centering
            \includegraphics[width=1\linewidth]{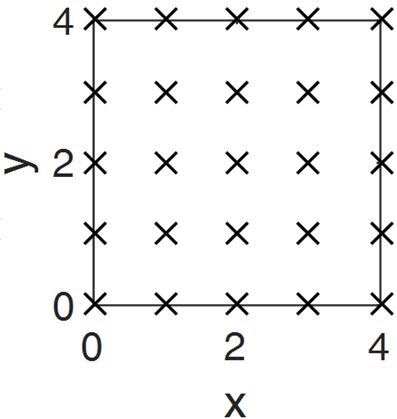}
        \end{minipage}
    }
    \subfloat[A \acl{H} grid]{
        \begin{minipage}[t]{0.31\textwidth}
            \centering
            \includegraphics[width=1\linewidth]{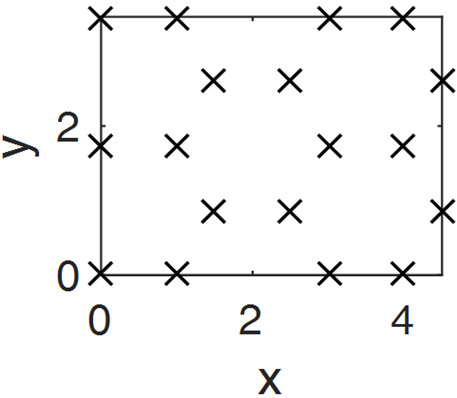}
        \end{minipage}
    }
    \caption[Three regular tessellation patterns]{Three regular tessellation patterns}
    \label{fig:1}
\end{figure}

The tessellation of a plane surface is tiling it with one or more geometric shapes. Each of the shapes can be called as a tile. When considering the same tiles for tessellation, three forms for regular tiling can be used which are \acl{S}, equilateral triangle, and hexagon as seen in Figure \ref{fig:1}. If only considers the circular sensing range for targets, this problem is the same as allocating base stations to get the maximum coverage area in telecommunication. Then based on \cite{RN303}, the T grid should have the least number of vertices. However, in this report, the communication range and the obstacle sensing range are also discussed. Thus, the result in \cite{RN303} may no longer be available. However, \cite{RN1,RN2,RN101,RN232,RN233} still used the T grid to design the algorithm and based on this pattern to find the relation between the target sensing range and the communication range. The sensing range for obstacles was set equal to the side length of the tile. However, in real applications, the search or coverage algorithms may be applied to different robots. Thus, the problem needs to be thought oppositely, namely, choosing the grid based on the parameters of robots. There are two kinds of requirements for selecting the grid. One is strict as in \cite{RN2}, and the other is loose which can be seen in \cite{RN1,RN122,RN279,RN119}. Next, the assumptions for different ranges are given followed by the discussion subjected to these two requirements.

\subsection{Definitions and Assumptions}
\label{s1}
This section takes advantage of the method of definitions in \cite{RN133}. Examples to illustrate the relation between ranges are given using a T grid.

\begin{assumption} 
    The search area $ A\subset{\mathbb{R}^{2}} $ is a bounded, connected and Lebesgue measurable set. Robots know nothing about the area initially.\label{a1}
\end{assumption}  

There are $ m $ robots labeled as $ rob_{1} $ to $ rob_{m} $ in $ A $. For the basic tiling shape, the side length of the shape is represented as $ a $ and robots can only move the distance $ a $ along the side in each step from one vertex of the grid to one of the nearest neighbor vertices.

\begin{assumption} 
    In the 2D area, all the ranges for robots are circular.\label{a2}
\end{assumption}

\begin{figure}
    \centering
    \includegraphics[width=0.8\linewidth]{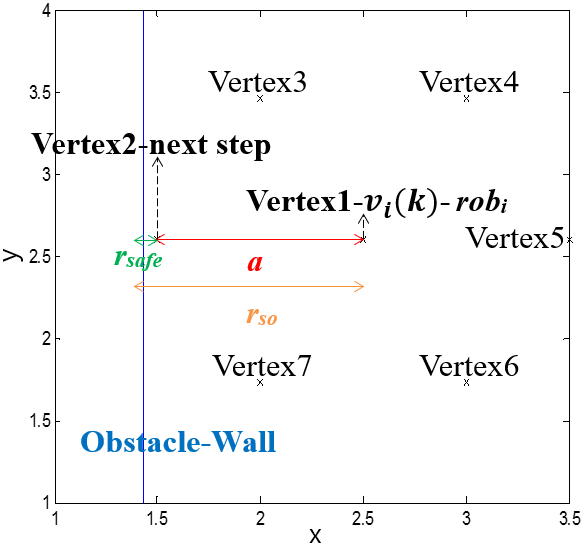}
    \caption[The illustration for $ a $, $ r_{safe} $ and $ r_{so} $]{The illustration for $ a $, $ r_{safe} $ and $ r_{so} $}
    \label{fig:5r_ob}
\end{figure}

Robots have a swing radius $ r_{rob} $. Let the maximum possible error containing odometry errors and measurement errors be $e$. Then a safety radius $ r_{safe} $ can be defined to avoid collision.
\begin{definition}
    To avoid collisions with other things before move, a safe distance $ r_{safe} $ should include both $ r_{rob} $ and $e$ (see Figure \ref{fig:5r_ob}). So $ r_{safe}\geq{r_{rob}+e} $.\label{d1}
\end{definition}

\begin{figure}
    \centering
    \includegraphics[width=0.8\linewidth]{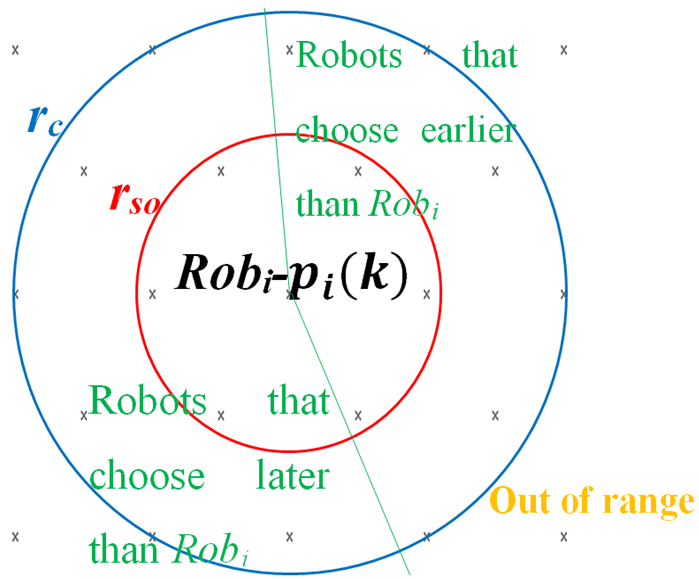}
    \caption[$r_{so}$, $ r_{c} $ and the selection order in search]{$r_{so}$, $ r_{c} $ and the selection order in search}
    \label{fig:6rangeorder}
\end{figure}

Thus, vertices which are within $ r_{safe} $ away from the boundaries or obstacles should not be visited. In Figure \ref{fig:5r_ob}, vertex2 is within $ r_{safe} $ of the obstacle so it cannot be reached. This means the obstacles around vertex2 need to be detected from $ rob_{i} $ so it will not choose to move there. Thus, sensors on robots should have a large enough sensing range $ r_{so} $ for obstacles. So $ r_{so} $ should satisfy the following assumption and can be seen in Figure \ref{fig:5r_ob} and Figure \ref{fig:6rangeorder}. 
\begin{assumption}
    $ r_{so}\geq{a+r_{safe}} $.\label{a3}
\end{assumption}
The vertices which are within $ r_{so} $ of $ rob_{i} $ are called sensing neighbors of $ rob_{i} $. 

Robots are equipped with wireless communication tools with the range $ r_{c} $. Those robots within $ r_{c} $ of $ rob_{i} $ are called communication neighbors of $ rob_{i} $. The communication between robots is temporary as in an ad-hoc network so each robot can be considered equal with a local communication range. To avoid having the same choice with other robots in the same step, robots should choose in a certain order and tell the choice to neighbor robots which are one or two vertices away from it. A possible order for search tasks given in Figure \ref{fig:6rangeorder} is that robots on the right side and the top of the current robot should have higher priorities. This order will be explained in Chapter 4 in detail. Considering the error $e$, $r_{c} $ should satisfy Assumption \ref{a4}.
\begin{assumption}
    $ r_{c}\geq{2*a+e}$.\label{a4}
\end{assumption}

Now, the length of a step can be shown as an inequality:
\begin{equation}\label{e1}
a\leq{min(r_{so}-r_{safe},(r_{c}-e)/2)}
\end{equation}

\subsection{The Strict Requirement}
\begin{figure}
    \centering
    \includegraphics[width=0.7\linewidth]{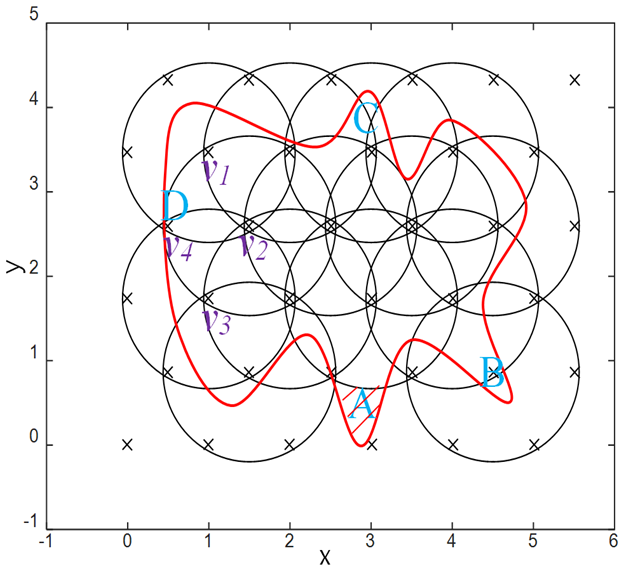}
    \caption[Curvature example for the \acl{T} grid]{Curvature example for the \acl{T} grid}
    \label{fig:7curve}
\end{figure}

\begin{figure}
    \centering
    \includegraphics[width=0.7\linewidth]{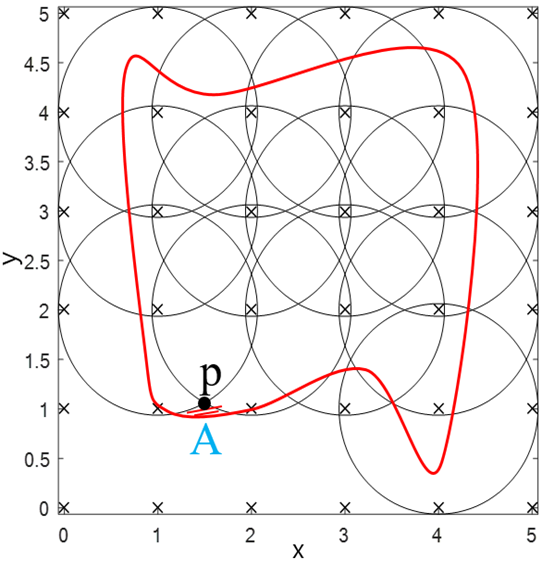}
    \caption[Example of the curvature in the S grid]{Example of the curvature in the S grid}
    \label{fig:8sqcurve}
\end{figure}

\begin{figure}
    \centering
    \includegraphics[width=1\linewidth]{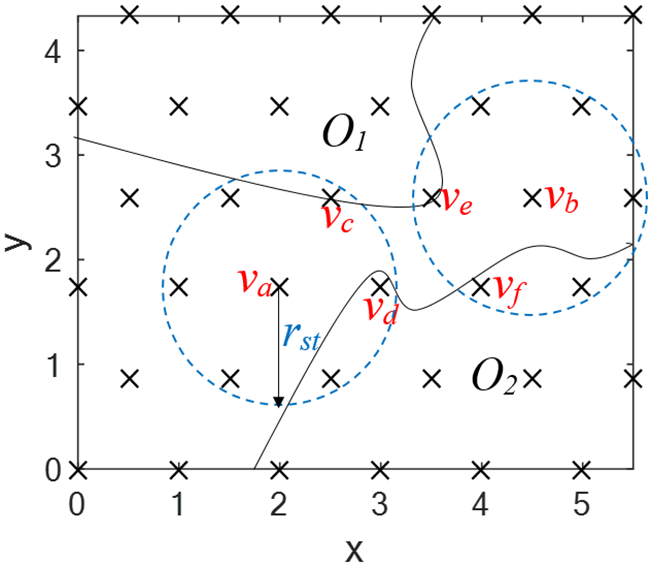}
    \caption[A narrow path]{A narrow path}
    \label{fig:9narrow}
\end{figure}

This requires robots to have a complete search through the grid. Only a T grid can be used under this circumstance with the assumptions \ref{a5} \ref{a6} and \ref{a7} below.
\begin{assumption}
    $ r_{st} \geq{a+r_{safe}}$.\label{a5}
\end{assumption}
This assumption for the sensing range for targets guarantees that areas around inaccessible vertices can still be detected. For example, in Figure \ref{fig:7curve}, black circles have a radius of $a+r_{safe} $ and the red line is the boundary. It can be seen that vertex $ v_{4} $ is less than $ r_{safe} $ away from the boundary so it cannot be visited. But the Assumption \ref{a5} guarantees that the area D near $v_{4}$ can be detected by robots at vertices $ v_{1} $, $ v_{2} $ and $ v_{3} $.
\begin{assumption}
    The curvature of the concave part of the boundary should be smaller than or equal to $1/ r_{st} $.\label{a6}
\end{assumption} 
Without this assumption, some parts of the searched area may not be detected such as section A in Figure \ref{fig:7curve}. In that figure, circles enclose all the sensed area and section A, B and C have segments with curvatures greater than $1/r_{st} $. It can be seen clearly that section A cannot be detected or visited although section B and C can be detected luckily. 
The reason that the S grid and the H grid are not used is that there are no such kinds of limitations to guarantee a complete coverage. For the S grid, Figure \ref{fig:8sqcurve} demonstrates that the setting of curvature will be useless as there is no vertex near the intersection point $ p $ of sensing ranges. Thus, a curve with any curvature could go beyond $p$ and left a section being undetected like section A. The H grid has the same problem. 
\begin{assumption}
    Let $ W_{pass} $ represents the minimum width of the passages between obstacles or between obstacles and boundaries. Then $ W_{pass}\geq{a+2*r_{safe}} $. \label{a7}
\end{assumption} 
This assumption guarantees that there is always at least one accessible vertex in the path. Otherwise, if the passage is not large enough as in Figure \ref{fig:9narrow}, robots at $ v_{a} $ will not arrive at $ v_{b} $ as robot can only move $a$ in a step but vertices $ v_{c} $, $ v_{d} $, $ v_{e} $ and $ v_{f} $ are inaccessible, leaving a section in between undetectable. 
\subsection{The Loose Requirement}
This only requires that all vertices are sensed because it has an assumption about the boundary effect which can be seen in \cite{RN147,RN171,RN122,RN279}.
\begin{assumption}
    The search area is usually very large. So, $a$ is much smaller than the length and the width of area $A$. Therefore, the boundary effect is negligible. \label{a8}
\end{assumption} 
This assumption means that there are no targets in the area near the boundary such as the slashed area under point $p$ in Figure \ref{fig:8sqcurve} so robots do not need to completely sense the area and Assumption \ref{a6} is not needed. Thus, all the three kinds of grids can be discussed. Then Assumptions \ref{a5} and \ref{a7} for $ r_{st} $ and $ W_{pass} $ are made differently based on the structure of each grid pattern. Here, $ W_{pass} $ is still used to design the minimum passage width so the passage can be passed by a robot. But the usage of $ r_{st} $ is to ensure that all the sections between accessible vertices are completely sensed which is different from the requirement that all sections are completely sensed in the strict requirement. Table \ref{t1} shows the relation between $ r_{st} $, $a$, $ r_{safe} $ and $r_{rob}$ and the value of $ W_{pass} $ represented by $a$ and $ r_{safe} $ for each grid pattern. The subscript of $a$ is the notation for the name of the corresponding grid.

\begin{table}
    \centering
    \renewcommand\arraystretch{2}
    \begin{tabular}{|c|c|c|}
        \hline 
        Shape&$ r_{st} $&$ W_{pass} $  \\ 
        \hline 
        T&$a_{T}\leq{\sqrt{3} ( r_{st} - r_{safe} +r_{rob} )}$&$a_{T}+2r_{safe} $  \\ 
        \hline 
        S&$a_{S}\leq{\sqrt{2} ( r_{st} - r_{safe} + r_{rob} )}$&$\sqrt{2}a_{S} +2r_{safe} $  \\ 
        \hline 
        H&$a_{H}\leq { r_{st} - r_{safe} + r_{rob} }$&$2a_{H}+2r_{safe}$  \\ 
        \hline 
    \end{tabular}  
    \caption[$ W_{pass} $ and $ r_{st} $ for each grid]{$ W_{pass} $ and $ r_{st} $ for each grid}
    \label{t1}
\end{table}

To use the least number of vertices in the search area, for the same $a$, one vertex needs to have the fewest neighbor vertices so that the polygon (tile) occupied by each vertex has the maximum area. Also, the maximum $a$ should be used. \cite{RN126} found the grid pattern which results in the minimum number of vertices using different $ r_{c} $/$ r_{st} $ for a static coverage problem. This report considers three parameters in choosing the grid pattern, namely, $ r_{c} $, $ r_{so} $ and $ r_{st} $, which will be different based on the used devices. So, the discussion regarding these three parameters for general situations is provided based on the maximum $a$ in the primary results in Inequality \ref{e1} and in Table \ref{t1}. In this subsection, the result in Inequality \ref{e1} is labeled as $a_{1}$ and the result from Table \ref{t1} is labeled as $a_{2}$. To simplify the discussion and illustrate the procedure of the calculation only, $e$ and $ r_{safe} $ which depend on the selected equipment are ignored while collision avoidance is still considered. So the primary results to maximize $a$ are simplified as $a_{1}=min( r_{so} , r_{c}/2)$, $a_{2T}=\sqrt{3} r_{st} $, $a_{2S}=\sqrt{2} r_{st} $ and $a_{2H}=r_{st} $. Then the maximum $a$ can be written as $a=min(a_{1}, a_{2})$. If $a_{1}>a_{2}$, the area is represented by $ r_{st} $ which is the same for the three grid patterns as it is a fixed parameter from the chosen robot. Otherwise, $a=a_{1}$ and the area is represented by $a_{1}$ which is also the same in the three grids. Table \ref{t2} demonstrates the area $S$ occupied by each vertex and $ W_{pass} $ in different conditions. $S$ and $W_{pass}$ have subscripts to indicate the types of the grid. The data in the table have three colors representing the ranking of the values with {\color{red} red} for the large value, {\color{orange} orange} for the medium value and {\color{cyan} cyan} for the small one. According to this table, if $ a_{1}\geq{1.52r_{st}}$, the T grid should be used to have the least number of vertices. If $  1.52 r_{st} \geq{a_{1}}\geq{1.14 r_{st}}$, the S grid should be used. An H grid will be used when $ 1.14r_{st}\geq{a_{1}}$. Although larger $a$ leads to fewer vertices and less search time, it will increase $ W_{pass} $ and thus will shrink the scope of applicability.
\begin{table}
    \centering
    \renewcommand\arraystretch{2}
    \begin{tabular}{|p{2.7em}|p{7em}|p{7em}|p{7em}|p{7em}|}
        \hline 
        Range&$a_{1}\geq{1.73r_{st}} $&$1.73r_{st}\geq{a_{1}}\geq{1.52r_{st}}$&$1.52 r_{st}\geq{a_{1}}\geq{1.41r_{st}}$ &$1.41 r_{st} \geq{a_{1}}\geq{1.23r_{st}}$\\ 
        \hline 
        $S_{T}$&${\color{red} 2.60r_{st}^{2}}$&${\color{red} 0.87{a_{1}}^{2}}$&$ {\color{orange} 0.87{a_{1}}^{2}}$&$ {\color{orange} 0.87{a_{1}}^{2}}$\\ 
        \hline 
        $S_{S}$&$ {\color{orange} 2{r_{st}}^{2}}$&$ {\color{orange}2{r_{st}}^{2}}$ &${\color{red}2{r_{st}}^{2}}$&${\color{red} {a_{1}}^{2}}$ \\ 
        \hline 
        $S_{H}$&${\color{cyan}1.30{r_{st}}^{2}}$&${\color{cyan}1.30{r_{st}}^{2}}$&${\color{cyan}1.30{r_{st}}^{2}}$&${\color{cyan}1.30{r_{st}}^{2}}$\\ 
        \hline 
        $W_{pass_{T}}$&${\color{cyan}1.73r_{st}}$&${\color{cyan}a_{1}}$&${\color{cyan}a_{1}}$&${\color{cyan}a_{1}}$ \\ 
        \hline 
        $W_{pass_{S}}$&${\color{red}2r_{st}} $&${\color{red}2r_{st}} $&${\color{red}2r_{st}} $&$ {\color{orange} 1.41a_{1}}$ \\ 
        \hline 
        $W_{pass_{H}}$&${\color{red}2r_{st}} $&${\color{red}2r_{st}} $&${\color{red}2r_{st}} $&${\color{red}2r_{st}} $\\ 
        \hline 
        Range&$1.23r_{st}\geq{a_{1}}\geq{1.14r_{st}}$&$1.14r_{st}\geq{a_{1}}\geq{r_{st}}$&$ r_{st}\geq{a_{1}}$& \\ 
        \hline 
        $S_{T}$&${\color{cyan} 0.87 {a_{1}}^{2}}    $&${\color{cyan}0.87 {a_{1}}^{2}}    $&${\color{cyan}0.87 {a_{1}}^{2}}$& \\ 
        \hline 
        $S_{S}$&${\color{red}{a_{1}}^{2}}    $&$ {\color{orange} {a_{1}}^{2}}    $&$ {\color{orange} {a_{1}}^{2}}$& \\ 
        \hline 
        $S_{H}$&$ {\color{orange} 1.30{r_{st}}^{2}}$&${\color{red}1.30{r_{st} }^{2}}$&${\color{red}1.30{a_{1}}^{2}}$& \\ 
        \hline 
        $W_{pass_{T}}$&${\color{cyan}a_{1}}$  &${\color{cyan}a_{1}}$   & ${\color{cyan}a_{1}}$  & \\ 
        \hline 
        $W_{pass_{S}}$&$ {\color{orange} 1.41a_{1}}$   &$ {\color{orange} 1.41a_{1}}$  &$ {\color{orange} 1.41a_{1}}$  & \\ 
        \hline 
        $W_{pass_{H}}$&${\color{red}2r_{st}} $  &${\color{red}2r_{st}} $& ${\color{red}2a_{1}} $& \\ 
        \hline 
    \end{tabular} 
    \caption[$ W_{pass} $ and the area occupied by each polygon]{$ W_{pass} $ and the area occupied by each polygon}
    \label{t2}
\end{table}

\subsection{The Choice for the Pioneer 3-DX Robot}
\begin{figure}
    \centering
    \includegraphics[width=0.6\linewidth]{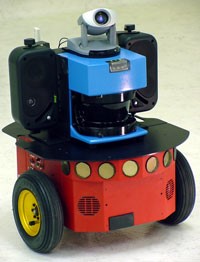}
    \caption[The Pioneer 3-DX robot]{The Pioneer 3-DX robot}
    \label{fig:10robot}
\end{figure}

\begin{figure}
    \centering
    \includegraphics[width=0.8\linewidth]{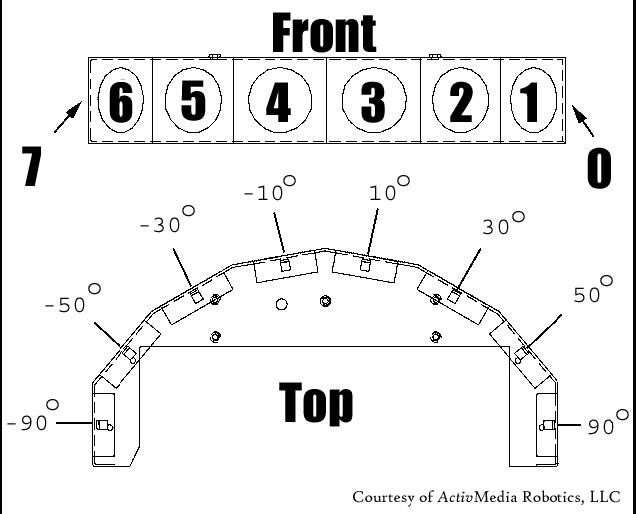}
    \caption[Layout of 8 sonars in the front]{Layout of 8 sonars in the front}
    \label{fig:11sonar}
\end{figure}

\begin{figure}
    \centering
    \includegraphics[width=0.8\linewidth]{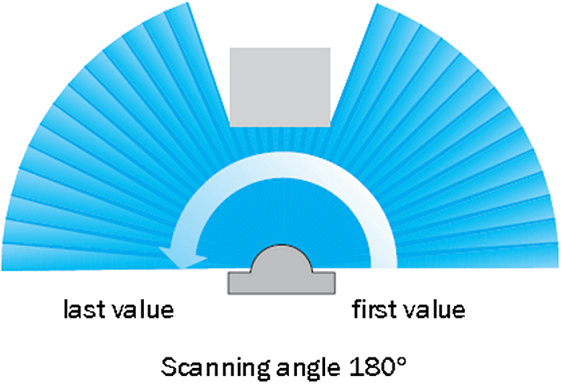}
    \caption[Range of the laser SICK-200]{Range of the laser SICK-200}
    \label{fig:103laser}
\end{figure}

Pioneer 3-DX robots in Figure \ref{fig:10robot} is widely used in research. The second search algorithm in this report also uses it to verify the performance. Therefore, this robot is introduced and the suitable grid pattern for it will be selected. Based on the above discussion, for a complete coverage, a T grid must be used, so only the grid pattern under the loose requirement need to be chosen based on the parameters of this robot. The robot has 16 sonars around which is the yellow circle in the figure. The layout of the 8 sonars in the front can be seen in Figure \ref{fig:11sonar}. There are other 8 sonars at the symmetrical rear side of the robots. Thus, the robots can sense the environment through 360 degrees. The sonar is used to sense obstacles with $0.17\text{m}<r_{so}\leq{5\text{m}}$. The target is sensed by the laser which is the cyan part in Figure \ref{fig:10robot}. The laser model is SICK-200 which has $r_{st}=32m$ for a range of 180 degree which is illustrated in Figure \ref{fig:103laser}. To detect 360 degrees, the robot needs to turn around after the detection of one side is finished. For communication, the robot uses a Cisco Aironet 2.4 GHz Articulated Dipole Antenna AIR-ANT4941 with $ r_{c}=91.4 $ at 1Mbps typically in an indoor environment. In calculations, equality conditions of the inequalities are used. Therefore, the maximum $a_{1}=5$ as $ a_{1}\leq{min(5,91.4/2)} $. As $1.14r_{st}>a_{1}$, the H grid should be used. Base on Table \ref{t2}, this belongs to the last (7th) situation.  From Table \ref{t1}, $a_{2H}=32 $m. So the side length of the H grid is $a=min(a_{1}, a_{2H})=5$m. Then $W_{pass}=10$m.

\section{Grids for 3D Areas}
This section proposes the general method to select the grid pattern to cover a 3D area with the fewest vertices. Different relations between the communication range, the target sensing range, and the obstacle sensing range are discussed in a Table. Then the suitable grid for the UUV, Bluefin-21, will be chosen using that table.

\begin{figure}
    \centering
    \subfloat[\acl{C}]{
        \begin{minipage}[t]{0.3\textwidth}
            \centering
            \includegraphics[width=1\linewidth]{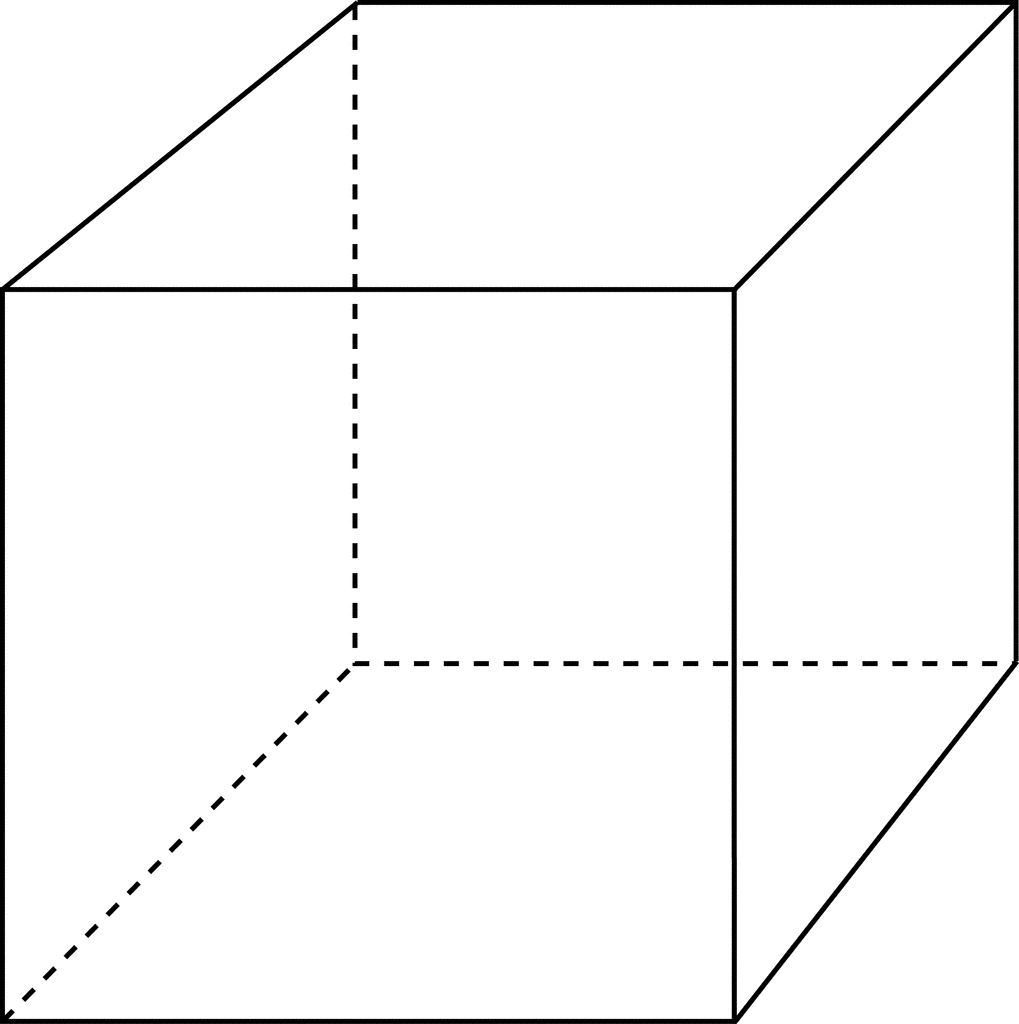}
        \end{minipage}
    }
    \subfloat[\acl{HP}]{
        \begin{minipage}[t]{0.3\textwidth}
            \centering
            \includegraphics[width=1\linewidth]{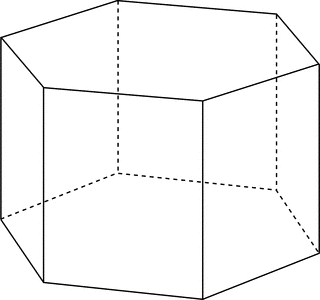}
        \end{minipage}
    }
    \\
    \subfloat[\acl{RD}]{
        \begin{minipage}[t]{0.3\textwidth}
            \centering
            \includegraphics[width=1\linewidth]{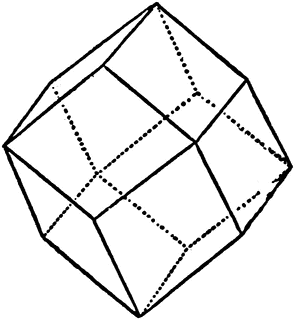}
        \end{minipage}
    }
    \subfloat[\acl{TO}]{
        \begin{minipage}[t]{0.3\textwidth}
            \centering
            \includegraphics[width=1\linewidth]{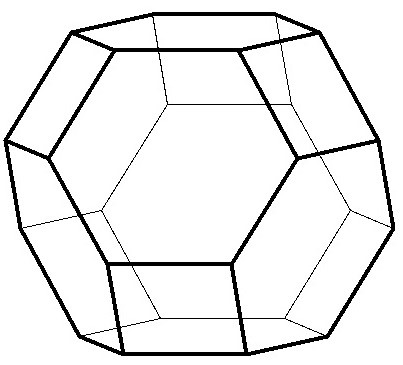}
        \end{minipage}
    }
    \caption[Four uniform honeycombs]{Four uniform honeycombs}
    \label{fig:123D}
\end{figure}

Tessellation in a 3D area is called honeycomb which is filling a 3D space with cells. As discussed in 2D area, the same cell is used to simplify the representation of the area and the control algorithm design. Honeycomb is more complicated than tessellation as there are no proven results for which grid pattern will use the fewest cells to cover a grid if the same sensing range is employed in all grid patterns. However, based on similar problems such as Kepler's conjecture and Kelvin's conjecture \cite{RN122,RN279}, possible polyhedrons include the C grid, the HP grid, the RD grid and the TO grid as seen in Figure \ref{fig:123D}. \cite{RN122,RN279} also discussed which grid should be chosen under different $r_{c} $/$ r_{st} $. Then \cite{RN229} chose the TO grid without giving the communication range. However, $r_{c}\geq{1.549r_{st}} $ $(\text{which is, } 2\sqrt{3}/\sqrt{5}r_{st})$ is enough to connect all vertices. But when $ 1.587r_{st}\geq{r_{c}}\geq{1.549r_{st}}$, an HP grid should be used. Then \cite{RN119,RN118,RN138,RN234,RN230} considered the situation for $r_{c}\geq{1.789r_{st}}$ $ (\text{which is, }4/\sqrt{5}r_{st})$. However, as collision avoidance was not discussed in detail, these papers did not have $ r_{so} $. Moreover, they were considering the question of designing the best grid pattern to cover the area and setting the ranges based on this grid. But, in reality, the search or coverage algorithms should be applied to different robots carrying devices with various ranges. So this chapter is based on the range related parameters of the robots to find the corresponding suitable grid pattern. Different from 2D problems, this section only discusses the grid selection which complies with the loose requirement. 
\subsection{Definitions and Assumptions}
\label{s2}
This section will take advantage of the method of definitions in \cite{RN147}. Examples to illustrate the relations between ranges are given in a C grid.
\begin{assumption}
    A three-dimensional area $ A\subset{\mathbb{R}^{3}} $ is a bounded, connected and Lebesgue measurable set.\label{a9}
\end{assumption}
There are $ m $ robots labeled as $ rob_{1} $ to $ rob_{m} $ in $ A $. 

\begin{definition}
    If there are edges with different lengths in the grid and robots can visit all vertices through various sets of lengths of edges, then the maximum length of each set is compared with others, and the smallest one is chosen to be the edge length $ a$. Then robots move the distance $a$ from one vertex straightly to an accessible neighbor vertex in one step. \label{d53Da}
\end{definition}

3D grids based on a uniform honeycomb are used, and the center of the polyhedron is the vertex of the grid. Then the grid is built up by connecting the vertices to its nearest surrounding neighbor vertices.  
\begin{assumption}
    All sensors and communication equipment on robots have spherical ranges. The range using the word radius also refers to a spherical radius .\label{a10}
\end{assumption}

\begin{figure}
    \centering
    \includegraphics[width=1\linewidth]{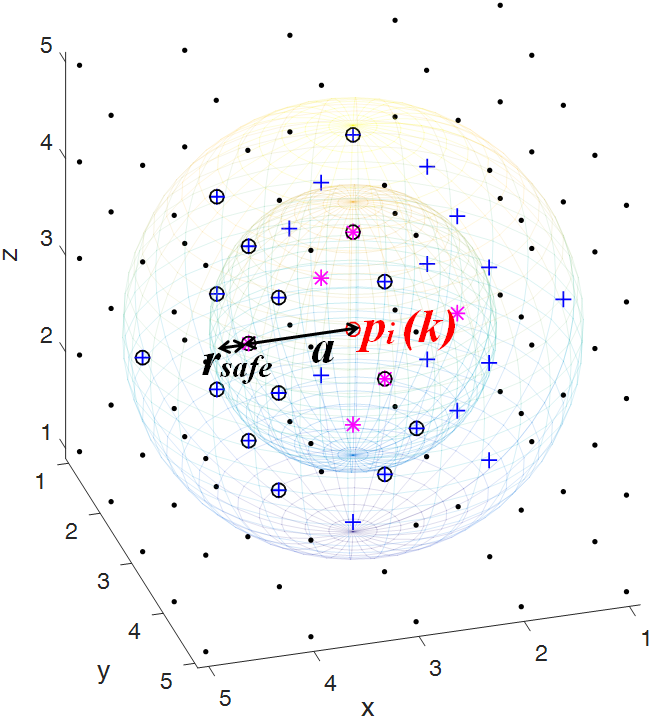}
    \caption[$a,\ r_{safe},\ r_{so},\ r_{c}$ and the order of selection in search.]{$a,\ r_{safe},\ r_{so},\ r_{c}$ and the order of selection in search.}
    \label{fig:16range3d}
\end{figure}

Let the swing radius of robots be $ r_{rob} $. Then the total error of robots is represented as $e$ and the safety radius is $r_{safe}$. Then we have a similar definition as Definition \ref{d1} and two similar assumptions for the obstacle sensing range $r_{so} $ and the communication range $r_{c}$ as Assumptions \ref{a3} and \ref{a4}. The only difference is that the ranges in these definitions are spherical instead of circular. 
\begin{definition}
    $ r_{safe}\geq{r_{rob}+e}$ so that the collisions with other things could be avoid by path planning before a move (see Figure \ref{fig:16range3d}).\label{d2}
\end{definition}
\begin{assumption}
    $ r_{so}\geq{a+r_{safe}} $.\label{a11}
\end{assumption} 
\begin{assumption}
    $ r_{c}\geq{2*a+e}$.\label{a12}
\end{assumption}

So, the length of a step is  
\begin{equation}\label{e2}
a\leq{min(r_{so}-r_{safe},(r_{c}-e)/2)}.
\end{equation}

Vertices which are within $ r_{so} $ of $ rob_{i} $ are called sensing neighbors of $ rob_{i} $. Robots within $ r_{c} $ of $ rob_{i} $ are called communication neighbors of $ rob_{i} $. The communication in the 3D space also uses the ad-hoc network with an omnidirectional antenna to get the spherical coverage. 
\subsection{The Loose Requirement}
In the loose requirement, the area between vertices must be fully covered, and the area between outer sensing spheres to the boundaries can be ignored as the following assumption is used.
\begin{assumption}
    The search task for a team of robots usually has a vast area. So, $a$ is much smaller than the length, the width and the height of area $A$. Therefore, the boundary effect is negligible \cite{RN122,RN279}. \label{a13}
\end{assumption} 

When selecting the type of the polyhedron, the number of polyhedrons to fill the 3D space needs to be minimized, so the maximum $a$ needs to be used. For the same $a$, one vertex needs to have the fewest sensing neighbors to allow each polyhedron to have the maximum volume. For the HP cell, there are two kinds of edges to connect vertices and robots must move along both to visit all vertices so that $a$ equals to the longer one. For a TO cell, there are also two kinds of edges, but robots can visit all vertices through the shorter one only so $a$ equals to the shorter one to maximize the volume of the polyhedron. The relation between $ r_{st} $ and $a$ and listed in Table \ref{t3}. 
\begin{table}
    \centering
    \renewcommand\arraystretch{2}
    \begin{tabular}{|c|c|}
        \hline 
        Shape&$ r_{st} $  \\ 
        \hline 
        C&$a_{C}\leq{\sqrt{3} ( r_{st} - r_{safe} +r_{rob} )/2}$   \\ 
        \hline 
        HP&$a_{HP}\leq{\sqrt{2} ( r_{st} - r_{safe} + r_{rob} )}$ \\ 
        \hline 
        RD&$a_{RD}\leq{\sqrt{2} ( r_{st} - r_{safe} + r_{rob} )}$ \\ 
        \hline 
        TO&$a_{TO}\leq{ 5(r_{st} - r_{safe} + r_{rob} )/2/\sqrt{15}}$  \\ 
        \hline 
    \end{tabular}  
    \caption[$ r_{st} $ for each 3D grid]{$ r_{st} $ for each 3D grid}
    \label{t3}
\end{table}

In the 3D space, the three parameter of robots, namely, $ r_{c} $, $ r_{so} $ and $ r_{st} $, are still considered in the discussion to find the grid with the fewest vertices in different situations. Thus, the results in \cite{RN122,RN279} which considered $r_{c}$ and $r_{st}$ only for a similar problem cannot be used. The discussion in this section uses the maximum $a$ in the primary results in Section \ref{s2} and in Table \ref{t3}, which are represented as $a_{1}$ and $a_{2}$ here. Similarly, $e$ and $ r_{safe} $ are ignored as to simplify the discussion and only illustrate the procedure of calculation. These two parameters are affected by the accuracy of the equipment and the disturbance in the environment. However, collision avoidance is still considered in the algorithm design. The primary results to maximize $a$ are $a_{1}=min( r_{so} , r_{c}/2)$, $a_{2C}=\sqrt{3} ( r_{st} - r_{safe} +r_{rob} )/2 $, $a_{2HP}=\sqrt{2} ( r_{st} - r_{safe} + r_{rob} $,$a_{2RD}=\sqrt{2} ( r_{st} - r_{safe} + r_{rob} $  and $a_{2TO}=5(r_{st} - r_{safe} + r_{rob} )/2/\sqrt{15} $. Thus, the maximum $a=min(a_{1}, a_{2})$. To represent the volume occupied by each polyhedral cell, the same method as in 2D is used. So when $a_{1}>a_{2}$, the area is represented by $ r_{st} $ which is a fixed parameter from the selected robot. Otherwise, $a=a_{1}$ and the area is represented by $a_{1}$ which is calculated using the other fixed parameter $ r_{so} $ $ r_{so}$. Therefore, the value in the table are comparable for the four grid patterns. Table \ref{t4} only shows the volume $V$ with subscript to describe the shape of the polyhedron occupied by each vertex. Colored values are also used in the table to show the comparison result with {\color{red} red} for the large value, {\color{orange} orange} for the medium value and {\color{cyan} cyan} for the small one. The table shows that the C grid should be chosen when $ a_{1}\leq{1.2599r_{st}}$ and the HP or RD grid should be employed when $  1.4142 r_{st} \leq{a_{1}}\leq{1.5431 r_{st}}$. For the TO grid, it is used when $  1.2599 r_{st} \leq{a_{1}}\leq{1.4142 r_{st}}$  and $  a_{1}\leq{1.5431 r_{st}}$ to have the least number of vertices.
\begin{table}
    \centering
    \renewcommand\arraystretch{2}
    \begin{tabular}{|p{2.7em}|p{7em}|p{7em}|p{7em}|p{7em}|}
        \hline 
        Range&$a_{1}\leq{1.1547r_{st}} $&$1.1547r_{st}\leq{a_{1}}\leq{1.2599r_{st}}$&$1.2599 r_{st}\leq{a_{1}}\leq{1.2961r_{st}}$ &$1.2961 r_{st} \leq{a_{1}}\leq{1.4142r_{st}}$\\ 
        \hline 
        $V_{C}$&${\color{red} a_{1}^{3}}$&${\color{red} {1.1547r_{st}}^{3}}$&$ {\color{orange} {1.1547r_{st}}^{3}}$&$ {\color{cyan} {1.1547r_{st}}^{3}}$\\ 
        \hline 
        $V_{HP}$&$ {\color{cyan} 0.7071{a_{1}}^{3}}$&$ {\color{cyan}0.7071{a_{1}}^{3}}$ &${\color{cyan}0.7071{a_{1}}^{3}}$&${\color{orange} 0.7071{a_{1}}^{3}}$ \\ 
        \hline 
        $V_{RD}$&$ {\color{cyan} 0.7071{a_{1}}^{3}}$&$ {\color{cyan}0.7071{a_{1}}^{3}}$ &${\color{cyan}0.7071{a_{1}}^{3}}$&${\color{orange} 0.7071{a_{1}}^{3}}$  \\  
        \hline 
        $V_{TO}$&${\color{orange} 0.7698{a_{1}}^{3}}$&${\color{orange}0.7698{a_{1}}^{3}}$&${\color{red}0.7698{a_{1}}^{3}}$&${\color{red}0.7698{a_{1}}^{3}}$\\ 
        \hline 
        Range&$1.4142r_{st}\leq{a_{1}}\leq{1.5431r_{st}}$&$1.5431r_{st}\leq{a_{1}}\leq{1.5492r_{st}}$&$ r_{st}\leq{1.5492a_{1}}$& \\ 
        \hline 
        $V_{C}$&${\color{cyan} {1.1547r_{st}}^{3}}    $&${\color{cyan}{1.1547r_{st}}^{3}}    $&${\color{cyan}{1.1547r_{st}}^{3}}$& \\ 
        \hline 
        $V_{HP}$&${\color{red} {1.4142r_{st}}^{3}}    $&$ {\color{orange} {1.4142r_{st}}^{3}}    $&$ {\color{orange} {1.4142r_{st}}^{3}}$& \\ 
        \hline 
        $V_{RD}$&${\color{red} {1.4142r_{st}}^{3}}    $&$ {\color{orange} {1.4142r_{st}}^{3}}    $&$ {\color{orange} {1.4142r_{st}}^{3}}$&  \\ 
        \hline 
        $V_{TO}$&${\color{orange}0.7698{a_{1}}^{3}}$&${\color{red}0.7698{a_{1}}^{3}}$&${\color{red}{1.5492r_{st}}^{3}}$&\\ 
        \hline 
    \end{tabular} 
    \caption[The area occupied by each cell]{The area occupied by each cell}
    \label{t4}
\end{table}
\subsection{The Choice for the Bluefin-21 Robot}
\begin{figure}
    \centering
    \includegraphics[width=1\linewidth]{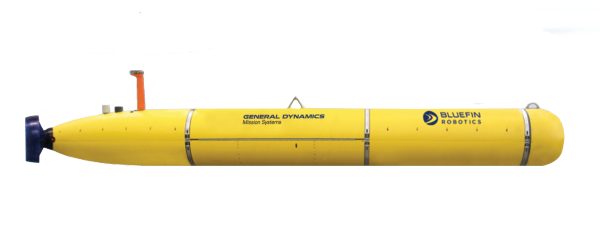}
    \caption[Bluefin-21]{Bluefin-21\cite{RN124}}
    \label{fig:17bluefin}
\end{figure}

Bluefin-21 is a \ac{UUV} (see Figure \ref{fig:17bluefin}) which carries multiple sensors and payload for tasks such as search, salvage and mine countermeasure \cite{RN124}. It is famous in searching MH370 which is also the inspiration of the research of this report. In the search task, the sensor for underwater locator beacon has a target sensing range as $ r_{st}=1$km at least in the normal condition \cite{RN123}. The sensor for obstacles is the EdgeTech 2200-M side scan sonar. It has different sensing ranges at different frequencies and resolutions. To guarantee the obstacle avoidance is successful, the minimum one $ r_{so}=75$m is used to ensure the accuracy. For the local communication, there is no information in the datasheet for the equipment used and for the range that it can communicate in so that the parameter of the 2D robot is used which is a Wi-Fi communication with $ r_{c}=91.4$m. In calculations, all the equality conditions of the inequalities are used. Thus, based on Inequality \ref{e2}, the maximum $a_{1}=min(75,91.4/2)=45.7$m. As $ a_{1}<1.1547r_{st}$, the first column of Table \ref{t4} is used and the C grid is chosen. Then $a=min(a_{1}, a_{2})=45.7$m and the corresponding assumptions for passage width $ W_{pass} $ is given.

\begin{figure}
    \centering
    \includegraphics[width=0.8\linewidth]{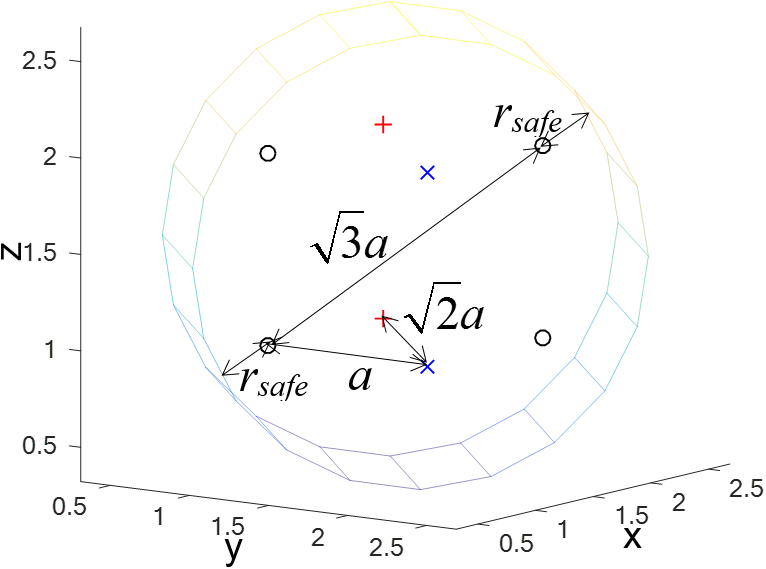}
    \caption[Passage width for a 2D area]{Passage width for a 2D area}
    \label{fig:18wpass3D}
\end{figure}

\begin{assumption}
    All passages between obstacles or between the obstacles and borders have a cross-section which can include a circle with a radius of $\sqrt{3}a+2r_{safe}$. This can be seen in Figure \ref{fig:18wpass3D}
\end{assumption} 
So the theoretical $ W_{pass} $ without $r_{safe}$ is $79.1547$m. This guarantees that there is always at least one accessible vertex in the cross-section which is illustrated in Figure \ref{fig:18wpass3D}. In the figure, the robot moving from a blue vertex ({\color{blue}{x}}) to a red vertex ({\color{red}{+}}) must go through a black vertex (o) as the robot can only move $a$ in a step. Although it is obvious that decreasing $a$ will expand the scope of applicability, the number of vertices will increase so the search time will also increase. 
\section{Summary}
In summary, this chapter proposed that a grid pattern should be used as the final location to deploy the \acl{WMSN} and the representation of the area for both search and complete coverage tasks. Three regular tessellations in a 2D field and four uniform honeycombs in a 3D area were considered. Then Table \ref{t2} for 2D and Table \ref{t4} for 3D are provided as the tables of selection criteria for different relations between the communication range, the target sensing range and the obstacle sensing range. Then typical robots which are Pioneer 3-DX robots in 2D and Bluefin-21 robot in 3D were used to exemplify the selection procedure. Based on the result, later simulation and experiment will mostly use the T grid in 2D tasks and the C grid in 3D tasks.  
\chapter{Optimal Collision-Free Self-Deployment for Complete Coverage}   
\chaptermark{Optimal Collision-Free Deployment for Complete Coverage}
Complete coverage is an attractive topic for multi-robot systems. It means each point of the task area is detected by at least one robot so the team can cover the area completely for surveillance or data collection. In this task, communication between the robots is usually required so each robot should be connected to at least one robot to form a network. Thus, any incident in the area or any intruder moving into the area will be known by the whole network to take further action. The mobile robotic sensor network in this chapter is an example of a networked control system with limited communication \cite{RN344,RN345,RN346,RN347}. As the area to cover is unknown or may be dangerous, robots could not be deployed manually and thus, need to be released from a known entrance to the area. While moving to the final location, the process should be as fast as possible, so the optimal algorithm to drive the robots while avoiding collisions is found in this chapter.

Self-deployment algorithm for 2D and 3D areas can be found in \cite{RN2} and \cite{RN138}. In their deployment phases, all robots applied the \ac{DR} algorithm at the same time which is impossible to cover the area completely because the algorithm could only work when there is a sequence of choice and robots inform the selection to neighbor robots which, however, was not mentioned in \cite{RN2,RN138}. Thus, the absorbing state in their proof may not be reached, and the algorithm may be invalid. The DR algorithm has a disadvantage that robots need to convert received coordinates to their coordinate system which leads to a large calculation load. To compare to the DR algorithm, robots in this chapter also start from certain vertices of a shared grid in some possible ways. Then, an optimal navigation algorithm is designed to have the fewest steps for coverage.

This chapter presents a novel distributed algorithm for deploying multiple robots in an unknown area to achieve complete coverage. The task area can be \acl{2D} or \acl{3D} and it is bounded and arbitrary with obstacles and certain assumptions. The robots with limited sensing ranges and a limited communication range move in sequence via grid patterns which were chosen in Chapter 2 to reach the complete coverage. The algorithm sets a reference point and drives the furthest robot first to move to one of the most distant neighbor vertices in each step with no collisions. Finally, every point of the area can be detected by at least one robot and robots achieve this using the fewest steps. If the task area is enlarged later, the algorithm can be modified easily to fit this requirement. The convergence of the algorithm is proved with probability one. Simulations are used to compare the algorithm with valid modification of the DR algorithm in different areas to show the effectiveness and scalability of the algorithm.

The rest of the chapter will be divided into two sections for the 2D problem and the 3D problem respectively. Each section will start with the problem statement followed by the description of the algorithm. Then MATLAB simulations and the comparison with another algorithm are shown, and finally, a summary will be provided.

\section{Completer Coverage in 2D}
This section provides a decentralized collision-free algorithm to deploy multiple robotic sensors in a task area $ A $ with obstacles to achieve the 100\% complete coverage using a T grid pattern based on Chapter 2. Thus, intruders entering the covered area can be detected. This section takes advantage of the methods for the definition and assumption in \cite{RN1,RN133,RN155}. Some assumptions and definition about the area, ranges, and the curvature under the strict requirement have been claimed in Section \ref{s1} so they are used directly without explaining. The original work of this section comes from \cite{RN133}.
\subsection{Problem Statement}
\label{s3}
\begin{assumption}
    There are a finite number of obstacles $ O_{1} $, $ O_{2} $,$\ldots $, $O_{l} $ and obstacles $ O_{i} $ are non-overlapping, closed, bounded and linearly connected for any $i>0$. The obstacles are static, arbitrary, and unknown to the robots a priori. \label{a14}
\end{assumption}
\begin{definition}
    Let $O:=\bigcup{O_{i}}$ for all $i>0$. Then $A_{c}:=A\setminus{O}$ represents the area that needs to be covered. Area $A_{c}$ can be static or be expanding.\label{d3}
\end{definition}  
In each move, a robot translates straightly from a vertex represented as $v_{i}$ of the grid to the selected vertices $v_{j}$ which is one of the nearest neighbors. Thus, the moved distance is the side length $a$ of the basic grid pattern. In $A_{c}$, the set of accessible vertices is denoted by $V_{a}$. If $\lvert\bullet\rvert$ is used to represent the number of elements in set $\bullet$, $\lvert V_{a} \rvert$ will represent the number of accessible vertices. $m$ robots with the same sensing and movement ability are released from $N_{i}$ initial known vertices to cover $A_{c}$. Then for the optimal deployment, let $N_{o}$ represent the number of steps needed. For the complete coverage, the number of robots prepared needs to satisfy
\begin{assumption}
     $m\geq{\lvert V_{a} \rvert}$. \label{a15}
\end{assumption}

The set containing sensing neighbors at discrete time $k$ is denoted as $N_{s,i}(k)$ and its $j$th element is at position $n_{s,i,j}(k)$. $N_{c,i}(k)$ denotes the set containing those communication neighbors at time $k$. As all robots are equal, the ad-hoc network is used for communication between $rob_{i}$ and robots in $N_{c,i}(k)$. These local networks are temporary and need to be rebuilt when robots arrive at new vertices. In the future tests on robots, Wi-Fi networks in ad-hoc mode will be used with \ac{TCP} socket communication. 

At time $k$, the chosen position of $rob_{i}$ is $p_{i}(k)$. A reference position $p_{r}$ is selected from the initial positions. Some definitions about distances are provided below.
\begin{definition}
    The distance from $p_{i}(k)$ to $p_{r}$ is denoted as $d(p_{i}(k),p_{r})$. Similarly, the distance from an accessible neighbor vertex $n_{s,i,j}(k)$ to $p_{r}$ is $d(n_{s,i,j}(k), p_{r})$ and the maximum distance for all $n_{s,i,j}(k)$ is denoted as $max(d(N_{s,i}(k), p_{r}))$ with $N_{max}$ sensing neighbors with this value. A random choice from those $N_{max}$ values is represented by $c$.\label{d4}
\end{definition}
\begin{assumption}
    $W_{pass}\geq{a+2*r_{safe}}. $\label{a16}
\end{assumption}

\subsection{Algorithm}
\label{s8ch3algo}
The algorithm includes two parts. The first part is to allocate robots to a common T grid with a common coordinate system. In the second part, a decentralized collision-free algorithm is designed to deploy robots using the grid.  
\subsubsection{Initial Allocation}
\begin{figure}
    \centering
    \includegraphics[width=0.5\linewidth]{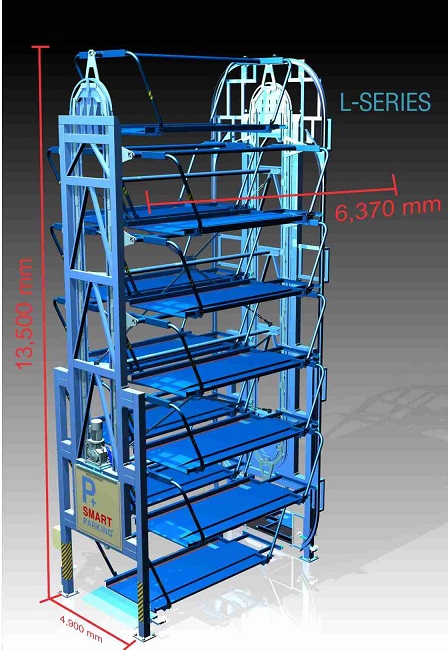}
    \caption[The multilevel rotary parking system]{The multilevel rotary parking system}
    \label{fig:19Vcarpark2}
\end{figure}

\begin{figure}
    \centering
    \includegraphics[width=1\linewidth]{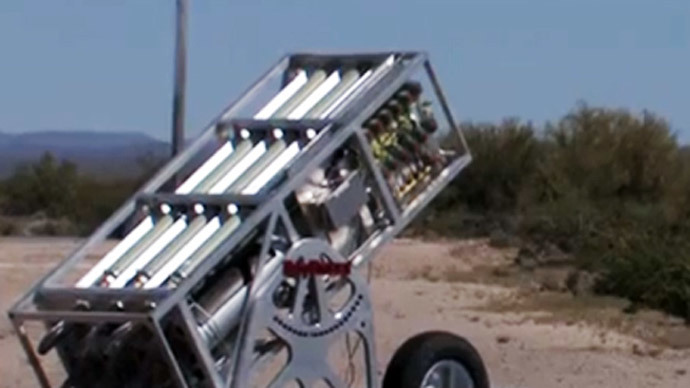}
    \caption[The launcher for multiple UAVs in the LOCUST project]{The launcher for multiple UAVs in the LOCUST project}
    \label{fig:20launcher}
\end{figure}

Before deployment, \cite{RN138,RN155} could only build a common grid with small errors in certain situations and one vertex can have more than one robot which means collisions are allowed. To compare to that algorithm and design a practical initialization with a common coordinate system, this chapter manually selects a few vertices of a grid near the known entrance of the area as initial positions to allocate robots and uses one point from them as $p_{r}$. Initial positions and $p_{r}$ are set to the place where robots could have more choices in the first step based on the limited knowledge of the area after applying the algorithm so that robots could leave the initial vertices earlier without blocking each other. To allocate multiple robots at the same point in different steps, the multi-level rotary parking system as in Figure \ref{fig:19Vcarpark2} from \cite{RN310} could be used. Thus, if a robot moves away from a vertex, the empty parking place will rotate up, and another robot will be rotated down to the same start position. Another possible method is to release robots using the rocket launcher such as the tubes employed in the LOCUST project for the Coyote unmanned aircraft system \cite{RN130} in Figure \ref{fig:20launcher}. As the relative positions of the tubes are fixed, the position of the robot in one tube can be set as the reference to allow all robots to have the same coordinate system and the same grid.
\subsubsection{Distributed Algorithm}
\begin{figure}
    \centering
    \includegraphics[width=1\linewidth]{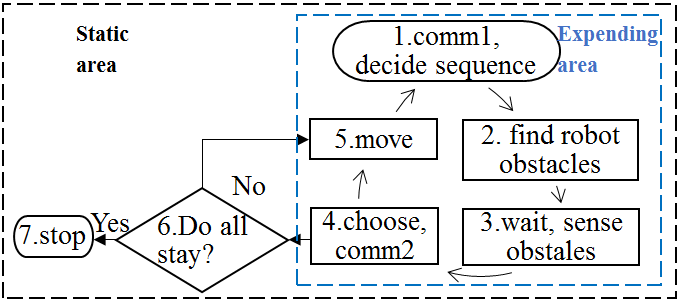}
    \caption[The flow chart for the 2D complete coverage]{The flow chart for the 2D complete coverage}
    \label{fig:21flow2Dcover}
\end{figure}

The distributed algorithm for deployment is designed based on the search algorithm in \cite{RN134,RN133,RN132,RN147}. The flow chart of the algorithm is in Figure \ref{fig:21flow2Dcover} and the abbreviation \textquoteleft{comm}\textquoteright{} in that figure is short for communication. Step 6 and Step 7 only happen when the area is static. If the area will be extended, the algorithm only includes the loop with Step 1 to Step 5. In both situations, robots will sense intruders continuously and broadcast the positions of intruders when they are found. However, for the static area, intruder detection may only be required after the deployment phase is stopped in Step 7 in some applications. 

\begin{enumerate}[fullwidth,itemindent=2em]
\item \textbf{Comm1, decide sequence:} A robot calculates $d(p_{i}(k),p_{r})$ at the beginning of each loop and exchanges $d(p_{i}(k),p_{r})$ and $p_{i}(k)$ with its communication neighbors. Then, the one with larger $d(p_{i}(k),p_{r})$ has higher priority. If more than one robot have the same $d(p_{i}(k),p_{r})$, another parameter of robots needs to be used to decide the selection sequence. As only addresses, such as IP addresses, are different between robots, this chapter allows the robot with a lower IP address to choose with a higher priority, however, if the one with higher address chooses first, the result should be the same. 
\item \textbf{Find robot obstacles:} The positions of robots are known, so robots within $r_{so}$ of $rob_{i}$ are thought as obstacles, and the occupied vertices are inaccessible for $rob_{i}$. The $N_{i}$ vertices to release robots are always considered as obstacles so robots will not choose them as the next step, and new robots can be added to the area.
\item \textbf{Wait, sense obstacles:} A robot needs to wait for the choices from the robots with higher priorities because the positions of those choices can be thought as detected points in the next step and may also be obstacles for $rob_{i}$. If all robots are connected, the last one in the choosing sequence needs to wait for every other robot and the processing time for the whole selection process should be set to the value in this condition because this is the longest time and using this time, all the robots could be synchronized in each loop. 

\setlength{\parindent}{2em}In the turn for $rob_{i}$, it detects the objects in $r_{so}$ but avoids the section occupied by existing neighbor robots. The way to judge whether the object $p$ is an obstacle or not is illustrated in Figure \ref{fig:22judgeobs}. The general rule is that if there is any sensed object on the way to vertex $n_{s,i,j}(k)$ in $N_{s,i}(k)$, that object is treated as an obstacle. So that vertex $n_{s,i,j}(k)$ cannot be reached and is removed from $N_{s,i}(k)$. In Figure \ref{fig:22judgeobs}, $r_{new}$ is the distance between $p_{i}(k)$ and the left arc where $r_{new}=\sqrt{(a^{2}+r_{safe}^{2})}$. $d_{pr}$ represents the distance from the sensed object $p$ to the position of $rob_{i}$, $p_{i}(k)$. $d_{pn}$ denotes the distance between $p$ and $n_{s,i,j}(k)$ and $d_{pe}$ denotes the distance from $p$ to the nearest edge of the grid. Let the angle between the facing direction of $rob_{i}$ and the segment which connects $p_{i}(k)$ and $p$ be $\theta_{pr}$. Due to the safe distance $r_{safe}$, when a robot goes straight to $n_{s,i,j}(k)$, it needs a related safe region which is the light brown part in the figure. When an object is detected, $d_{pr}$ and $\theta_{pr}$ are known so coordinates of that object can be calculated. Thus, $d_{pn}$ and $d_{pe}$ can be determined. If $(d_{pr}<r_{new}\&d_{pe}<r_{safe})|d_{pn}<r_{safe}$, point $p$ is an obstacle. If no objects are sensed, this robot will still wait for the same amount of processing time so all robots are synchronized in this step.
\begin{figure}
    \centering
    \includegraphics[width=1\linewidth]{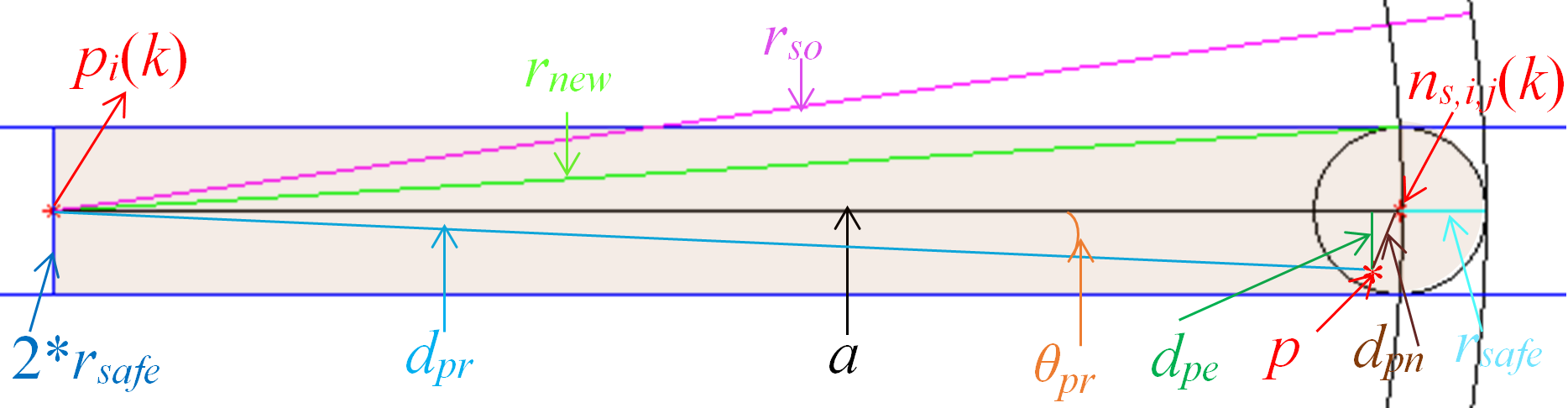}
    \caption[Judge obstacles]{Judge obstacles}
    \label{fig:22judgeobs}
\end{figure}
\item \textbf{Choose comm2:} $rob_{i}$ calculates $max(d(N_{s,i}(k), p_{r}))$ for its sensing neighbors and randomly chooses one vertex from $N_{max}$ neighbor vertices which could result in $max(d(N_{s,i}(k), p_{r}))$. If there are no accessible vertices to visit for $rob_{i}$, namely $N_{max}=0$, $rob_{i}$ will stay at its current position. Mathematically, it can be written as:
\begin{equation}
p_{i}(k+1)=
\begin{cases}
c \text{ with prob. } 1/N_{max}, \text{ if } N_{max}\neq{0},\\
p_{i}(k),  \text{ if } N_{max}={0}.
\end{cases}
\label{e3}
\end{equation}

Then, the choice will be sent to robots with lower priorities. For the static area, a robot also records and sends the IP addresses of robots which will move in the current step. After communication, the robot still waits until the time for choice past. However, robots have no knowledge about the number of robots used at time $k$. Thus, the total time for making a choice is equal to the time for selection and communication of one robot multiples the total number of robots prepared for deployment. In this way, robots can be synchronized.
\item \textbf{Move:} All robots move together and those who do not move need to wait for the time for the movement to get synchronized. The movement includes rotating the head of the robot to aim at the target vertex and translating to the vertex. All robots should move in the same pattern at the same time, and the way to move should match the realistic constraints. 

Each movement step consists of two phases: phase I is the pure rotation, phase II is the pure translation. The rotation will align the robot with the target vertex. Specially, if $rob_{i}$ was facing to its next vertex before phase I, it will not move but wait for the rotation time. Then $rob_{i}$ can go straight along the selected edge. Many robots use the nonholonomic model with standard hard constraints on velocity and acceleration. This report also considers the constraints of real robots so that both phases use trapezoidal velocity profiles based on parameters of the area and the test robots. In this section for 2D areas, Pioneer 3-DX robots are used, and parameters are shown in simulations (Section \ref{s4}).
\item \textbf{Do all Stay? :} Based on the sequence of choice, the robot at $p_{r}$ will be the last to choose the next vertex. If no IP addresses from robots which go to new vertices are received, and the choice of the robot at $p_{r}$ is also staying at the current position, namely $p_{i}(k+1)=p_{i}(k)$ for all robots, then the self-deployment should be stopped. Otherwise, all robots execute Step 5.
\item \textbf{Stop:} The stop procedure depends on the applications. In this chapter, robots are designed to detect intruders into the area so the robot at $p_{r}$ could broadcast the stop signal, and all robots stop moving and obstacles sensing but only keep the sensors for target intruders on. If an intruder is found, robots broadcast the position of the intruder, and other corresponding actions could be taken. Specially, under the strict requirement, if the sensing range for the target is changeable, the sensors which cannot sense the boundaries could decrease their sensing range to the minimum value which can meet the loose requirement, namely, $ r_{st} =a_{T}/\sqrt{3}+r_{safe}-r_{rob}$. Thus, the energy of those sensors will be saved, and they can monitor the area for a longer time.
\end{enumerate}
\begin{remark}
     The broadcast is only used for a few vital information including the stop signal and positions of intruders, so it will not consume too much energy. If only local communication could be used, the stop signal can still be transmitted to all robots at the time slot for comm1 and receivers will do all the actions in Step 7 directly except broadcasting the stop signal.
\end{remark}
By applying the above algorithm, robots which are further away from $p_{r}$ have higher priorities to choose to go to the furthest accessible vertices, which leaves spaces for the newly added robots near $p_{r}$ to go to. If there are enough available choices for robots, $ N_{i}$ robots could be added in each loop and an optimal deployment with $N_{o}$ steps can be reached. Thus, the equation for $N_{o}$ can be written as:
\begin{equation}
N_{o}=ceil(\lvert V_{a} \rvert/N_{i}). \label{e4}
\end{equation}
In the above formula, the $ceil()$ is the ceiling function because the result of the division may be a decimal figure. However, less than one step means not all the robots need to move in that step, but that should still be finished in one step.
\begin{theorem}
    \label{th2Dc}
    For a static area, suppose all assumptions in Section \ref{s1} and \ref{s3} hold for the loose requirement or the strict requirement, and algorithm \ref{e3} with the judgment step is used. Then there exists such a time $ k_{0}>0$ with probability 1 that for any vertex $v_{j}\in{V_{a}} \text{ where } j=1,2,\ldots,\lvert V_{a} \rvert$, the relationship $v_{j}=p_{i}(k)$ holds for some $i=1,2,\ldots,m$  and all $k\geq{k_{0}}$.
\end{theorem}
\begin{prove}
The proposed Algorithm \ref{e3} and the judgment method in Step 6 form an absorbing Markov chain with many transient states and absorbing states. The transient states are the states that robots have accessible sensing neighbors to choose. The absorbing states are the states that all robots have no sensing neighbors to go and stop at their current vertices forever, namely the states that they cannot leave. The proposed algorithm continuously adds robots to the area to occupy every vertex, which enables robots to reach the absorbing states from any initial states with any $N_{i}$. This implies the absorbing state will be achieved with probability 1. This completes the proof of Theorem \ref{th2Dc}.
\end{prove}
\subsection{Simulation Results}
\label{s4}
\begin{figure}
    \centering
    \includegraphics[width=0.8\linewidth]{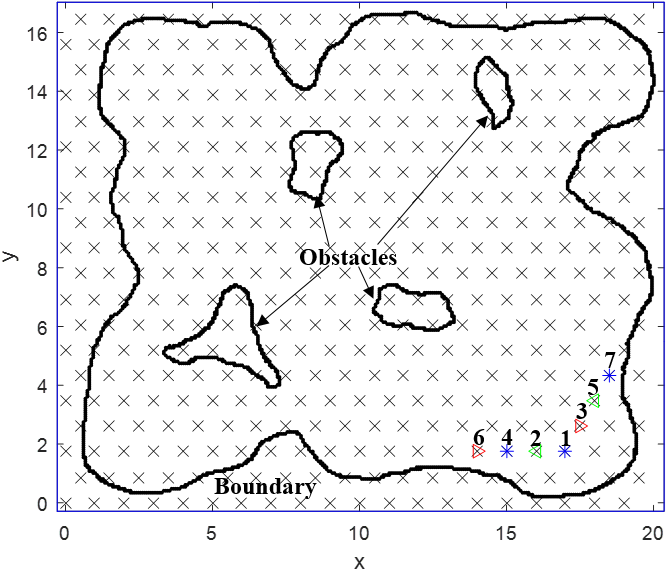}
    \caption[The task area and the initial setting]{The task area and the initial setting}
    \label{fig:23simucover2D}
\end{figure}

The proposed algorithm for the complete coverage is simulated in MATLAB2016a in an area with 20*20 vertices and $\lvert V_{a} \rvert=256$ as seen in Figure \ref{fig:23simucover2D}. $N_{i}=3, 5, 7$ are chosen to verify the results and those known initial vertices are at the bottom right corner to release robots. The numbers in the graph are the order to add vertices, and vertex 1 is the reference vertex at $p_{r}$. In this arrangement, each robot will have vacant vertices to go without blocking each other in the first loop. 

The parameters in the simulation are designed based on parameters of Pioneer 3-DX robots in Figure \ref{fig:10robot} and the assumptions satisfying the strict requirement. Table \ref{t5} shows the parameters in the simulation which are designed based on the datasheet of the pioneer 3-DX robot. Row P, R, and S are the names of the parameters, parameters for robots and parameters in simulations respectively. The underlined column names refer to the known values for robots. $r_{safe}$ and $t_{move}$ need to be estimated from the experiment. Then $a$ and $W_{pass}$ can be calculated in turn based on assumptions in Section \ref{s1} and $r_{safe}$. In calculations, equality conditions of the inequalities are used. In the experiment, robots will have at most a 0.08m shift to the left or right after moving 8 meters forward or backward, so $e$ is set to 0.09m and $r_{safe}=0.35$m. In an experiment with $a=1$m and $v_{max}=0.3$m/s, moving time is roughly estimated as 5.75 seconds which is 4.5s for the translation and $1.25s$ for the rotation using trapezoidal velocity profile and the maximum rotation speed $\omega_{max}=\SI{300}{\degree}$/s of the robot. In simulations, $r_{rob}$ and $\omega_{max} $ from the robot are used. $a$, $v_{max}$ and $t_{move}$ from the experiment are used. Although different areas with different sizes will be simulated, for simplicity, the error in the simulation is always set as $e=0.09$ so $r_{safe}=0.35$. Then other values are calculated based on assumptions in Section \ref{s1}.  
\begin{table}
    \centering
    \renewcommand\arraystretch{2}
    \begin{tabular}{|c|c|c|c|c|c|}
        \hline 
        P&$    \underline{r_{rob}}$&$    r_{safe}$&$    \underline{r_{so}}$&$    \underline{r_{st}}$&$    a$  \\ 
        \hline 
        R&    0.26m&0.35m&5m&32m    &4.65m\\ 
        \hline 
        S&0.26&0.35&1.35&1.35&1 \\ 
        \hline 
        P&$    \underline{r_{c}}$&$    \underline{v_{max}}$& $    \underline{\omega_{max}}$&$    t_{move}$&$    W_{pass}$\\
        \hline
        R&91.4m&1.5m/s&$\SI{300}{\degree}$/s&5.75s&5.35m\\
        \hline
        S&2.09&0.3/s&300&5.75s&1.7\\
        \hline
    \end{tabular} 
    \caption[Parameters of robots and simulation]{Parameters of robots and simulation}
    \label{t5}
\end{table}

\begin{figure}
    \centering
    \includegraphics[width=0.8\linewidth]{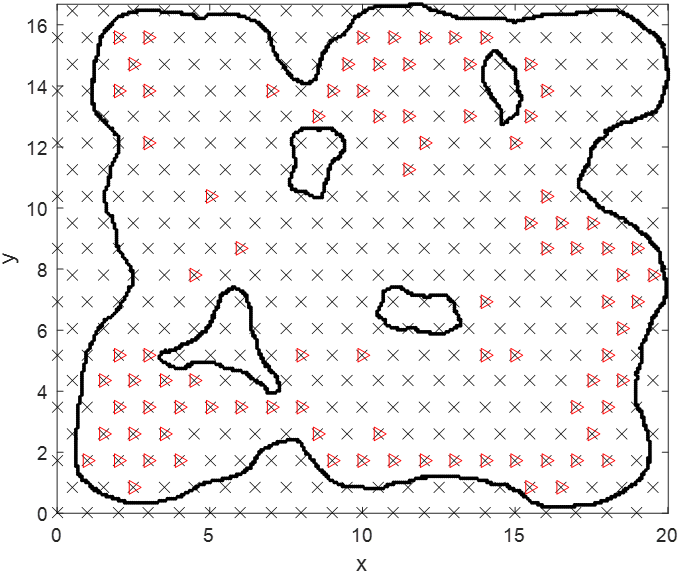}
    \caption[Positions of robots when $k=30, \lvert V_{a} \rvert=256, N_{i}=3$]{Positions of robots when $k=30, \lvert V_{a} \rvert=256, N_{i}=3$}
    \label{fig:24robk30}
\end{figure}

\begin{figure}
    \centering
    \includegraphics[width=0.8\linewidth]{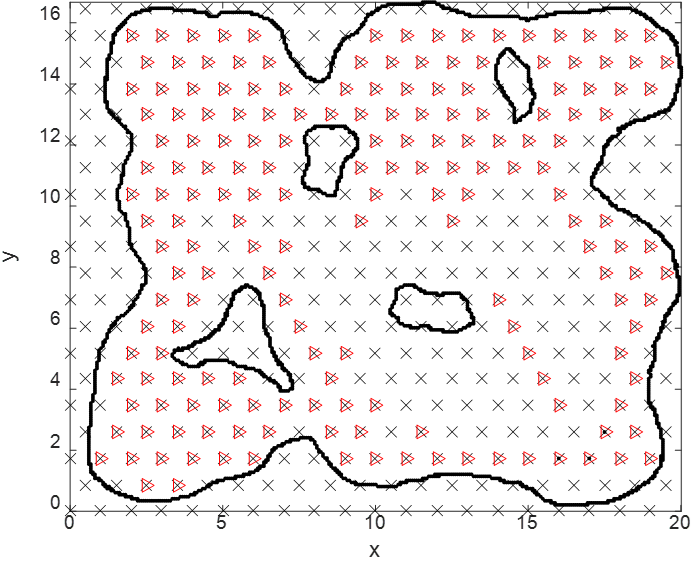}
    \caption[Positions of robots when $k=60, \lvert V_{a} \rvert=256, N_{i}=3$]{Positions of robots when $k=60, \lvert V_{a} \rvert=256, N_{i}=3$}
    \label{fig:25robk60}
\end{figure}

\begin{figure}
    \centering
    \includegraphics[width=1\linewidth]{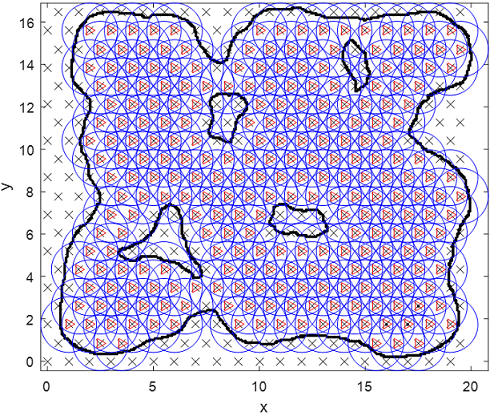}
    \caption[Positions of robots when $k=86, \lvert V_{a} \rvert=256, N_{i}=3$]{Positions of robots when $k=86, \lvert V_{a} \rvert=256, N_{i}=3$}
    \label{fig:26robk86}
\end{figure}

An example progress for $N_{i}=3$ is shown in Figure \ref{fig:24robk30}, \ref{fig:25robk60} and \ref{fig:26robk86} for $k=30$, $k=60$ and $k=86$ respectively and the self-deployment finished in 86 loops. The graph shows that robots tended to go to vertices which were far from $p_{r}$ and left space for newly added robots, which made the deployment fast. In Figure \ref{fig:26robk86}, the circles represent the $r_{st}$ of each robot and every point of the task area $A_{c}$ was covered at the end.

The proposed algorithm is compared to the modified version of the DR algorithm in \cite{RN2,RN119}. To modify it, firstly, the IP addresses of the robots are used as the order numbers to set the sequence of choice. Therefore, for $rob_{i}$ and its communication neighbors, the robot with the lowest address will choose first. Then the robot needs to inform its communication neighbors with lower priorities about its choice in this loop. 
\begin{figure}
    \centering
    \includegraphics[width=0.5\linewidth]{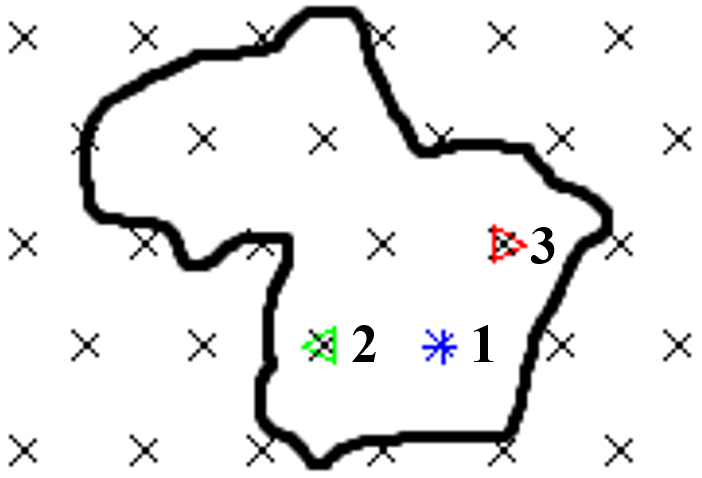}
    \caption[An example of the narrow path with $k=3, \lvert V_{a} \rvert=6, N_{i}=3$]{An example of the narrow path with $k=3, \lvert V_{a} \rvert=6, N_{i}=3$}
    \label{fig:27narrowexample}
\end{figure}

Table \ref{t6No20} illustrates the $N_{o}$ calculated using formula \ref{e4}, average results of 100 simulations for the proposed algorithm and average results of 100 simulations for the DR algorithm. It also shows the ratio of the steps in the proposed algorithm to that in the DR algorithm. It can be seen that the proposed algorithm is always better than the DR algorithm, and the advantage is clearer when $N_{i}$ is larger. Compared to $N_{o}$, the simulation result of the proposed algorithm is equal to it when $N_{i}=3$. However, for $N_{i}=5$ or 7, the simulation results are a little bit larger but no greater than 1 step. This result means some loops had less than $N_{i}$ robots added because less than $N_{i}$ robots found vacant vertices to go to. This can be caused by some narrow vacant areas which can be seen in the example in Figure \ref{fig:27narrowexample} where $N_{i}=3$ and $\lvert V_{a} \rvert=6$. From formula \ref{e4}, $N_{o}=2$, however, the narrow path only allows one robot to be added in each loop. So, robots need four loops to finish the complete coverage. Another reason is that $rob_{i}$ can choose a $n_{s,i,j}(k)$ whose $d(n_{s,i,j}(k), p_{r})$ is smaller than $d(p_{i}(k),p_{r})$ but other robots within $r_{so}$ of $rob_{i}$ have higher priorities so they have finished choosing the next step. Thus, $p_{i}(k)$ would be vacant in step $k+1$ and the movement of $rob_{i}$ is towards the reference point. So it does not result in a vacant vertex near initial vertices for adding a new robot.
\begin{table}
    \centering
    \renewcommand\arraystretch{2}
    \begin{tabular}{|c|c|c|c|}
        \hline 
        \multirow{2}{*}{Type of Results}&\multicolumn{3}{c|}{$N_{i}$} \\ 
        \cline{2-4} 
        & 3 & 5 &7  \\ 
        \hline 
        $N_{o}$&86&52&37\\
        \hline
        Proposed algorithm&86&52.06&37.96\\ 
        \hline 
        DR algorithm&310.39&273.74&254.92\\ 
        \hline 
        Proposed/DR&27.7\%&19.0\%&14.9\%\\ 
        \hline 
    \end{tabular} 
    \caption[Steps from calculations and simulations with $\lvert V_{a} \rvert=256$]{Steps from calculations and simulations with $\lvert V_{a} \rvert=256$}
    \label{t6No20}
\end{table}

To test the scalability of the algorithm, areas with the same shape but different sizes are used with $N_{i}=3$. So areas with 10*10 vertices and 30*30 vertices are used with $\lvert V_{a} \rvert=57$ and $\lvert V_{a} \rvert=598$ respectively. The average results of the two algorithms, the ratio between them, and $N_{o}$ are shown in Table \ref{t7No102030}. The results demonstrate that the proposed algorithm is scalable and effective. Its advantage is clearer when the area is larger. The steps for the areas with 20*20 and 30*30 vertices is optimal, but the steps for the area with 10*10 vertices is larger than $N_{o}$ because the shrunk area has more narrow paths as discussed in Figure \ref{fig:27narrowexample}.

\begin{table}
    \centering
    \renewcommand\arraystretch{2}
    \begin{tabular}{|c|c|c|c|}
        \hline 
        \multirow{2}{*}{Type of Results}&\multicolumn{3}{c|}{Size of the Area} \\ 
        \cline{2-4} 
        & 10*10&    20*20&    30*30  \\ 
        \hline 
        $N_{o}$&    19    &86&    200\\
        \hline
        Proposed algorithm&23.96&    86&    200\\ 
        \hline 
        DR algorithm&70.62    &310.39    &766.71\\ 
        \hline 
        Proposed/DR&    36.0\%&    27.7\%&    26.1\%\\ 
        \hline 
    \end{tabular} 
    \caption[Steps from calculations and simulations for different areas with $N_{i}=3$]{Steps from calculations and simulations for different areas with $N_{i}=3$}
    \label{t7No102030}
\end{table}
\subsection{Section Summary}
This section provided a decentralized algorithm for multiple robots to build a robotic sensor network to completely cover an unknown task area without collisions using an equilateral triangular grid pattern. The convergence of the proposed algorithm was mathamatically proved with probability 1. Results of this algorithm were very close to the least number of steps and were equal to that in some situations. So the algorithm is optimal under the limitation of the area. Simulation results demonstrated that the proposed algorithm is effective comparing to another decentralized algorithm and it is scalable for different sizes of areas.

\section{Completer Coverage in 3D}
This section solves the complete coverage problem in a 3D area by using a decentralized collision-free algorithm for robots to form a coverage network. Based on Chapter 2, the C grid is used in this section for the Bluefin-21 robot. The assumptions and definition for the area and ranges from Section \ref{s2} are used directly. Others are similar as Section \ref{s3} except the ranges are spherical, and the shape of objects are \acl{3D} which are based on \cite{RN171}. However, these assumptions and definitions are still claimed here to make the explanation and the proof clear and will be used in later Chapters. 
\subsection{Problem Statement}
\begin{assumption}
    Area A has a finite number of \acl{3D} obstacles $ O_{1} $, $ O_{2} $,$\ldots $, $O_{l} $ and all these obstacles $ O_{i} $ are non-overlapping, closed, bounded and linearly connected for any $i>0$. The obstacles are unknown to the robots. \label{a17obs}
\end{assumption}
\begin{definition}
    Let $O:=\bigcup{O_{i}}$ for all $i>0$. Then the 3D area that needs to be covered can be written as $A_{c}:=A\setminus{O}$. Area $A_{c}$ can be either static or be expanding.\label{d5}
\end{definition}

In each step, robots move $a $ from one vertex $v_{i}$ to an accessible nearest neighbor vertex $v_{j}$ through the edge of the grid. The number of steps needed in the optimal situation for the complete coverage is denoted as $N_{o}$. The set of accessible vertices in $A_{c}$  is denoted by $V_{a}$ and the number of members in $V_{a}$ is $\lvert V_{a} \rvert$. 
\begin{assumption}
    To occupy every vertex in $A_{c}$, $m\geq{\lvert V_{a} \rvert}$. Robots are fed into the area from $N_{i}$ initial vertices near the boundary.\label{a18Ac} 
\end{assumption}

The set of sensing neighbors of $rob_{i}$ at discrete time $k$ (step $k$) is denoted as $N_{s,i}(k)$ in which the position of the $j$th member is $n_{s,i,j}(k)$. 
The set of communication neighbors of $rob_{i}$ in time $k$ is denoted as $N_{c,i}(k)$. The communication in 3D is also employing ad-hoc network which is rebuilt after each robot reached its next vertex. 

The position of $rob_{i}$ at time $k$ is $p_{i}(k)$. A reference position $p_{r}$ which belongs to the $N_{i}$ initial positions is set. Then some definitions about distances are given. 
\begin{definition}
    The distance from $p_{i}(k)$ to $p_{r}$ is represented by $d(p_{i}(k),p_{r})$ and the distance from $n_{s,i,j}(k)$ to $p_{r}$ is $d(n_{s,i,j}(k), p_{r})$. The maximum $d(n_{s,i,j}(k), p_{r})$ is denoted by $max(d(N_{s,i}(k), p_{r}))$ with $N_{max}$ neighbor vertices resulting to this value. The position of the choice from those $N_{max}$ values is represented by $c$.\label{d6}
\end{definition}

To allow robots to go through each passage to reach each vertex, the assumption for passage width $W_{pass}$ is needed.
\begin{assumption}
    $W_{pass}\geq{ a\sqrt{3}+2r_{safe}}$.\label{a193Dwpass}
\end{assumption}

\subsection{Algorithm}
The algorithm starts with the method to set the initial locations of robots followed by the decentralized self-deployment algorithm with collision avoidance to form the complete coverage network to detect intruders.  
\subsubsection{Initial Allocation}
 The initialization in a 3D area is similar to that in a 2D area. $N_{i}$ nearby vertices of a grid near the known entrance of the area are selected as the initial positions to allocate robots, and one of them is set as $p_{r}$. The chosen points should allow robots to leave the initial positions without blocking each other and it is better to give a robot more choices. In practice, method in Figure \ref{fig:20launcher} and \ref{fig:19Vcarpark2} can be used. For the marine applications, $N_{i}$ ships or submarines could be used to carry robots and stop at the selected places to launch UUVs.
\subsubsection{Distributed Algorithm}
The 3D algorithm is the same as that in a 2D area, so the flowchart in Figure \ref{fig:21flow2Dcover} is still available. So Algorithm \ref{e3} and Equation \ref{e4} are still used here. 
\begin{equation}
p_{i}(k+1)=
\begin{cases}
c \text{ with prob. } 1/N_{max}, \text{ if } N_{max}\neq{0},\\
p_{i}(k),  \text{ if } N_{max}={0}.
\end{cases}
\label{e33D}
\end{equation}
\begin{equation}
N_{o}=ceil(\lvert V_{a} \rvert/N_{i}) \label{e43D}.
\end{equation}
To apply them to the 3D area, the user needs to think all parameter in the 3D situation such as the distance in obstacle detection and setting the priority. Then the algorithm in 2D will handle the task in 3D successfully. 
\subsection{Simulation Results}
\begin{figure}
    \centering
    \includegraphics[width=0.8\linewidth]{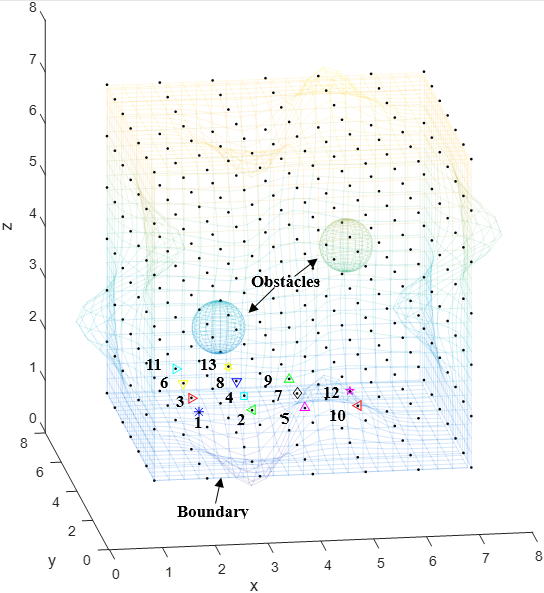}
    \caption[The task area and the initial setting]{The task area and the initial setting}
    \label{fig:283Dcoverarea}
\end{figure}

The proposed algorithm is initially simulated in MATLAB2016a for a static area with 7*7*7 vertices and $\lvert V_{a} \rvert=129$ in Figure \ref{fig:283Dcoverarea}. In the simulation, $N_{i}=3$, 5, 7, 9, 11 and 13 are used with initial vertices at the bottom layer of the task area. In the figure, numbers indicate the order to added vertices and the position of vertex 1 is selected as $p_{r}$. In this setting, all robots have accessible sensing neighbors to go in loop one. 

The parameters in the simulation are based on the Bluefin-21 underwater vehicle which was used in searching MH370 and details can be found in \cite{RN124}. The data for Bluefin-21 and simulation is shown in Table \ref{t8} where P, R, and S stands for parameter names, parameters for robots and parameters for simulations. The underlined parameter names mean those parameters for robots are from the datasheet. As $\omega_{max}$ is unknown in the datasheet, the $t_{move}$ cannot be estimated. The length of the robot is 4.93m, and the diameter is 0.53m, so $r_{rob}=4.93$m is set. $r_{safe}$ is calculated based on the $e$. In the datasheet, real-time accuracy $\leq0.1\%$ of the distance traveled. So $r_{safe}$ can be set using the distance traveled in simulations. Then $a$ and $W_{pass}$ are calculated based on $r_{safe}$ and assumptions in Section \ref{s2}. In simulations, $r_{rob}$ use the data from robot directly. Then $a=15$ is set to avoid collisions while robots are moving. To find $r_{safe}$ of the robot, the maximum travel distance in simulations in this section is used which is 155.885 as the body diagonal of the area with 8*8*8 vertices in comparison. So the corresponding $e=0.156$ and $r_{safe}=5.086$. Then other parameters in simulations can be calculated. Then $r_{safe}=5.086$m is set in the parameters for the robot. The rotation speed of this UUV is unknown, and there are no Bluefin-21 robots to test in the author's lab. Thus, the simulation results will not be compared to the experiment result so guess an angular speed and a total move time is meaningless. So in simulations, $t_{move}$ is still set as 5.75s with 4.5s for translation and 1.25s for rotation to show the effect of the algorithm, although it will be smaller than the actual value. 

\begin{figure}
    \centering
    \includegraphics[width=0.7\linewidth]{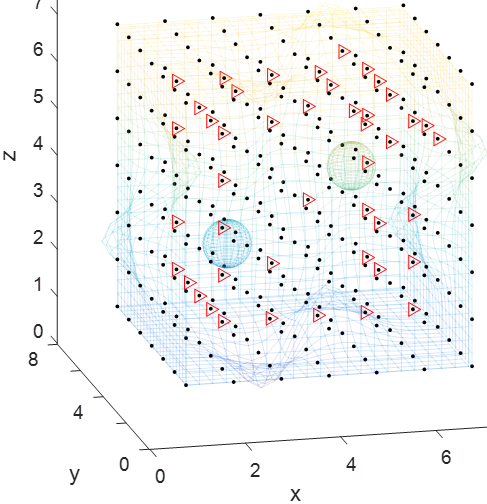}
    \caption[Positions of robots when $k=15$, ${\lvert V_{a} \rvert}=129$, $N_{i}=3$]{Positions of robots when $k=15$, ${\lvert V_{a} \rvert}=129$, $N_{i}=3$}
    \label{fig:293Dk15}
\end{figure}

\begin{figure}
    \centering
    \includegraphics[width=0.7\linewidth]{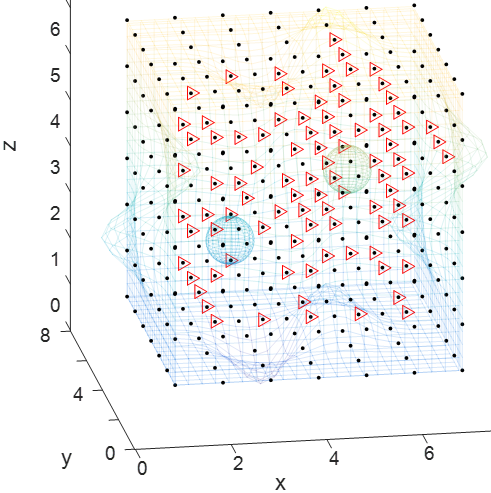}
    \caption[Positions of robots when $k=30$, ${\lvert V_{a} \rvert}=129$, $N_{i}=3$]{Positions of robots when $k=30$, ${\lvert V_{a} \rvert}=129$, $N_{i}=3$}
    \label{fig:30k30}
\end{figure}

\begin{figure}
    \centering
    \includegraphics[width=1\linewidth]{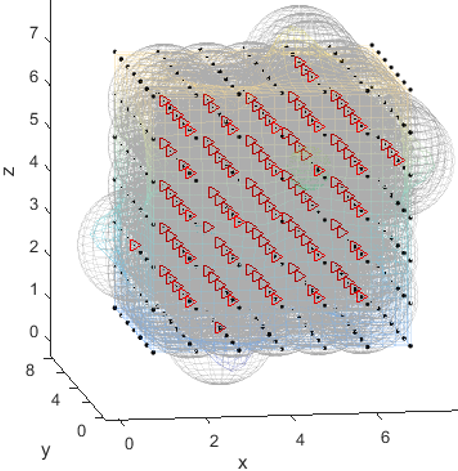}
    \caption[Positions of robots when $k=44$, ${\lvert V_{a} \rvert}=129$, $N_{i}=3$]{Positions of robots when $k=44$, ${\lvert V_{a} \rvert}=129$, $N_{i}=3$}
    \label{fig:31k44}
\end{figure}

An example of the procedure with $N_{i}=3$ is shown in Figure \ref{fig:293Dk15}, \ref{fig:30k30} and \ref{fig:31k44} for $k=15$, $k=30$ and $k=44$ separately and the deployment finished in 44 steps. The graphs show that robots will occupy the sections which are far from $p_{r}$ first, thus, vertices near initial positions are vacant for adding new robots. Therefore, the self-deployment result could be close to the optimal result. In Figure \ref{fig:31k44}, the light grey spheres demonstrate the $r_{st}$ for each vertex. The result shows that only the corners near the boundaries of the area are not covered. Based on Assumption \ref{a13}, those corners are ignored, and the complete coverage of the area is achieved.

\begin{table}
    \centering
    \renewcommand\arraystretch{2}
    \begin{tabular}{|c|c|c|c|c|c|}
        \hline 
        P&$\underline{r_{rob}}$&$    r_{safe}$&$    \underline{r_{so}}$&$    \underline{r_{st}}$&$    a$  \\ 
        \hline 
        R&    4.93m&    5.086m    &75m&    1km    &69.9140m    \\ 
        \hline 
        S&    4.93&    5.086&20.086    &    20.086&    15     \\ 
        \hline 
        P&$    \underline{r_{c}}$&$    \underline{v_{max}}$& $    \underline{\omega_{max}}$&$    t_{move}$&$    W_{pass}$ \\
        \hline
        R&91.4m&    2.315m/s    &N/A &N/A&131.2666m  \\
        \hline
        S& 30.156&    N/A&N/A    &5.75s& 36.1528 \\
        \hline
    \end{tabular} 
    \caption[Parameters of robots and simulation]{Parameters of robots and simulation}
    \label{t8}
\end{table}

The modified DR algorithm in \cite{RN138} is compared to the proposed algorithm. To make that algorithm work in the practical situation, the IP address of the robot which is unique for each robot is used as the order of choice. For $rob_{i}$ and its neighbors in $N_{c,i}(k)$, it is set that the one with lower IP address will choose earlier. $rob_{i}$ also informs its choice robots in $N_{c,i}(k)$ with lower priorities. Thus, communication neighbors will not choose the same vertex as $rob_{i}$ in the same step.

\begin{figure}
    \centering
    \includegraphics[width=0.8\linewidth]{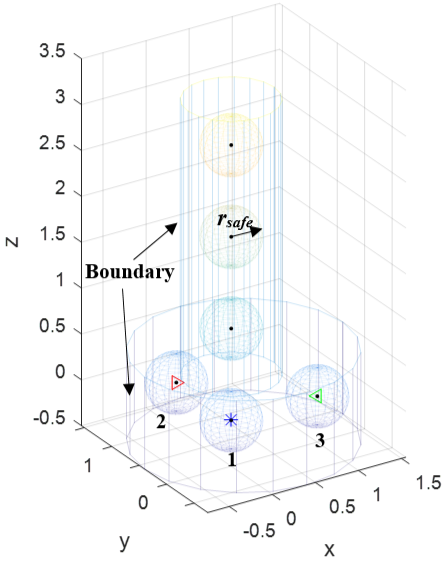}
    \caption[An example of the narrow path with $k=3$, $\lvert V_{a} \rvert=6$, $N_{i}=3$]{An example of the narrow path with $k=3$, $\lvert V_{a} \rvert=6$, $N_{i}=3$}
    \label{fig:323Derror}
\end{figure}

The calculated $N_{o}$ and the average simulation results of 300 tests for both the proposed algorithm and the DR algorithm are shown in Table \ref{t9va129}. It also shows the ratios of steps in the proposed algorithm to the steps in the DR algorithm. The numbers of steps of the proposed algorithm are less than 1/3 of that in the DR algorithm, and the advantage is clearer when $N_{i}$ is larger. The results of the proposed algorithm are only less than two steps more than $N_{o}$, so they are very close to the optimal numbers of steps. However, this also means, in some steps, there were less than $N_{i}$ robots added because less than $N_{i}$ robots could find vacant vertices to move to. One possible reason is that there were some narrow sections in the task area. An example of narrow section is shown in Figure \ref{fig:323Derror} where $N_{i}=3$, $\lvert V_{a} \rvert=6$ and vertex 1 is at $p_{r}$. The area to be covered is made up of two cylinders, and the balls show the $r_{safe}$ around each vertex so that it is clear that vertex 2 and 3 are blocked by boundaries and only vertex 1 can add new robots to the section of the thin cylinder. So, based on rule \ref{e3}, robots need four loops to finish the complete coverage. However, according to formula \ref{e4}, $N_{o}$=2. Another reason for the nonoptimal result is that $rob_{i}$ chose a $n_{s,i,j}(k)$ with the $d(n_{s,i,j}(k), p_{r})$ which was smaller than $d(p_{i}(k),p_{r})$, which means $rob_{i}$ moved toward $p_{r}$. If other robots at sensing neighbors of $rob_{i}$ had higher priorities than $rob_{i}$, they should have finished their choices so $p_{i}(k)$ could not be chosen in this step and would be left vacant in step $k+1$. Thus, the choice of $rob_{i}$ did not result in an accessible vertex near initial vertices to feed in a new robot.
\begin{table}
    \centering
    \renewcommand\arraystretch{2}
    \begin{tabular}{|c|c|c|c|c|c|c|}
        \hline 
        \multirow{2}{*}{Type of Results}&\multicolumn{6}{c|}{$N_{i}$} \\ 
        \cline{2-7} 
        & 3 & 5 &7 &    9    &11&    13 \\ 
        \hline 
        $N_{o}$&43&    26    &19    &15&    12&    10\\
        \hline
        Proposed algorithm&44.27&    27.84&    20.29&    16.57&    13.53&    11.66\\ 
        \hline 
        DR algorithm&    139.65&    93.04&    68.36&    62.65&    53.91&    46.26\\ 
        \hline 
        Proposed/DR&31.7\%&    29.9\%&    29.7\%&    26.5\%&    25.1\%&    25.2\%\\ 
        \hline 
    \end{tabular} 
    \caption[Steps from calculations and simulations with $\lvert V_{a} \rvert=129$]{Steps from calculations and simulations with $\lvert V_{a} \rvert=129$}
    \label{t9va129}
\end{table}

To test the scalability of the proposed algorithm, areas with the same shape but different sizes with $N_{i}=3$ are used. The test areas have 6*6*6 vertices and 8*8*8 vertices with $\lvert V_{a} \rvert=62$ and $\lvert V_{a} \rvert=210$ respectively. The average results of 300 simulations of both algorithms, the ratios of the two algorithms and $N_{o}$ are in Table \ref{t10No678}. It shows that the proposed algorithm is always better than the DR algorithm. As the area expands, the ratio decreases which means the advantage is clearer than that in a small area. The difference between the proposed algorithm and $N_{o}$ also decreases because there are less narrow areas. The above simulations demonstrate that the algorithm is scalable and effective.

\begin{table}
    \centering
    \renewcommand\arraystretch{2}
    \begin{tabular}{|c|c|c|c|}
        \hline 
        \multirow{2}{*}{Type of Results}&\multicolumn{3}{c|}{Size of the Area} \\ 
        \cline{2-4} 
        & 6*6*6    &7*7*7&    8*8*8  \\ 
        \hline 
        $N_{o}$&    21&    43    &70\\
        \hline
        Proposed algorithm&    23.70&    44.27&    70.98\\ 
        \hline 
        DR algorithm&    66.03&    139.65    &248.24\\ 
        \hline 
        Proposed/DR&    35.9\%&    31.7\%    &28.6\%\\ 
        \hline 
    \end{tabular} 
    \caption[Steps from calculations and simulations for different areas with $N_{i}=3$]{Steps from calculations and simulations for different areas with $N_{i}=3$}
    \label{t10No678}
\end{table}

\subsection{Section Summary}
This section provided a decentralized self-deployment algorithm for multiple mobile robotic sensors to cover a 3D area completely. The area is arbitrary unknown with obstacles, and the coverage algorithm used a grid pattern to have the least number of robots and avoid collisions. This section proved that the algorithm could converge with probability 1 and the results of this complete coverage algorithm are very close to the fewest number of steps. Comparing to another decentralized algorithm with the same initial condition, the proposed algorithm is effective and scalable.
\section{Summary}
A decentralized complete coverage algorithm for the mobile robotic sensor network was proposed for both 2D and 3D area. The task area is arbitrary unknown with obstacles and robots deploy themselves using the grid without any collisions. In 2D areas, the 100\% coverage was reached under the strict requirement, and in 3D areas, each grid point in the task area was occupied by a mobile sensor. Comprehensive simulation of the algorithm for areas with different sizes and different initial points were run based on the parameters of real robots. The simulation results illustrate that the step in the algorithm is very close to the ideal value and this algorithm is much better than the DR algorithm. This chapter also proved the convergence of the algorithm with probability 1. The algorithm is applicable in both 2D and 3D areas, and only slight changes in the dimension of some values are needed. 
\chapter{A Collision-Free Random Search Algorithm}
\chaptermark{A Collision-Free Random Search Algorithm}
Chapter 4, 5 and 6 provide three decentralized collision-free algorithms for search tasks in an unknown area by multiple robots. They all use grid patterns as discussed in Chapter 2 and all the targets are static. So the problem statement and the algorithm in this chapter will be explained and discussed in detail. Although some of them will be the same as those in the complete coverage, including them here will make this chapter self-contained and make readers understand the discussion easier. Chapter 5 and 6 will refer to this chapter and only describe the problem statement and algorithm briefly and focus on the different parts.

According to Chapter 3, in a large task area, monitoring the whole area requires a significant number of robots which will be costly. Therefore, with a limited number, robots need to move around to detect the entire area gradually or find some targets which will be the search task. There are two types of search tasks in an unknown environment. If the number of targets is known to the robot a priori, robots only need to find all the targets to finish the work. This situation is called situation I which can be seen in the search for a certain number of people lost in a forest. Otherwise, the number of targets is unknown, it is called situation II, and robots need to detect every corner of the area to achieve the goal such as the mine countermeasures in the battlefield.  

This chapter proposes a grid-based decentralized random algorithm for search tasks in an unknown area. The performance of the algorithm will be analyzed using simulations. The proposed algorithm is suitable for both 2D and 3D areas. The main differences are the grid pattern and the dimension of some parameters as discussed in Chapter 2. The problem in a 2D area will be discussed in Section \ref{s5search2D1}, and the task in a 3D space will be in Section \ref{s6search3D1}.

\section{2D Tasks}
\label{s5search2D1}
This section aims to design a collision-free decentralized algorithm for a group of robots to search targets in an unknown area with static obstacles by visiting vertices of a grid pattern.
\subsection{Problem Statement}
\begin{figure}
    \centering
    \includegraphics[width=1\linewidth]{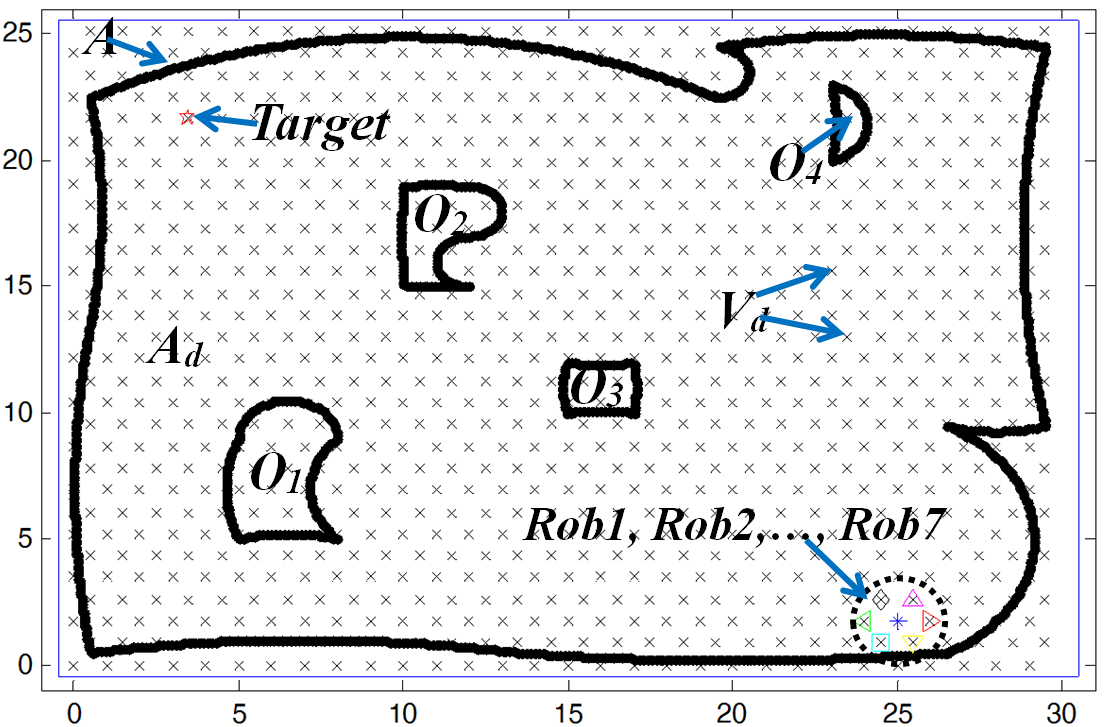}
    \caption[Initial settings of the map]{Initial settings of the map}
    \label{fig:332Dinitialsetting}
\end{figure}

The searched area in this section is a two-dimensional area $A\subset\mathbb{R}^{2}$ with limited number of obstacles $ O_{1} $, $ O_{2} $,$\ldots $, $O_{l} $. There are $m$ robots labeled as $rob_{1}$ to $rob_{n}$ in this area as shown in Figure \ref{fig:332Dinitialsetting}. An arbitrary known T grid pattern with a side length of $a$ is used, and robots can only move along the edge connection neighbor vertices in each step. 
\begin{assumption}
    The area $A$ is a bounded, connected and Lebesgue measurable set. The obstacles $O_{i}$ are non-overlapping, closed, bounded and linearly connected sets for any $i\geq{0}$. They are both unknown to the robots. \label{a:1area}
\end{assumption}

\begin{definition}
    Let $O:=\bigcup{O_{i}}$ for all $i>0$. Then $A_{d}:=A\setminus{O}$ represents the area that needs to be detected.\label{d:1obs}
\end{definition}
There are $n$ static targets $t_{1}$ to $t_{n}$ which are labeled as red stars in Figure \ref{fig:332Dinitialsetting}. Let $T$ be the set of target coordinates and $T_{ki}$ represents the targets set known by $rob_{i}$. The symbol $\lvert\bullet\rvert$ in this report represents the number of elements in set $\bullet$. Then $\lvert T\rvert$ denotes the number of targets in the search area and $\lvert T_{ki}\rvert$ denotes the number of targets known by robot $rob_{i}$. There are two kinds of targets searching problems. In situation I, $\lvert T\rvert$ is known and in situation II, no information about $\lvert T\rvert$ is given.

\begin{assumption}
    The initial condition is that all robots use the same T grid pattern and the same coordinate system. They start from different vertices near the boundary which can be seen as the only known entrance of the area. \label{a:Initial_P}
\end{assumption}

Although \cite{RN119,RN118,RN229,RN234,RN1,RN101,RN232,RN233} provided a consensus algorithm to achieve this assumption, it needs infinite time and has collisions between robots, so the algorithm is impractical. If it is only approximately reached in a short period, the sensing range and the communication range should be designed large enough to ignore small errors between coordinate systems, but this was not mentioned. In MATLAB simulations, robots are manually set to start from vertices. In robot tests, all robots could be released from the same vertex but configured to move to different nearby vertices at the beginning. Another possible solution could be using one robot and its coordinate system as a standard, and each other robot is put close to a distinct vertex near the standard. After sensing and communication with the standard robot directly or indirectly, other robots will know the positions of themselves and the corresponding closest vertices. Thus, they can go to those vertices without any collision.

\begin{assumption}
    All the ranges and radius in 2D search problems are circular. \label{a:circular}
\end{assumption}

The swing radius of robots is $ r_{rob} $. To avoid collision with other objects such as boundary and other robots, errors need to be considered. Let the maximum possible error containing odometry errors and measurement errors be $e$. Then a safety radius $ r_{safe} $ which is larger than $ r_{rob} $ for collision avoidance can be defined as follows.
\begin{definition}
    To avoid collisions with other things before move, a safe radius $ r_{safe} $ should include both $ r_{rob} $ and $e$ (see Figure \ref{fig:5r_ob}). So $ r_{safe}\geq{r_{rob}+e} $.\label{d:rsafe}
\end{definition}
\begin{assumption}
     As the T grid is used, all passages $W_{pass}$ between obstacles or between an obstacle and a boundary are wider than $a+2*r_{safe}$ based on Table \ref{t1}.\label{a:wpass}
\end{assumption}

This guarantees that there is always an approachable vertex in the path and robots can go through the passage with no collision. Obviously, decreasing $a$ will expand the scope of applicability. However, the number of vertices increases so the search time will also increase. To search with collision avoidance, robots carry sensors to detect the environment including both targets and obstacles. 
\begin{assumption}
    The sensing radius for obstacles is $r_{so}$ where $r_{so}\geq{a+r_{safe}}$.\label{a:rso}
\end{assumption}

Therefore, if the neighbor vertices of a robot are less than $r_{safe}$ away from the boundary or obstacle, $r_{so}$ is large enough to detect that so the robot will not visit those vertices to avoid collision. $r_{so}$ can be seen in Figure \ref{fig:5r_ob} and \ref{fig:6rangeorder}. The vertices which are within $r_{so}$ of $rob_{i}$ are called sensing neighbors of $rob_{i}$. Let $V_{a}$ represents the set which includes all the accessible vertices. Then let $N_{s,i}(k)$ represents the sensing neighbors set that contains the closest vertices $n_{s,i,j}(k)$ around $rob_{i}$ at discrete time $k$ where $j\in{\mathbb{Z}^{+}}$ is the index of the neighbor vertices. In $N_{s,i}(k)$, let the set of visited vertices of $rob_{i}$ be $V_{v,i}(k)$ and the corresponding set for unvisited vertieces be $V_{u,i}(k)$. So $0\leq{\lvert V_{v,i}(k)\rvert}\leq{6}$ and also $0\leq{\lvert V_{u,i}(k)\rvert}\leq{6} $. Then let the choice from $V_{v,i}(k)$ be $c_{v}$ and the choice from $V_{u,i}(k)$ be $c_{u}$.

Apart from obstacle sensing, robots also need to have a sensor to detect targets which has a range of $r_{st}$. Under the strict requirement, it is similar to $r_{so}$. An assumption for $r_{st}$ is needed to guarantees that areas around inaccessible vertices can still be detected.
\begin{assumption}
    $ r_{st} \geq{a+r_{safe}}$.\label{a:rst}
\end{assumption}

In the strict requirement, each point should be detected, and a T grid is needed, the curvature assumption will guarantee this. 
\begin{assumption}
    The curvature of the concave part of the boundary should be smaller than or equal to $1/r_{st} $.\label{a:2D1curvature}
\end{assumption} 

For the loose requirement, the above two assumptions are not needed as $r_{st}$ should be chosen based on Table \ref{t1} other than Assumption \ref{a:rst} and assumption for boundary effect will replace \ref{a:2D1curvature} to ensure a complete search which is stated below. 
\begin{assumption}
    The search task for a team of robots usually has a vast area. So, $a$ is much smaller than the length, the width and the height of area $A$. Therefore, the boundary effect is negligible \cite{RN122,RN279}. \label{a:2D1boundary}
\end{assumption} 

The communication range of robots is represented as $ r_{c} $. Robots within $ r_{c} $ of $ rob_{i} $ are named as the communication neighbors of $ rob_{i} $.  As the range is limited, the communication between robots is temporary as in an ad-hoc network, so each robot is considered equal with local communication ability. To avoid choosing the same vertex with other robots in the same step, robots could choose in sequence use the proposed random algorithm and tell the choice to communication neighbors who may select it. Thus, others will think that vertex as an obstacle. The way to set the sequence of choice is described in Section \ref{s72Dalgo}. Based on the T grid, those robots are one or two vertices away from it. Therefore, $r_{c} $ should satisfy Assumption \ref{a:rc} with error $e$ considered.
\begin{assumption}
    $ r_{c}\geq{2*a+e}$ (see Figure \ref{fig:6rangeorder}). \label{a:rc}
\end{assumption}

Let $ N_{c,i}(k)$ denotes a set of communication neighbors around $rob_{i}$ that it must communicate with. According to Figure \ref{fig:6rangeorder}, max($\lvert N_{c,i}(k) \rvert)=18$. 
\subsection{Procedure and Algorithm} 
\label{s72Dalgo}
\begin{figure}
    \centering
    \includegraphics[width=1\linewidth]{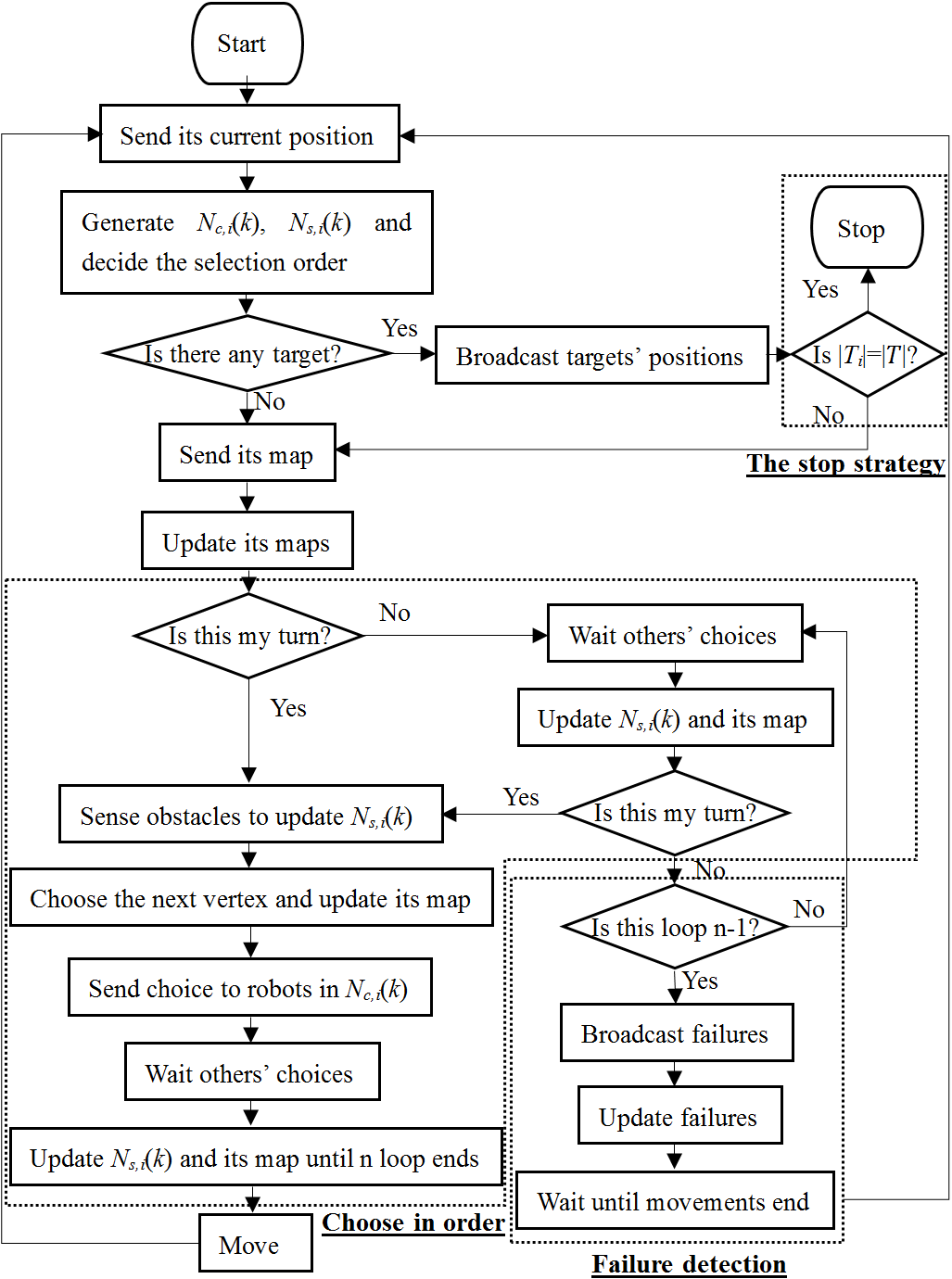}
    \caption[The flow chart for situation I]{The flow chart for situation I}
    \label{fig:342dflowsitu1}
\end{figure}

\begin{figure}
    \centering
    \includegraphics[width=1\linewidth]{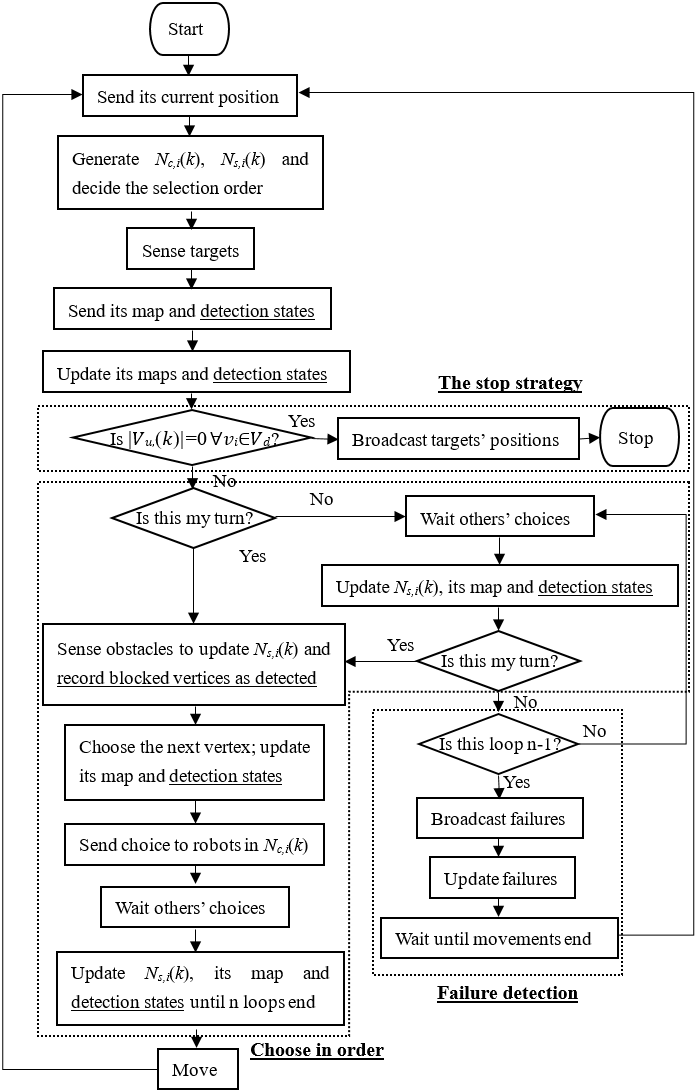}
    \caption[The flow chart for situation II]{The flow chart for situation II}
    \label{fig:352dflowsitu2}
\end{figure}

The algorithm directs the robots to explore the area gradually vertex by vertex. The whole progress can be seen in Figure \ref{fig:342dflowsitu1} for situation I and in Figure \ref{fig:352dflowsitu2} for situation II. The stop strategy and the time to do it in the two situations are different. The description of the algorithm is based on situation I. Extra steps for situation II are mainly about detection states of the vertices. From the previous section, a vertex is said to be detected in two conditions. If a vertex has obstacles within $r_{safe}$, the detection state for this vertex and surrounding area is judged from robots at its sensing neighbors. If the vertex is accessible, this vertex and surrounding area are said to be detected after it is visited. They are underlined in Figure \ref{fig:352dflowsitu2} and will be specially noted if necessary. 

\subsubsection{Preparation for the Choice}
\label{algo2Dchoose}
Initially, robots exchange their positions with robots within $r_{c}$. Based on the relative positions, the recognized communication neighbors of $rob_{i}$ will form the set $N_{c,i}(k)$. The recognized robots staying at sensing neighbors will be thought as obstacles by $rob_{i}$ and their occupied vertices will be removed from $N_{s,i}(k)$. Then the sequence of choice can be decided to avoid choosing the same next vertex by different robots. For $rob_{i}$ at $p_{i}(k)$ and its communication neighbor $rob_{j}$ at $p_{j}(k)$, if $p_{jx}>p_{ix}\mid(p_{jy}>p_{iy}\&p_{jx}=p_{ix})$, $rob_{j}$ will have higher priority to make choice. Figure \ref{fig:6rangeorder} visualizes the above relation. In that figure, robots with higher priorities should be on the vertices at the right side of the green curve. So, robots at vertices on the left will choose later than $rob_{i}$. In this way, robots at half of the neighbor vertices will choose earlier than $rob_{i}$ and there will be no contradiction when judging relative orders from the views of different robots.

Subsequently, robots sense the local area within $r_{st}$ to detect targets. As positions of other robots are known, their occupied zone will not be detected. In situation I, when a target is detected, the information will be broadcast to all robots. The broadcast is used as the positions of targets are the key results to find for all robots, and it only happens $\lvert T\rvert$ times, so this will not consume too much energy. After this, all robots will go to stop strategy. If robots are not ready to stop or no targets are sensed, robots will go back to the regular loop on the left side of the figure. In situation II, the information of targets will only be broadcast after the stop is confirmed. Robots only need to move to the next step of the flowchart regardless of the sensing results. 

Next, the robot shares its explored map which is updated based on the position received in the first step. Then robots can update their maps again. In situation II, whether the sensing neighbors of the vertices are detected or not will also be sent and received to update the detection states of all vertices. 
\subsubsection{Wait, Choose, and Communicate}
$rob_{i}$ needs to wait for the choices from the robots with higher priorities because the positions of those choices can be thought as detected points in the next step and may also be obstacle areas for $rob_{i}$. If all robots are connected, the last one in the choosing sequence needs to wait for every other robot and the processing time for the whole selection process should be set to the value in this condition because all the robots should be synchronized in each loop. When it is the turn of $rob_{i}$, it will detect the local area but avoid the section of existing neighbor robots. The way to judge whether the object $p$ is an obstacle or not is illustrated in Figure \ref{fig:22judgeobs} in Section \ref{s8ch3algo}. If there is any sensed object on the way to the vertex $n_{s,i,j}(k)$ in $N_{s,i}(k)$, that object is treated as an obstacle. So that the vertex $n_{s,i,j}(k)$ cannot be reached and is removed from $N_{s,i}(k)$. For situation II, that vertex is also recorded as detected. If no objects are sensed in these two situations, this robot will still wait for the same amount of processing time, so all robots are synchronized in this step.

Choices are made using the following random decentralized algorithm.
\begin{equation}
p_{i}(k+1)=
\begin{cases}
c_{u} \text{ with prob. } 1/\lvert V_{u,i}(k)\rvert, \text{ if } \lvert V_{u,i}(k)\rvert\neq{0},\\
c_{v} \text{ with prob. } 1/\lvert V_{v,i}(k)\rvert, \text{ if } (\lvert V_{u,i}(k)\rvert=0)\&(\lvert V_{v,i}(k)\rvert\neq{0}),\\
p_{i}(k),  \text{ if } \lvert N_{s,i}(k) \rvert=0
\end{cases}
\label{e52Dchoose}
\end{equation}
This algorithm means when there are unvisited sensing neighbors, $rob_{i}$ will choose one from them randomly. If there are only visited sensing neighbors around, $rob_{i}$ will randomly choose one from them. If there are no vacant sensing neighbors to go, $rob_{i}$ will stay at the current position $p_{i}(k)$.

The choice of $rob_{i}$ is also added to its map and sent to robots in $N_{c,i}(k)$. The receiver $rob_{j}$ with lower priority needs to remove $p_{i}(k+1)$ from $N_{s,j}(k)$ if it was in that set. In situation II, the detection state of vertices should also be updated after making a choice and receiving the choice. 

In contrast to the random choice, there is also an algorithm with deterministic choice, namely using the first available choice. The first choice is used as there may not be another option. However, if the first choice is defined based on heading angle of the potential movement, robots tend to move in the same direction instead of searching all directions evenly which may lead to more repeated visitation. Thus, it will have a longer search time. 
\subsubsection{Move}
All robots should move in the same pattern at the same time, and the way to move should match the realistic constraints. 

Each movement step consists of two phases: phase I is the pure rotation, and phase II is the pure translation. The rotation will aim the heading of the robot at the target vertex. Specially, if $rob_{i}$ has been facing to its selected vertex before phase I, it will not turn but wait for the rotation time. Then $rob_{i}$ can go straight along the chosen edge. Robots have standard hard constraints on the velocity and the acceleration so that both phases use trapezoidal velocity profiles based on parameters of the area and the test robots. In this section, parameters from Pioneer 3-DX are used in simulation as in Table \ref{t5}.
\subsection{Stop Strategies}
For situation I, a robot judges the stop condition after exchanging target information. $rob_{i}$ should stop at time $k$ when the number of targets that it finds equals to the number of targets in the area. The knowledge of $\lvert T_{ki}\rvert$ comes from its detection and the information received from other robots. Namely, for any robot, 
\begin{equation}
p_{i}(k+1)=p_{i}(k)  \text{, if } \lvert T_{ki}\rvert=\lvert T\rvert
\label{e62DstopI}
\end{equation}
In situation II, the judgment for stop condition happens in step 6 in the flowchart. For each visited vertex, the detection states of its sensing neighbors need to be checked. If all visited vertices have no undetected sensing neighbors, the robot should stop. Namely, 
\begin{equation}
p_{i}(k+1)=p_{i}(k)  \text{, if } \forall v_{i}\in V_{a}, \text{ } \lvert V_{u,i}(k)\rvert=0
\label{e72DstopII}
\end{equation}

The aims of the tasks can be various. So robots may have different missions after stop moving.  However, this paper focuses on search tasks only so looking for the coordinates of the targets is the only requirement. In situation I, when the judgment condition of $rob_{i}$ is satisfied, all robots stop the whole procedure as they have the same information of targets which is broadcast every time. However, in situation II, if $rob_{i}$ satisfies the stop condition, it needs to broadcast its knowledge of targets and a stop signal to inform all other robots. Then all robots will broadcast the positions of targets that they know and terminate the search.
\subsection{Broken Robots and Reliability}
Waiting for information to choose means the failure of robots with higher priorities will affect those with lower priorities and thus break the loop of the flowchart. To improve the reliability of the algorithm, the failure of robots need to be found and a solution to tackle it is provided as follows. If $rob_{i}$ has not received any information from robots with higher priorities until the $(n-1)$ waiting loop, it means that at least one robot is not working. Therefore, $rob_{i}$ will broadcast that the robots which should send their choices but did not do failed and claim that $rob_{i}$ itself is still working. If both failure and working states of one robot are received in this loop, that robot will be judged as working. Then, those robots which send information will stay at current positions and wait until the time slot for move passed. In the next loop, robots will not wait for the failed robots but just consider it as an obstacle. This scheme detects the failure after the first step in the flowchart and before the movement. However, the detection does not include the fault in communication. By having this broken robots detection scheme, this decentralized algorithm is more robust than centralized algorithms and robots will still work even when there is only one working robot left. Under this circumstance, the system could only fail in the extreme condition where broken robots blocked all the paths to a target so that the target cannot be detected.
\begin{theorem}
    \label{th2D1}
    Suppose that all assumptions hold and the random decentralized algorithms \ref{e52Dchoose} with a related judgment strategy \ref{e62DstopI} or \ref{e72DstopII} are used. Then for any number of robots, there is a time $k_{0}>0$ such that all targets are detected with probability 1.
\end{theorem}
\begin{prove}
    The algorithms \ref{e52Dchoose} \& \ref{e62DstopI} or \ref{e52Dchoose} \& \ref{e72DstopII} forms an absorbing Markov chain which includes both transient states and absorbing states. The transient states are the steps that robots visit the approachable vertices of the grid but do not stop. Absorbing states are the steps that all robots stop at the vertices. Applying this algorithm, robots tend to go to the unvisited neighbor vertices. If all sensing neighbors of a robot are visited before finding the targets, robots will choose a random sensing neighbor, continue the regular loop and only stop when all targets are found in Situation I or all approachable vertices are visited, and all inapproachable vertices are detected from their sensing neighbors in situation II. For \ref{e52Dchoose} \& \ref{e62DstopI}, the number of transient states will decrease until all robots know $\lvert T_{ki}\rvert=\lvert T\rvert$. For \ref{e52Dchoose} \& \ref{e72DstopII}, absorbing states are reached until $\lvert V_{u,i}(k)\rvert=0$ for any accessible vertices. Based on Assumption \ref{a:wpass}, absorbing states can be achieved from any initial state, namely with probability 1. This completes the proof of Theorem \ref{th2D1}.
\end{prove} 

\subsection{Simulation Results}    
Simulations of the situation where the number of robots is known, namely situation I, is used to verify the algorithm. The loose requirement is applied for the area, and the T grid is used for the robot to move through. In simulations, there is one target randomly put in the area $A$, and there are several robots to search it (see Figure \ref{fig:332Dinitialsetting}). The aim is to find the target position in $A$ by robots via the T grid pattern using the proposed algorithm. Based on this aim, when the target is found, the corresponding robot stops and shares this as global information. The search procedure should always have no collision. The bold black line made up by points in the map displays the boundary of area $A$ and obstacles. The distances between neighbor points are small enough which is set according to Assumption \ref{a:wpass} so that no robots can pass the boundary. Parameters in the simulation are designed based on the pioneer 3-DX robot. Notice that move time is roughly estimated as 5.75s based on Table \ref{t5}. To test all the situations in algorithm \ref{e52Dchoose}, at least seven robots are needed. So the simulations with one to seven robots for the same target starting from similar places are run to check the relation between the number of steps of choice, search time and the number of robots. To see the route clear and show the collision avoidance. The target is put far from the starting points with obstacles in between. For the test with seven robots, one robot will be enclosed by other robots as shown in Figure \ref{fig:332Dinitialsetting}. Then all six sensing neighbors will be seen as obstacles by the middle one which is the third situation of algorithm \ref{e52Dchoose}. In later loops, the first and second situations could happen. 

\begin{figure}
    \centering
    \includegraphics[width=0.8\linewidth]{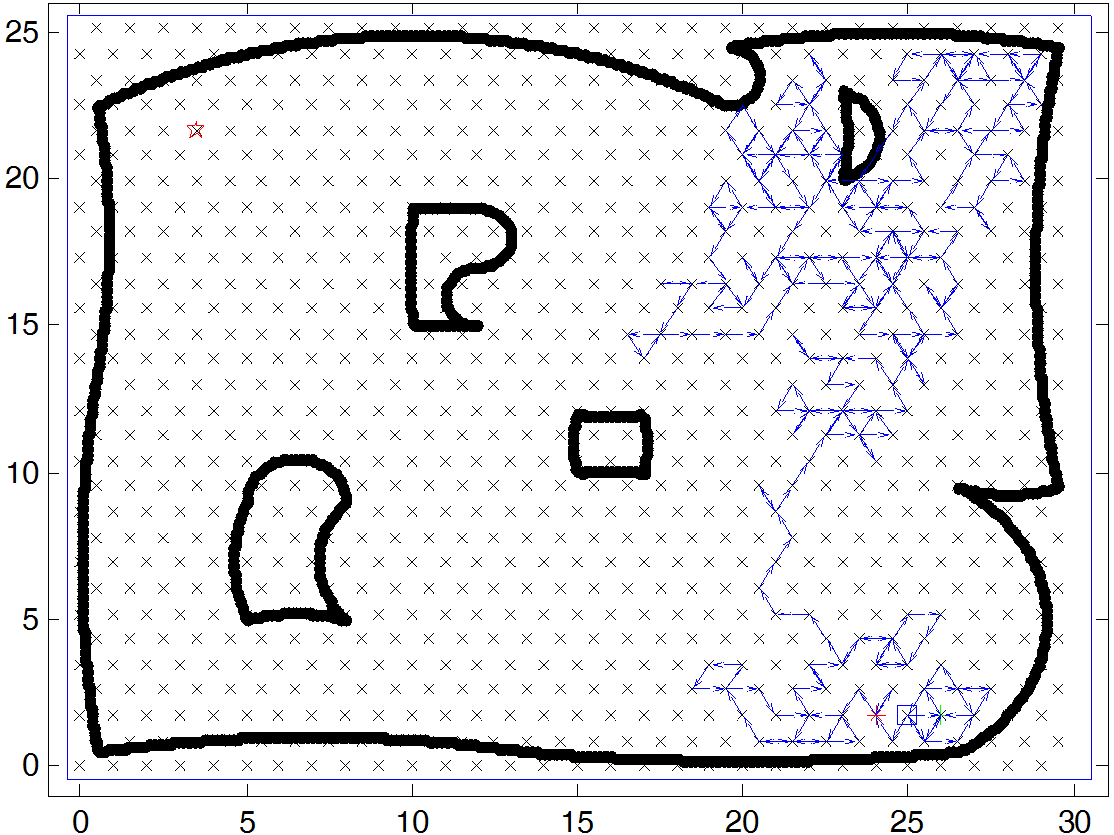}
    \caption[The route of robot 1 in a simulation of 3 robots]{The route of robot 1 in a simulation of 3 robots}
    \label{fig:36route1}
\end{figure}

\begin{figure}
    \centering
    \includegraphics[width=0.8\linewidth]{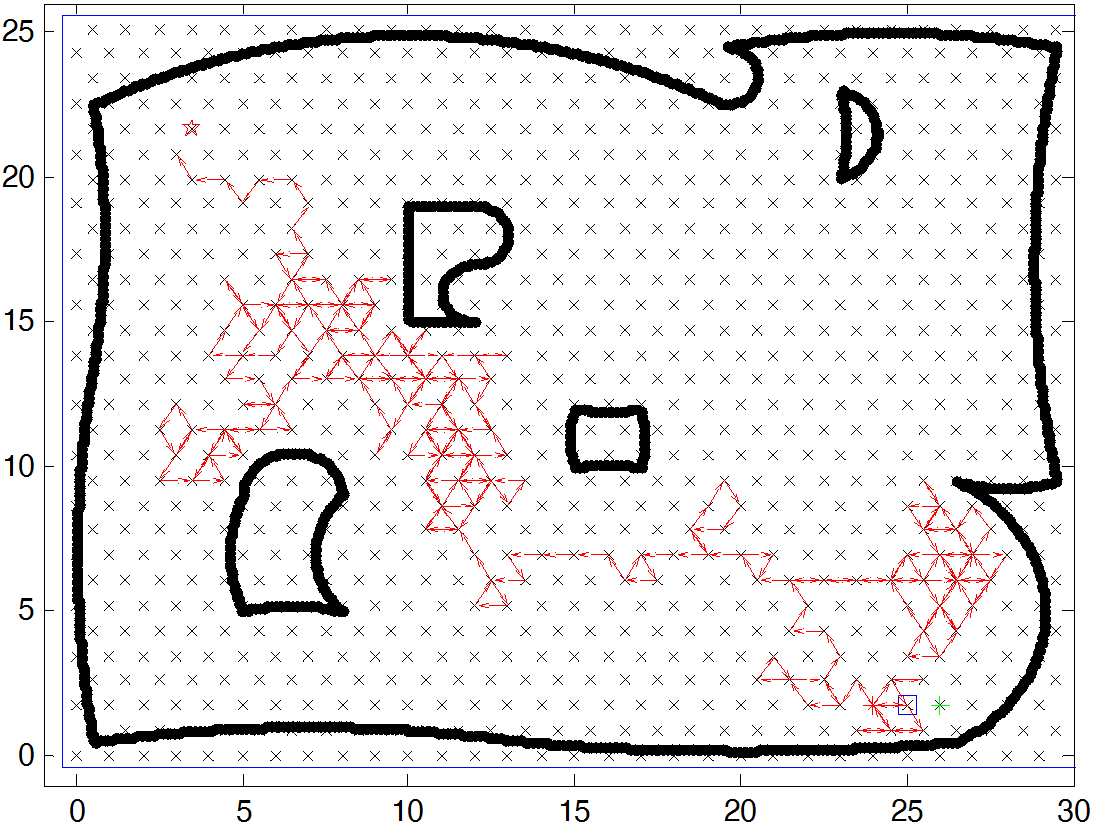}
    \caption[The route of robot 2 in a simulation of 3 robots]{The route of robot 2 in a simulation of 3 robots}
    \label{fig:37route2}
\end{figure}

\begin{figure}
    \centering
    \includegraphics[width=0.8\linewidth]{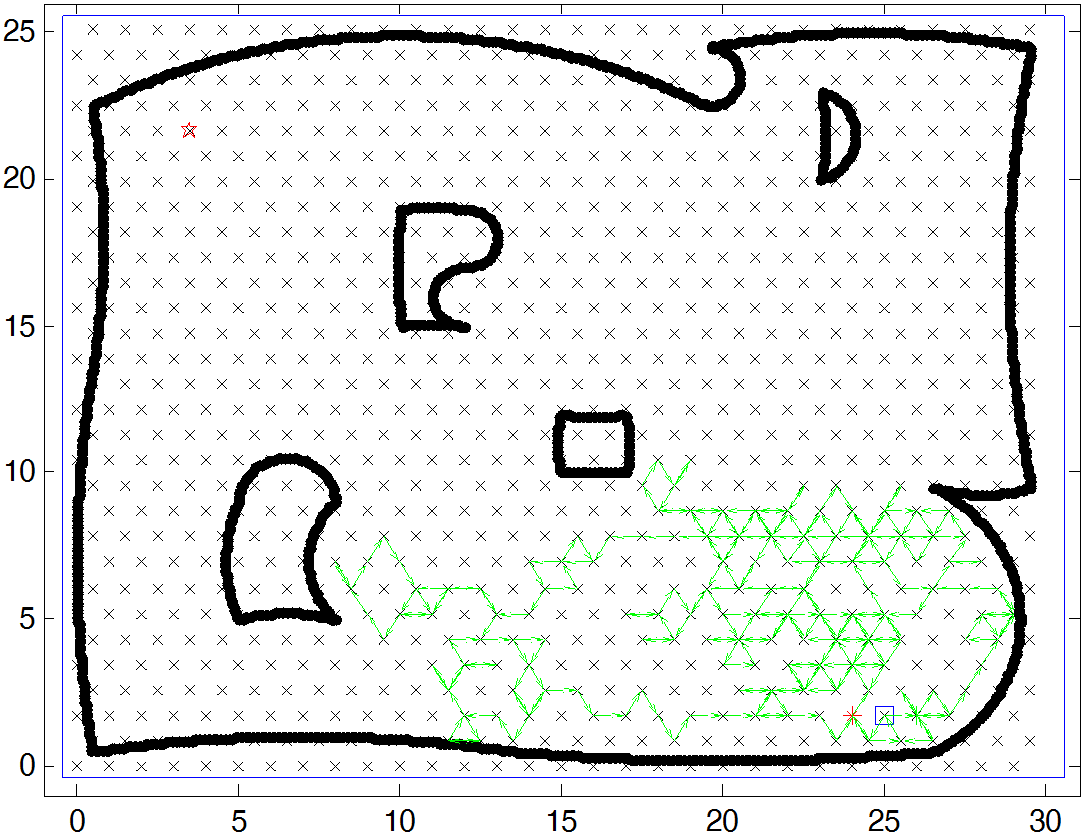}
    \caption[The route of robot 3 in a simulation of 3 robots]{The route of robot 3 in a simulation of 3 robots}
    \label{fig:38route3}
\end{figure}

\begin{figure}
    \centering
    \includegraphics[width=1\linewidth]{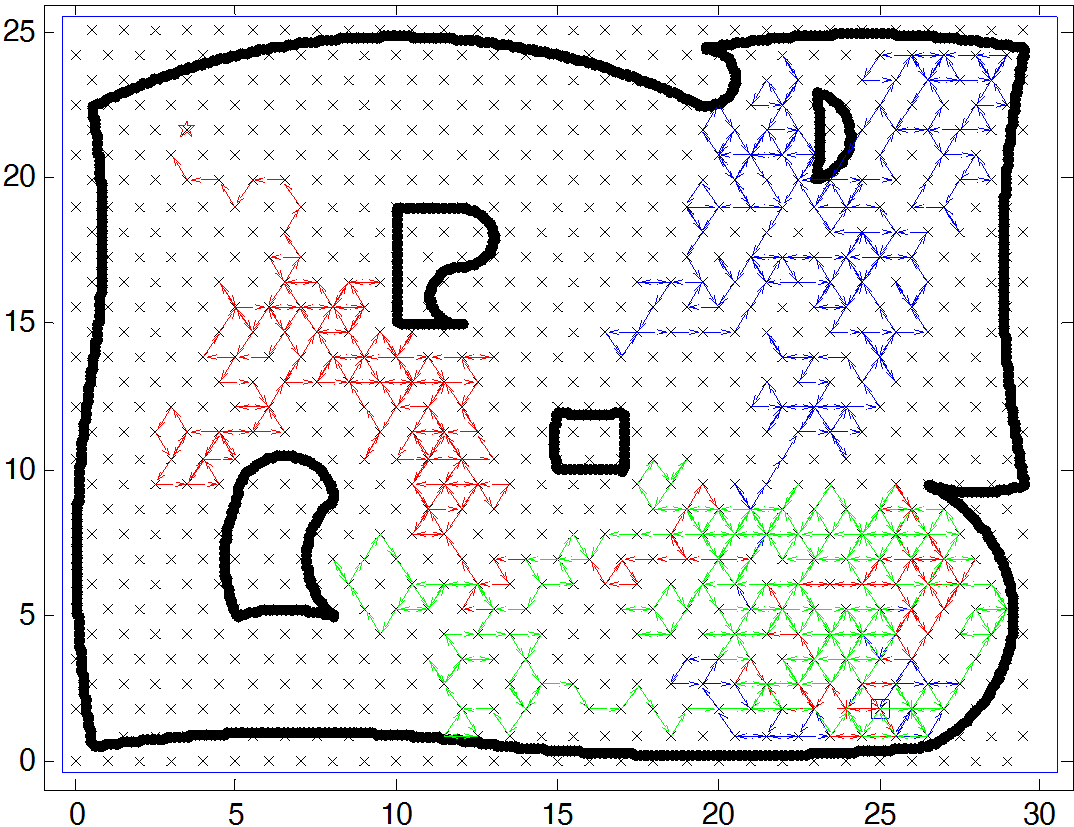}
    \caption[Combined routes in a simulation of 3 robots]{Combined routes in a simulation of 3 robots}
    \label{fig:39route123}
\end{figure}

Figure \ref{fig:36route1}-\ref{fig:39route123} show the routes of a simulation with 3 robots. Figure \ref{fig:36route1}-\ref{fig:38route3} display the route of each robot with arrows to illustrate the directions of the movements. Figure \ref{fig:39route123} is the combination of the three routes where the repeated sections of routes can be found due to distributed decisions made when the communication is not available. By analyzing the route of each robot which is recorded in each step, no collisions between robots are found, and the algorithm is functioning well.


\begin{figure}
    \centering
    \includegraphics[width=0.7\linewidth]{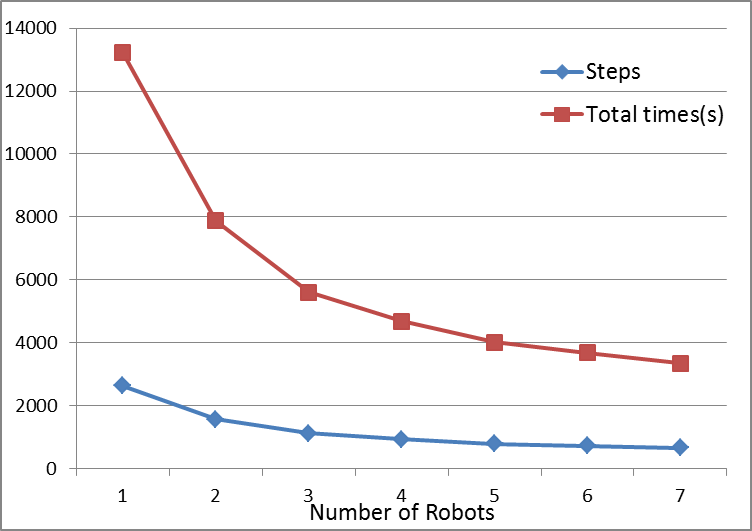}
    \caption[Time and steps]{Time and Steps}
    \label{fig:40algo1result}
\end{figure}

The results in Table \ref{t10algo1result} and corresponding Figure \ref{fig:40algo1result} are the average results of 100 tests for 1 to 7 robots. The row `Code' shows the time for the calculation in the algorithm. The row `Total' includes both the calculation time and the movement time. It demonstrates that if more robots are working together, the number of search steps will decrease significantly at the beginning and gently later. However, both the number of robots $n$ and the choosing time increase, so the waiting time in each step which is $n$ multiplies the time for `choose' also increases.  It can be seen clearly that the moving time is much larger than the algorithm time so that the trend of total time is similar to that of the number of steps, namely descending sharply from 1 to 5 robots and decreasing gradually from 5 to 7 robots. It can be predicted that if there are more robots in this search area, the total time may keep stable followed by a moderate rise as the increased algorithm time takes up larger proportion in the total time. 
\begin{table}
    \centering
    \renewcommand\arraystretch{2}
    \begin{tabular}{|c|c|c|c|c|c|c|c|}
        \hline 
        No.&    1&    2&    3&    4    &5&    6&    7  \\ 
        \hline 
        Steps&    2638&    1572&    1116&    931    &795&    730&    661  \\ 
        \hline 
        Code(s)    &22.0&    27.0&    29.0&    30.3&    33.7&    37.2&    39.0  \\ 
        \hline 
        Total(s)&    13211&    7885&    5607&    4687&    4009&    3688&    3345  \\ 
        \hline 
    \end{tabular}  
    \caption[$ W_{pass} $ and $ r_{st} $ for each grid]{$ W_{pass} $ and $ r_{st} $ for each grid}
    \label{t10algo1result}
\end{table}

In this algorithm, the number of robots is flexible, and the algorithm is scalable. The search time is also affected by the initial positions of the robots and the shape of the area. As that information is unknown in the problem discussed in this paper, the suitable range of the number of robots may only be estimated in simulations with a general area when the rough size is given.  Even for a fixed map, the optimal number of robots should be decided by comparing the benefits from the decline of the time and the cost of adding robots.
\subsection{Section Summary}
This section presented a decentralized random algorithm for multiple robots to search static targets in an unknown 2D area. This algorithm sets a sequence for making choices to avoid collisions and uses assumptions based on realistic conditions. A T grid pattern and the loose requirement of the area were used as an example. Robots started with a preset common T grid pattern and the same coordinate system. Then they used the proved algorithm with local information to choose future steps even if some robots failed in the calculation. Simulation results for 1-7 robots showed the flexibility and scalability of the algorithm.

\section{3D Tasks}
\label{s6search3D1}
This section describes a search task for multiple robots in a 3D area using the decentralized random algorithm proposed in the 2D situation. The main difference between these two areas is the dimension of the area and parameters. So the assumptions and definitions are given directly without repeated explanation. The performance of the algorithm will also be analyzed using simulation in MATLAB. As said in Chapter 2, the data of Bluefin-21 robots will be utilized, and the C grid will be employed in the simulations.
\subsection{Problem Statement}
Problems in this section is for a three-dimensional area $A\subset\mathbb{R}^{3}$ which has a limited number of obstacles $ O_{1} $, $ O_{2} $,$\ldots $, $O_{l} $. They need to satisfy the following assumption: 
\begin{assumption}
    Area $A$ is a bounded, connected and Lebesgue measurable set. The obstacles $O_{i}\subset\mathbb{R}^{3}$ are non-overlapping, closed, bounded and linearly connected sets for any $i>0$. They are both static and unknown to robots. \label{a:3Darea}
\end{assumption}
\begin{definition}
    Let $O:=\bigcup{O_{i}}$ for all $i>0$. Then $A_{d}:=A\setminus{O}$ is the area that needs to be detected.\label{d:3Dobs}
\end{definition}

There are $m$ robots labelled as $rob_{1}$ to $rob_{m}$ to search $n$ static targets $t_{1}$ to $t_{n}$. The coordinate of $rob_{i}$ is $p_{i}(k)$ at time $k$. Let $T$ be the set of coordinates of targets and $T_{ki}$ represents the set of coordinates fo targets known by $rob_{i}$.
\begin{assumption}
    Initially, all robots have the same grid with the same coordinate system and are allocated at different vertices along the border of the search area. In the simulation, robots are manually set at vertices of the C grid\label{a:3Dintialposition}
\end{assumption}
\begin{assumption}
    All sensors and communication equipment on robots have spherical ranges. Other radiuses mentioned are also spherical. \label{a:3D ranges}
\end{assumption}

Let the swing radius of robots be $r_{rob}$ and the safety radius to avoid collisions be $r_{safe}$. The total error is $e$. Then
\begin{definition}
    $ r_{safe}\geq{r_{rob}+e} $ (see Figure \ref{fig:16range3d}).\label{d:3Drsafe}
\end{definition}

The C grid has a side length of $a$. According to Chapter 2, the passage width $W_{pass}$ satisfies
\begin{assumption}
    $W_{pass}\geq{\sqrt{3}a+2r_{safe}}$ (see Figure \ref{fig:18wpass3D}).\label{a:3Dwpass}
\end{assumption}

\begin{figure}
    \centering
    \includegraphics[width=1\linewidth]{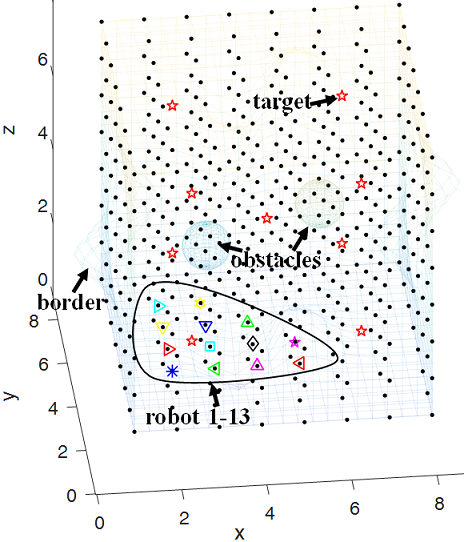}
    \caption[3D area and initial settings]{3D area and initial settings}
    \label{fig:413dinitialsetting}
\end{figure}

The initial state of simulations is demonstrated in Figure \ref{fig:413dinitialsetting} in which robots start from vertices near the boundary of the area. The colored net with curves is the border of the search area, and the two spheres inside are the obstacles. The vertices of the net are the discrete boundary points which are close enough so that robots will not pass the border. Passages between spheres and borders are designed to fit Assumption \ref{a:3Dwpass}. Black dots are the vertices of the cubic grid, and 13 robots are labeled with colored shapes in the black curve. Red stars are targets $t_{1}$ to $t_{9}$ evenly distributed in the space because the compared algorithm L\'evy fight works best for sparsely and randomly distributed targets.

To avoid collisions with both obstacles and borders in movements, the sensing range for obstacles $r_{so}$ satisfies:
\begin{assumption}
    $r_{so}\geq{a+r_{safe}}$ (see Figure \ref{fig:16range3d}).\label{a:3Drso}
\end{assumption}

If there is any obstacle on the way from $rob_{i}$ to one of its sensing neighbors, that neighbor will not be visited. Then, let $V_{a}$ represents the set which includes all the accessible vertices. Let the set of sensing neighbors $N_{s,i}(k)$ contains all the closest vertices $n_{s,i,j}(k)$ around $rob_{i}$ at discrete time $k$ in which $j\in\mathbb{Z}^{+}$ is the index of a neighbor robot. In $N_{s,i}(k)$, let the set containing visited vertices of $rob_{i}$ be $V_{v,i}(k)$ and the set containing unvisited vertices be $V_{u,i}(k)$. Then the choice of $rob_{i}$ from $V_{v,i}(k)$ is represented by $c_{v}$ and the choice from $V_{u,i}(k)$ is $c_{u}$. 
To avoid collisions resulting from having the same choice by different robots, robots choose the next steps in a certain order as described in section \ref{s83Dalgo} and at least need to tell the choice to those which are one vertex and two vertices away. Thus, the communication range $r_{c}$ satisfies:
\begin{assumption}
    $ r_{c}\geq{2*a+e}$(see Figure \ref{fig:16range3d}).\label{a:3Drc}
\end{assumption}

Robots which are in the range $r_c$ of $rob_{i}$ are called communication neighbors of $rob_{i}$. The communication uses ad-hoc networks, and all robots are in equal status. As the network is temporary, it will be rebuilt at the beginning of each loop. Let $ N_{c,i}(k)$ denotes the set of communication neighbors of $rob_{i}$ that it must communicate with. 

By having the concepts above, a further explanation about Figure \ref{fig:16range3d} is given. This figure demonstrates some parameters and the priority of making choices using a cubic grid centered at $rob_{i}$. The small sphere shows $r_{so}$ with 6 magenta asterisks ({\color{magenta}*}) as sensing neighbors. The larger sphere illustrates $r_{c}$ with 26 blue plus signs ({\color{blue}+}) and 6 magnet asterisks in it. Then, max($\lvert N_{c,i}(k) \rvert)=32$. Half of them (16 vertices) in the black circles (o) have higher priorities than $rob_{i}$ in making choice. 
\begin{assumption}
    The search task for a team of robots usually has a huge area. So, $a$ is much smaller than the length, the width and the height of area $A$. Therefore, the boundary effect is negligible which means there are no targets between the boundary and the frontiers of the target sensing spheres. \label{a:3Dboundaryeffect}
\end{assumption}
\begin{assumption}
    $r_{st}\geq{2a/\sqrt{3}+r_{safe}-r_{rob}}$ based on Table \ref{t3}.
\end{assumption}

The aim of the section is to design a collision-free decentralized algorithm for robots to search targets in a 3D area which satisfies the above assumptions using a cubic grid.
\subsection{Procedure and Algorithm}
\label{s83Dalgo}
The general procedure for the 3D task is similar to the 2D version. So, it uses the same rules for making choice (\ref{e52Dchoose}) 
\begin{equation}
p_{i}(k+1)=
\begin{cases}
c_{u} \text{ with prob. } 1/\lvert V_{u,i}(k)\rvert, \text{ if } \lvert V_{u,i}(k)\rvert\neq{0},\\
c_{v} \text{ with prob. } 1/\lvert V_{v,i}(k)\rvert, \text{ if } (\lvert V_{u,i}(k)\rvert=0)\&(\lvert V_{v,i}(k)\rvert\neq{0}),\\
p_{i}(k),  \text{ if } \lvert N_{s,i}(k) \rvert=0
\end{cases}
\label{e53Dchoose}
\end{equation}
and (\ref{e62DstopI} and \ref{e72DstopII}) for stop.
\begin{equation}
\text{In situation I: } p_{i}(k+1)=p_{i}(k)  \text{, if } \lvert T_{ki}\rvert=\lvert T\rvert
\label{e63DstopI}
\end{equation} 
\begin{equation}
\text{In situation : }p_{i}(k+1)=p_{i}(k)  \text{, if } \forall v_{i}\in V_{a}, \text{ } \lvert V_{u,i}(k)\rvert=0
\label{e73DstopII}
\end{equation}

But the way to set the selection sequence is different. The 2th step of flow chart \ref{fig:342dflowsitu1} and \ref{fig:352dflowsitu2} sets the order of choices as illustrated in Figure \ref{fig:16range3d}. Let $rob_{j}$ at $p_{j}$ be a communication neighbor of $rob_{i}$. Then if $p_{jx}>p_{ix}\mid(p_{jy}>p_{iy}\&p_{jx}\geq{p_{ix}})\mid(p_{jz}>p_{iz}\&p_{jy}\geq{p_{iy}}\&p_{jx}\geq{p_{ix}})$, $rob_{j}$ will have the higher priority. Thus in the figure, 16 vertices which is half of the max$\lvert N_{c,i}(k)\rvert$ in the black circles will have higher priorities than the middle robot $rob_{i}$. This will result in no contradiction from the views of different robots. 
 
\subsection{Simulation Results}
\begin{figure}
    \centering
    \includegraphics[width=0.65\linewidth]{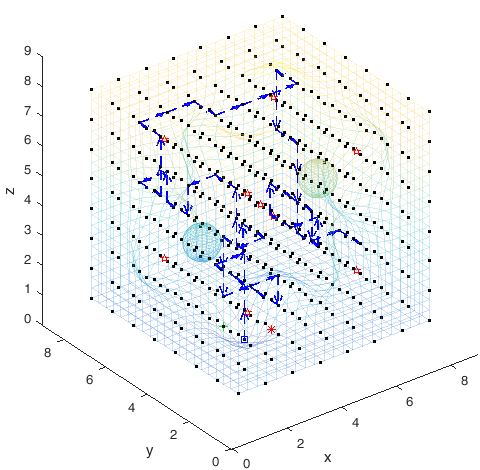}
    \caption[The route of robot 1 in a simulation of 3 robots]{The route of robot 1 in a simulation of 3 robots}
    \label{fig:423Droute1}
\end{figure}

\begin{figure}
    \centering
    \includegraphics[width=0.7\linewidth]{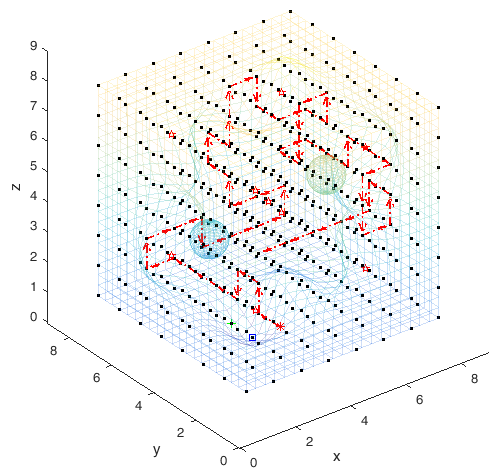}
    \caption[The route of robot 2 in a simulation of 3 robots]{The route of robot 2 in a simulation of 3 robots}
    \label{fig:433Droute2}
\end{figure}

\begin{figure}
    \centering
    \includegraphics[width=0.7\linewidth]{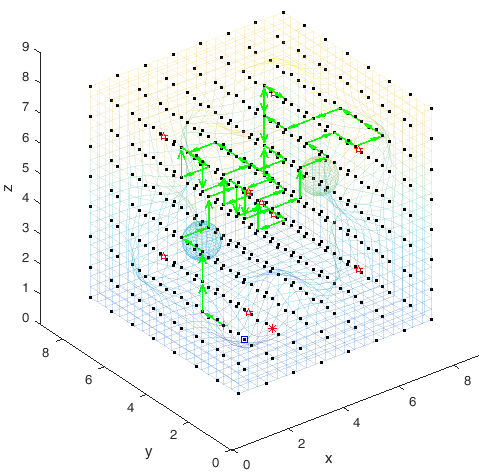}
    \caption[The route of robot 3 in a simulation of 3 robots]{The route of robot 3 in a simulation of 3 robots}
    \label{fig:443Droute3}
\end{figure}

\begin{figure}
    \centering
    \includegraphics[width=1\linewidth]{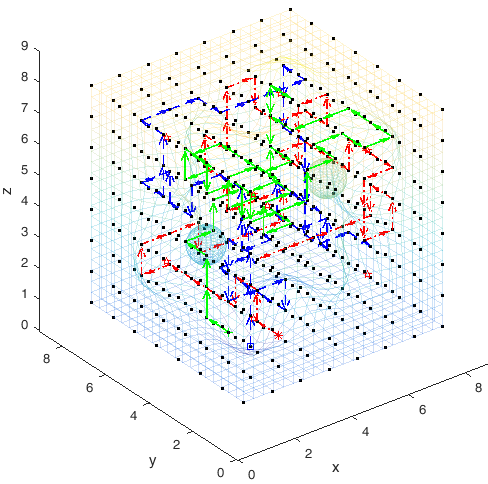}
    \caption[The combined routes of 3 robots]{The combined routes of 3 robots}
    \label{fig:453Droute123}
\end{figure}

The simulation of the algorithm is run in MATLAB2016a. To illustrate the detail in the route of robots apparently, the 8*8*8 grid in Figure \ref{fig:413dinitialsetting} is used with three robots label with red, blue and green color respectively. The route for situation I with 9 targets are shown in Figure \ref{fig:423Droute1}, \ref{fig:433Droute2} and \ref{fig:443Droute3} for the route of each robot and Figure \ref{fig:453Droute123} is the combination of the three figures. The routes are illustrated with arrows to show the directions of movements, and the route of each robot has a distinct color. The result indicates that each target was within $r_{st}$ of the sensing range of a visited vertex and there was no collision between robots based on the routes.

\begin{figure}
    \centering
    \includegraphics[width=0.75\linewidth]{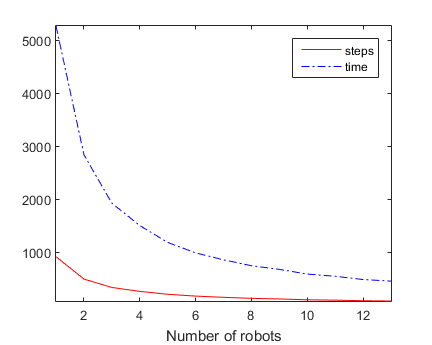}
    \caption[Time and steps for situation I]{Time and steps for situation I}
    \label{fig:803d1timestepi}
\end{figure}

\begin{figure}
    \centering
    \includegraphics[width=0.75\linewidth]{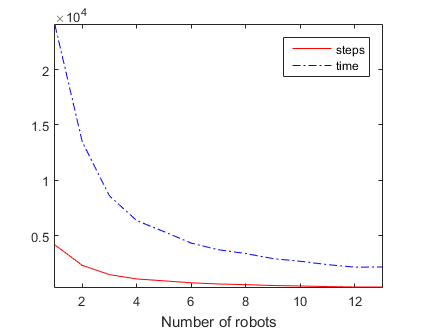}
    \caption[Time and steps for situation II]{Time and steps for situation II}
    \label{fig:813d1timestepii}
\end{figure}

Then, the algorithm is tested in a larger area which has 12*12*12 vertices in the grid and 1 to 13 robots are used which are still at the bottom layer of the area near point (1,1,1). Both situation I with Algorithm \ref{e62DstopI} and situation II with Algorithm \ref{e72DstopII} are considered and the simulations is run for 500 times to get the average search time. Figure \ref{fig:803d1timestepi} shows the result for situation I and Figure \ref{fig:813d1timestepii} demonstrate the result for situation II. Table \ref{t113Dalgo1result} is used to show the ratio of calculation time to total time for the two situations.

\begin{table}
    \centering
    \renewcommand\arraystretch{2}
    \begin{tabular}{|c|c|c|c|c|c|c|c|}
        \hline 
        No.&    1&    2&    3&    4    &5&    6&    7 \\ 
        \hline 
        Situation I&0.28\%&    0.31\%&    0.34\%&    0.37\%&    0.41\%&    0.49\%    &0.49\%  \\ 
        \hline 
        Situation II&0.79\%&    1.34\%&    1.56\%&    1.90\%&    2.27\%&    2.40\%&    2.99\%\\ 
        \hline 
        No.&8&9&10&11&12&13&\\
        \hline
        Situation I&    0.63\%    &0.56\%&    0.62\%    &0.66\%    &0.74\%    &0.76\%&    \\    
        \hline
        Situation II&    3.42\%&    3.54\%&    4.17\%&    4.67\%&    4.69\%&    5.67\%&\\
        \hline
    \end{tabular}  
    \caption[The ratio of calculation time to total time]{The ratio of calculation time to total time}
    \label{t113Dalgo1result}
\end{table}

The two figures illustrate that, as the number of robots increases, the amount of search time and the number of search steps decreases sharply at the beginning but slowly later. However, the time in wait step will increase as it equals the time for the choice of one robot multiplies the number of robots. However, the algorithm time only accounts for a small percentage ($\leq{5.7\%}$)of the total time, so the trend of the total time follows the trend of steps needed. As the percentage is increasing, it may need to be considered when there are a significant number of robots. When comparing the two figures, we can see that the time in situation II is more than four times of situation I. Thus, for time-sensitive tasks, it is better to spend some time to find the number targets if possible. 

The simulation shows that the algorithm is flexible with different numbers of robots and is scalable with areas in various sizes. However, other factors such as initial locations of robots, the number of robots, and the shape of the area are unknown in the problem discussed in this paper. So it is still hard to predict the time of a search task even with a fixed area. To choose the suitable number of robots, the user needs to seek a balance between the search time and the cost of robots. 
\subsection{Section Summary}
This section presented a decentralized search algorithm for a robot team in 3D area. A C grid is used to simplify the algorithm and avoid collisions with a selection order. In the algorithm, robots use local information and a random rule to explore the area with a reliable failure detection method. Performance of the algorithm is simulated for both situation I and situation II with 1 to 13 robots. The results show that the algorithm is flexible and scalable.

\section{Summary}
This chapter proposed the first decentralized random algorithm for search task using multiple mobile robots. The area which satisfies the assumptions is unknown with unknown arbitrary static obstacles and targets. Robots can move via a grid pattern without collisions. They start with a preset common grid pattern and the same coordinate system. With the assumed sensing ranges and the communication range, robots use the proved algorithm to choose future steps in order and move with no collision until the task is finished even if some robots failed. All assumptions are based on realistic conditions such as the volume and physical constraints of the robots. A rigorous mathematical proof of the convergence of the algorithm is provided. Simulation results show that, for the same task, more robots will use fewer steps and time. However, the time may stop decreasing when the number of robots increases to a certain value. Conclusively, the algorithm is scalable, flexible, and reliable. It is suitable for both 2D and 3D area and can be transferred between these two dimensions easily.
\chapter{A Collision-Free Random Search with the Repulsive Force}
\chaptermark{A Collision-Free Random Search with the Repulsive Force}
This chapter proposes the second grid based decentralized algorithm without collisions. Comparing to the 1st algorithm in Chapter 4, this algorithm still uses the random decision rule, but it adds the repulsive force of artificial potential field algorithm which improves the algorithm. The artificial potential field algorithm is used in many search and complete coverage problems such as \cite{RN262,RN15,RN300,RN301,RN302,RN296,RN55} as it is easy to apply. 

In this chapter, problem formulation and algorithm description will be simplified and refer to Chapter 4. Only the different parts will be emphasized. Similar works with grid algorithms, such as \cite{RN119,RN118,RN229,RN234,RN1,RN101,RN232,RN233}, will be compared. However, algorithms in those papers did not consider collision avoidance well or just ignored it. Other random algorithms are mainly inspired by animals such as random work or L\'evy flight related algorithms \cite{RN102,RN107,RN291,RN286,RN290,RN278,RN284}. They will also be compared. The factors related to search time will be discussed in depth for 2D areas. Also, experiments with Pioneer 3-DX robots are used to verify the algorithm in a 2D area. 
\section{2D Tasks}
This part proposed the second collision-free random algorithm for search task with potential field algorithm. The way to find the relation between the search time, the number of robots and the size of the area will be demonstrated. Other factors related to search time will also be discussed to have a deeper understanding of this algorithm. Unlike Chapter 4 which only analyzed the performance of the algorithm itself, the algorithm in this section will be compared to many other algorithms to show the effectiveness. Then an experiment using Pioneer 3-DX robots is done to check the performance of the algorithm in reality.
\subsection{Problem Statement}
In this section, the task area is a two-dimensional area $A$ with a limited number of static obstacles  $ O_{1} $, $ O_{2} $,$\ldots $, $O_{l} $. There are $m$ robots labelled as $rob_{1}$ to $rob_{n}$ to search static targets $t_{1}$ to $t_{n}$. An arbitrary known grid pattern with a side length of $a$ is used for robots to move in each step with the following definitions and assumptions. All these setting can be seen in Figure \ref{fig:712d2initialsetting}. Further explanations were given in Chapter 4. 
\begin{assumption}
    The area $A$ is a bounded, connected and Lebesgue measurable set. The obstacles $O_{i}$ are non-overlapping, closed, bounded and linearly connected sets for any $i\geq{0}$. They are both unknown to the robots. \label{a:2area}
\end{assumption}
\begin{definition}
    Let $O:=\bigcup{O_{i}}$ for all $i>0$. Then $A_{d}:=A\setminus{O}$ represents the area that needs to be detected.\label{d:2obs}
\end{definition}
\begin{assumption}
    The initial condition is that all robots use the same grid pattern and the same coordinate system. They start from different vertices near the boundary which can be seen as the only known entrance of the area.\label{a:2Initial_P}
\end{assumption}
\begin{assumption}
    All the ranges and radius in 2D search problems are circular. \label{a:2circular}
\end{assumption}
\begin{definition}
    To avoid collisions with other things before a move, a safe distance $ r_{safe} $ should include both $ r_{rob} $ and $e$ (see Figure \ref{fig:5r_ob}). So $ r_{safe}\geq{r_{rob}+e} $.\label{d:2rsafe}
\end{definition}
\begin{assumption}
    If the T grid is used, all passages $W_{pass}$ between obstacles or between an obstacle and a boundary are wider than $a+2*r_{safe}$. Other $W_{pass}$ for other grid patterns can be seen in Table \ref{t1}.\label{a:2wpass}
\end{assumption}
\begin{assumption}
    The sensing radius for obstacles is $r_{so}$ where $r_{so}\geq{a+r_{safe}}$.\label{a:2$r_{so}$}
\end{assumption}
\begin{assumption}
    $ r_{c}\geq{2*a+e}$ (see Figure \ref{fig:6rangeorder}). \label{a:2rc}
\end{assumption}

$V_{a}$ is the set of all the accessible vertices. $N_{s,i}(k)$ represents the set of sensing neighbors around $rob_{i}$. In $N_{s,i}(k)$, $V_{v,i}(k)$ and $V_{u,i}(k)$ denote the two sets for visited vertices and unvisited vertex and the choices from these two sets are $c_{v}$ and $c_{u}$ respectively.

If the strict requirement is used, 
\begin{assumption}
    The curvature of the concave part of the boundary should be smaller than or equal to $1/ r_{st} $.\label{a:2D2curvature}
\end{assumption} 
\begin{assumption}
    $ r_{st} \geq{a+r_{safe}}$.\label{a:2rst2strict}
\end{assumption}
If the loose requirement is used, the above two assumptions are not needed and following assumptions are needed.
\begin{assumption}
    The search area is usually vast. So, the side length $a$ is much smaller than the length and the width of area $A$. Therefore, the boundary effect is negligible. \label{a:2D2boundaryeffect}
\end{assumption} 
\begin{assumption}
    $ r_{st} \geq{a/\sqrt{3}+r_{safe}-r_{rob}}$.\label{a:2rst2loose}
\end{assumption}
 
The following definition is about variables in calculating repulsive force which is new. 
\begin{definition}
    In loop $k$, $p_{i,ave}(k)$ is defined as the average position of choices made by robots which chose earlier and positions of communication neighbors which have not chosen the next step yet. Then $ d_{c_{u},p_{i,ave}(k)} $ is the distance from $p_{i,ave}(k)$ to one unvisited vertex and $ d_{c_{v},p_{i,ave}(k)} $ is the distance from $p_{i,ave}(k)$ to a visited one. Then the maximum values of these two parameters about distance are represented as $ max(d_{c_{u},p_{i,ave}(k)})$ and $max(d_{c_{v},p_{i,ave}(k)}) $ separately. The corresponding sets of vertices which result in these maximum distances are represented as $ V_{umax,i}(k)$ and $ V_{vmax,i}(k)$. The chosen position of $rob_{i}$ at time $k+1$ is $p_{i}(k+1)$.
\end{definition}
\subsection{Procedure and Algorithm }
The algorithm directs the robots to explore the area gradually vertex by vertex. The general progress in the second search algorithm is the same as the first search algorithm and the flow charts in Figure \ref{fig:342dflowsitu1} and \ref{fig:352dflowsitu2} can still be used. The key difference is how to choose the next step. 

Choices are made using the following random decentralized algorithm with repulsive force which is improved based on Chapter 4. When a robot has no neighbor robots in $r_{so}$, Algorithm \ref{e82Dchoose2a} from Chapter 4 will be applied. It means, when there are vacant unvisited vertices, $rob_{i}$ will choose one from them randomly. If there are only visited sensing neighbors to go, $rob_{i}$ will randomly choose one from the visited vertices. If there is no vacant sensing neighbor around, $rob_{i}$ will stay at the current position $p_{i} (k)$.
\begin{equation}
p_{i}(k+1)=
\begin{cases}
c_{u} \text{ with prob. } 1/\lvert V_{u,i}(k)\rvert, \text{ if } \lvert V_{u,i}(k)\rvert\neq{0},\\
c_{v} \text{ with prob. } 1/\lvert V_{v,i}(k)\rvert, \text{ if } (\lvert V_{u,i}(k)\rvert=0)\&(\lvert V_{v,i}(k)\rvert\neq{0}),\\
p_{i}(k),  \text{ if } \lvert N_{s,i}(k) \rvert=0
\end{cases}
\label{e82Dchoose2a}
\end{equation}

When robots have neighbor robots in $r_{so}$, they apply the algorithm \ref{e82Dchoose2b} which uses the repulsive force of the artificial potential field. After receiving choices of previous robots, at time $k$, $rob_{i}$ will calculate $p_{i,ave}(k)$. Then it will randomly choose one vertex from those which result in $ max(d_{c_{u},p_{i,ave}(k)})$ if there are unvisited vertices to go, or from  $ max(d_{c_{v},p_{i,ave}(k)})$ if there are only visited vertices to visit. Under this circumstance, robots near each other will try to leave each other. So if there are many unvisited vertices which are within $r_{c}$ of robots at time $k$, robots may have more choices at time $k+1$ by using rule \ref{e82Dchoose2b} than using rule \ref{e82Dchoose2a}. If all the vertices around are visited, robots will try to leave this visited section in different directions simultaneously to have less repeated vertices comparing to using Algorithm \ref{e82Dchoose2a}. Thus, they may find unvisited sections earlier. These two algorithms ensure that all the places in $A_{d}$ are searched. 
\begin{equation}
p_{i}(k+1)=
\begin{cases}
c_{u} \text{ with prob. } 1/\lvert V_{umax,i}(k)\rvert, \text{ if } \lvert V_{u,i}(k)\rvert\neq{0},\\
c_{v} \text{ with prob. } 1/\lvert V_{vmax,i}(k)\rvert, \text{ if } (\lvert V_{u,i}(k)\rvert=0)\&(\lvert V_{v,i}(k)\rvert\neq{0}),\\
p_{i}(k),  \text{ if } \lvert N_{s,i}(k) \rvert=0
\end{cases}
\label{e82Dchoose2b}
\end{equation}

The stop strategy for this search algorithm is the same as that in Chapter 4. So, in situation I:
\begin{equation}
p_{i}(k+1)=p_{i}(k)  \text{, if } \lvert T_{ki}\rvert=\lvert T\rvert.
\label{e92DstopI}
\end{equation}
In situation II:
\begin{equation}
p_{i}(k+1)=p_{i}(k)  \text{, if } \forall v_{i}\in V_{a}, \text{ } \lvert V_{u,i}(k)\rvert=0.
\label{e92DstopII}
\end{equation}
\begin{theorem}
    Suppose that all assumptions hold and the random decentralized algorithms \ref{e82Dchoose2a} \& \ref{e82Dchoose2b} with a related judgment strategy \ref{e92DstopI} or \ref{e92DstopII} are used. Then for any number of robots, there is a discrete time $k_{0}>0$ such that all targets are found or all vertices are detected with probability 1.
    \label{th2D2}
\end{theorem}
\begin{prove}
    The algorithms \ref{e82Dchoose2a} \& \ref{e82Dchoose2b} with a corresponding judgment method form an absorbing Markov chain which includes both transient states and absorbing states. The transient states are the approachable vertices of the grid that robots visit but do not stop. Absorbing states are the vertices that the robots stop. Applying algorithms \ref{e82Dchoose2a} \& \ref{e82Dchoose2b}, robots tend to go to the unvisited neighbor vertices. If all sensing neighbors are visited before finding the targets, robots will randomly choose an accessible neighbor, continue to the next step and only stop when all targets are found, or all vertices are detected. For \ref{e92DstopI}, an absorbing state will be reached until all robots know $\lvert T_{ki}\rvert=\lvert T\rvert$. For \ref{e92DstopII}, an absorbing state will be reached until $\lvert V_{u,i}(k)\rvert=0$ for any vertex. As robot stop moving after the judgment conditions are satisfied, so absorbing states are impossible to leave.  The assumption about $W_{pass}$ guarantees that all the accessible vertices can be visited so absorbing states can be achieved from any initial states, namely with probability 1. This completes the proof of Theorem \ref{th2D2}
\end{prove}
\subsection{Simulations with the Strict Requirement}
\label{s9simustrict}
Simulations with algorithms \ref{e82Dchoose2a} \& \ref{e82Dchoose2b} for both situation I and II are run to verify the algorithm. For situation I, one to nine targets which are averagely allocated in the area are tested. 

\begin{figure}
    \centering
    \includegraphics[width=0.7\linewidth]{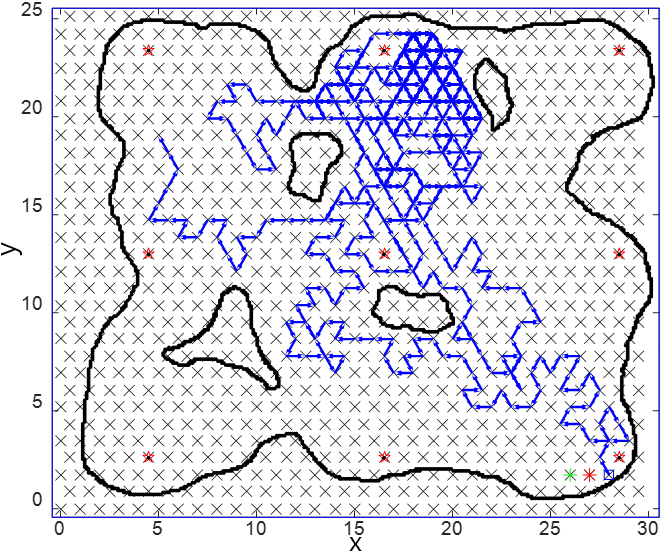}
    \caption[Route of robot 1 in a simulation of three robots]{Route of robot 1 in a simulation of three robots}
    \label{fig:462d2r1}
\end{figure}

\begin{figure}
    \centering
    \includegraphics[width=0.7\linewidth]{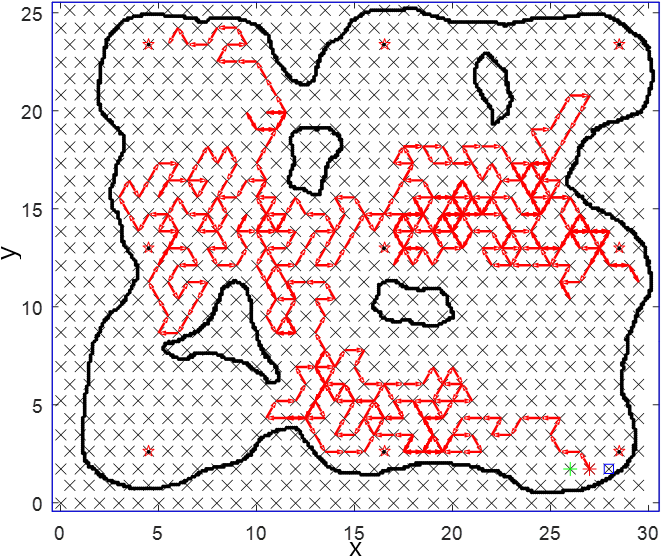}
    \caption[Route of robot 2 in a simulation of three robots]{Route of robot 2 in a simulation of three robots}
    \label{fig:472d2r2}
\end{figure}

\begin{figure}
    \centering
    \includegraphics[width=0.7\linewidth]{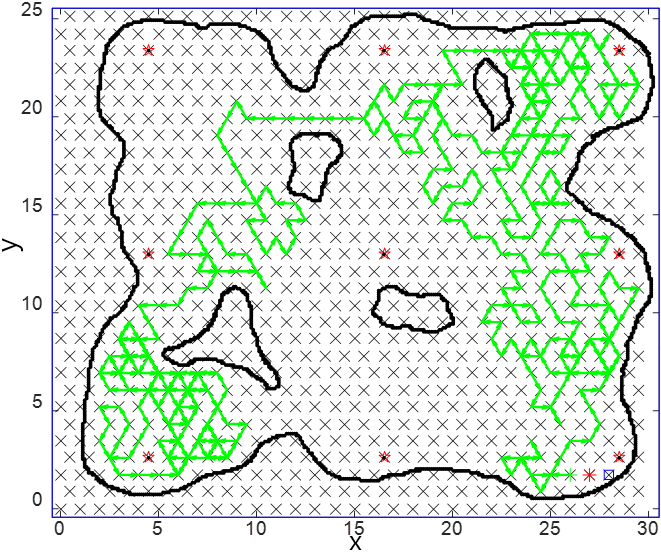}
    \caption[Route of robot 3 in a simulation of three robots]{Route of robot 3 in a simulation of three robots}
    \label{fig:482d2r3}
\end{figure}

\begin{figure}
    \centering
    \includegraphics[width=1\linewidth]{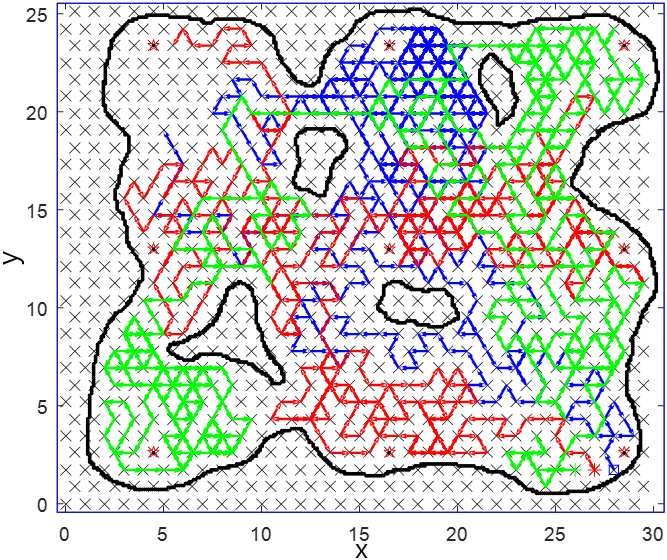}
    \caption[Combined routes in a simulation of three robots]{Combined routes in a simulation of three robots}
    \label{fig:492d2r4}
\end{figure}

For situation I, Figure \ref{fig:462d2r1}, \ref{fig:472d2r2}, \ref{fig:482d2r3} and \ref{fig:492d2r4} is an example of searching 9 targets by 3 robots. There are 900 vertices (30*30) in total in the graph. The boundaries of the search area and obstacles are illustrated by the bold black lines made up by discrete points. Distances between those points are designed to be small enough so robots cannot pass the boundaries. Table \ref{t5} shows the parameters in the simulation which are designed based on the datasheet of the Pioneer 3-DX robot. Robots had been one vertex away from each target which means every target is found.

\begin{figure}
    \centering
    \includegraphics[width=0.7\linewidth]{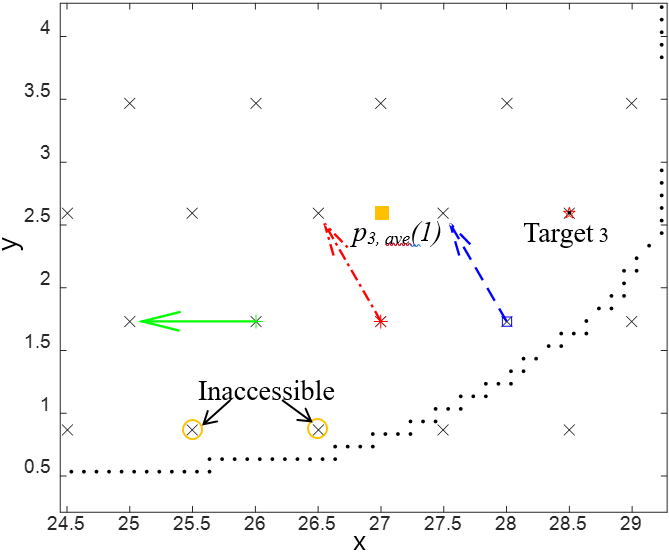}
    \caption[Choice in the first step of the simulation of three robots]{Choice in the first step of the simulation of three robots}
    \label{fig:502d2rinitial}
\end{figure}

Based on the recorded route, there were no collisions between robots. Figure \ref{fig:502d2rinitial} displays the initial step ($k=1$) of the above example. Orange circles are the vertices which cannot be accessed as they are less than $r_{safe}$ to the boundary points. Based on the method to decide selection order in Section \ref{algo2Dchoose}, the robot labeled by the blue square was the first to choose, the red one was the second, and the green one was the last. Then based on the first condition of the algorithm \ref{e82Dchoose2b}, the blue robot chose the unvisited vertex at the top left corner which is the only unvisited vertex in the six sensing neighbors. The vertex with target $t_{3}$ can be detected as an obstacle, so it does not need to be visited. Similarly, the red robot chose the vertex on the top left corner based on the first condition in \ref{e82Dchoose2b}. For the green robot, there were two accessible unvisited neighbors so it calculated $p_{3,ave}(k)$ which is the orange square and chose the vertex which results in $ max(d_{c_{u},p_{i,ave}(k)})$ , namely the vertex on the left side. 

\begin{figure}
    \centering
    \includegraphics[width=1\linewidth]{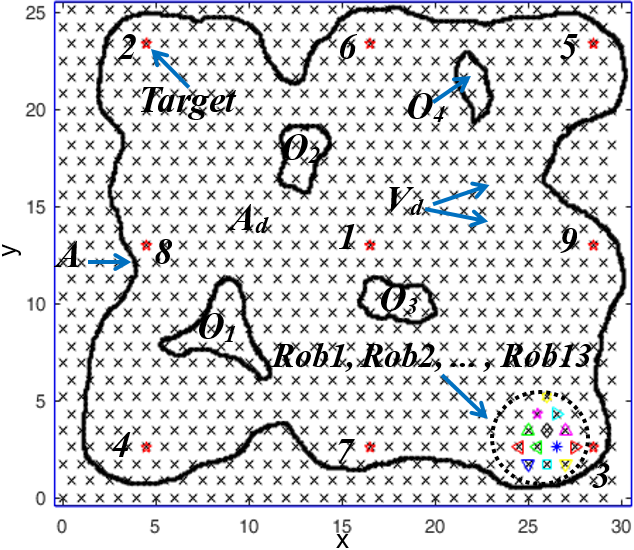}
    \caption[Initial setting of the map]{Initial setting of the map}
    \label{fig:712d2initialsetting}
\end{figure}

\begin{figure}
    \centering
    \includegraphics[width=0.7\linewidth]{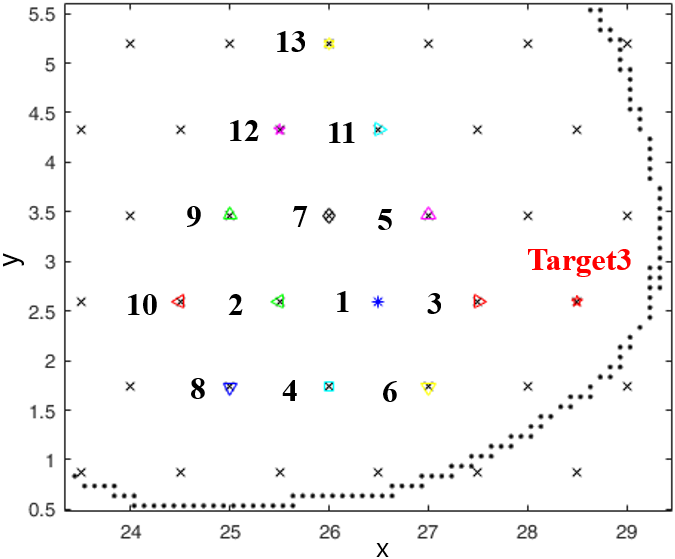}
    \caption[The order and initial positions of robots]{The order and initial positions of robots}
    \label{fig:512d2allocate}
\end{figure}

Search time may be affected by many factors including the number of robots, the size of the area, the number of targets, the initial allocation of robots and $W_{pass}$. The result of each situation is the average result for 150 simulations. In order to test every situation in rules \ref{e82Dchoose2a} \& \ref{e82Dchoose2b}, at least seven robots are needed to guarantee one robot can be blocked by six neighbors, which is the third condition of Algorithm \ref{e82Dchoose2b} and can be seen in Figure \ref{fig:712d2initialsetting}. The six neighbors are in the first and second conditions of Algorithm \ref{e82Dchoose2b} and other conditions will happen in later loops. In simulations, 1 to 13 robots were used, and the order of allocating robots is shown in Figure \ref{fig:512d2allocate}.

\begin{figure}
    \centering
    \includegraphics[width=0.7\linewidth]{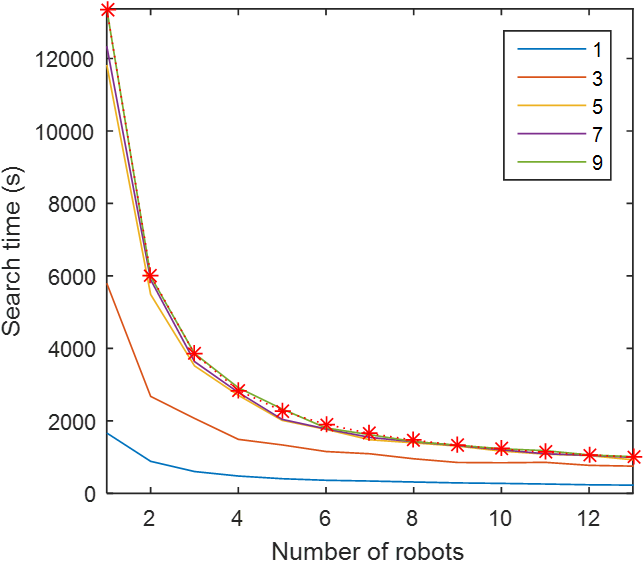}
    \caption[Search time for 1-13 robots with 1, 3, 5, 7 and 9 targets]{Search time for 1-13 robots with 1, 3, 5, 7 and 9 targets}
    \label{fig:52trob}
\end{figure}

Initially, simulations for one to thirteen robots with one, three, five, seven and nine targets in an area of 574.8243618m$^{2}$ (900 vertices) are run to check the relation between the search time and the number of robots as well as between the search time and the number of targets (see Figure \ref{fig:52trob} ). Here, 900 vertices include all the vertices in the graph, and this representation is clear to identify the relation between different areas. The size of the area is for the search area only which does not include the parts occupied by obstacles. The area and obstacles have random shapes with both concave and convex parts, and they satisfy the assumptions for the strict requirement. Thus, the result in Figure \ref{fig:52trob} can be thought as a general result for areas with obstacles. For each curve in Figure \ref{fig:52trob}, as the number of robots rises, the search time falls fast for 1 to 6 robots and slowly for 7 to 13 robots. The shape of the line is similar to an inverse proportional function and generally can be written in the form of the power function.
\begin{equation}
time=b*m^{c}+d \label{e10general} 
\end{equation}
where $b$, $c$ and $d$ are the unknown parameters and $m$ is the number of robots. As the search time for nine targets is compared with others in the next few paragraphs, the function for that line is estimated using the least squares method with four significance bits for accuracy, namely 
\begin{equation}
time=1295*10*m^{-1.209}+420.2. \label{e11tom}
\end{equation}

This function is the red dotted line with asterisks in \ref{fig:52trob}. When searching different numbers of targets using the same number of robots, the search time increases with the number of targets. However, it can be seen that time differences between five, seven, and nine targets are smaller than that between one, three and five targets as robots need to go to most parts of the area to find all the targets when there are five, seven, and nine targets. This is because targets $t_{2}$, $t_{3}$, $t_{4}$ and $t_{5}$ have taken up the four corners of the area. 

\begin{figure}
    \centering
    \includegraphics[width=0.7\linewidth]{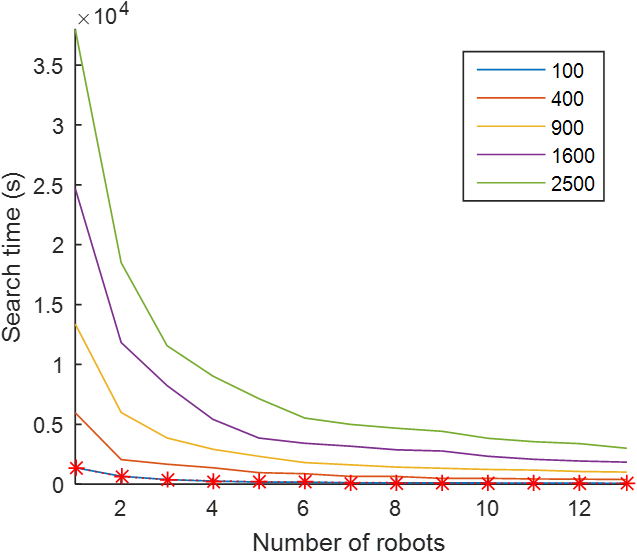}
    \caption[Search time for 9 targets with 1-13 robots in 5 areas]{Search time for 9 targets with 1-13 robots in 5 areas}
    \label{fig:53tarea}
\end{figure}

\begin{figure}
    \centering
    \includegraphics[width=0.7\linewidth]{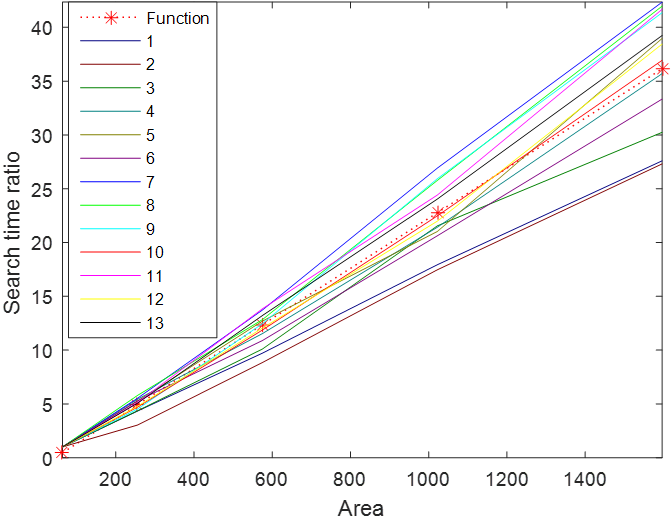}
    \caption[Time ratio based on 100 vertices for 1-13 robots in five areas]{Time ratio based on 100 vertices for 1-13 robots in five areas}
    \label{fig:54ratio}
\end{figure}

Secondly, 1 to 13 robots with 9 targets in areas of 63.87m$^{2}$ (100 vertices), 255.48m$^{2}$ (400 vertices), 574.82m$^{2}$ (900 vertices), 1021.91m$^{2}$ (1600 vertices) and 1596.73m$^{2}$ (2500 vertices) are tested to discover the relation between the search time, the number of robots (Figure \ref{fig:53tarea}) and the size of the area (Figure \ref{fig:54ratio}). It should be noted that these five areas are expanded or shrunk based on the 900 vertices area, so they have the same shape so the area ratio is the same as vertices ratio which is 1:4:9:16:25. All robots start from the bottom right corner near target $t_{3}$, so the starting points are controlled to have the minimal effect on the result. The curvatures are initially designed to satisfy the requirement in the area with 900 vertices, so the curvatures in the expanded areas also meet this requirement. Nonetheless, for the shrunk area, curves of some parts do not meet the requirement, so the grids are slightly moved to guarantee a complete search. The result in Figure \ref{fig:53tarea} agrees with Figure \ref{fig:52trob}. It also illustrates that as the size of area increases, the search time increases as well. If the search time for 100 vertices is considered as 1 for each number of robots, the ratio of search time in different areas to the search time in the area with 100 vertices can be demonstrated by Figure \ref{fig:54ratio}.  It can be claimed that there is always a linear relation between the search time and the area. The gradients of the lines are different, and there is no fixed relation between the gradient and the number of targets. So the average trend of the 13 lines is calculated using the least squares method and four significance bits, 
\begin{equation}
ratio=area*0.02328-1.024.  \label{e12ratio}
\end{equation}
This equation is the red dotted line in Figure \ref{fig:54ratio}. The base of this equation namely the time function for 100 vertices is
\begin{equation}
time=1401*m^{-1.146}-10.35. \label{e13base}
\end{equation}
It is the red dotted line in Figure \ref{fig:53tarea}. So, the general time function should be the multiplication of ratio \ref{e12ratio} and time \ref{e13base}:
\begin{equation}
time_{total}=32.62*m^{-1.146}*area-1435*m^{-1.146}-0.2410*area+10.60 \label{e14total}
\end{equation}                                                                                

\begin{figure}
    \centering
    \includegraphics[width=0.7\linewidth]{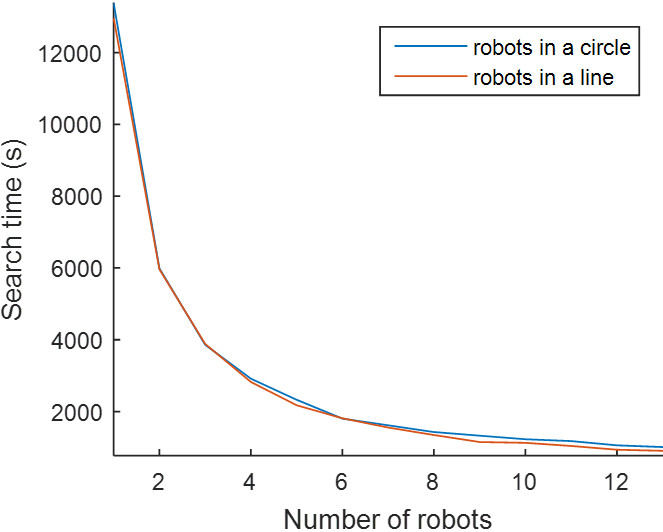}
    \caption[Search 9 targets in 900 vertices area with different allocations in situation I]{Search 9 targets in 900 vertices area with different allocations in situation I}
    \label{fig:55circleline}
\end{figure}

\begin{figure}
    \centering
    \includegraphics[width=0.7\linewidth]{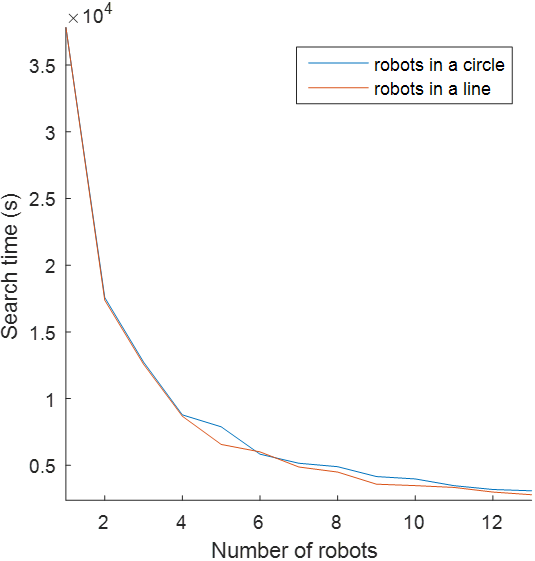}
    \caption[Search 9 targets in 900 vertices area with different allocations in situation II]{Search 9 targets in 900 vertices area with different allocations in situation II}
    \label{fig:56circleline2}
\end{figure}

\begin{figure}
    \centering
    \includegraphics[width=0.7\linewidth]{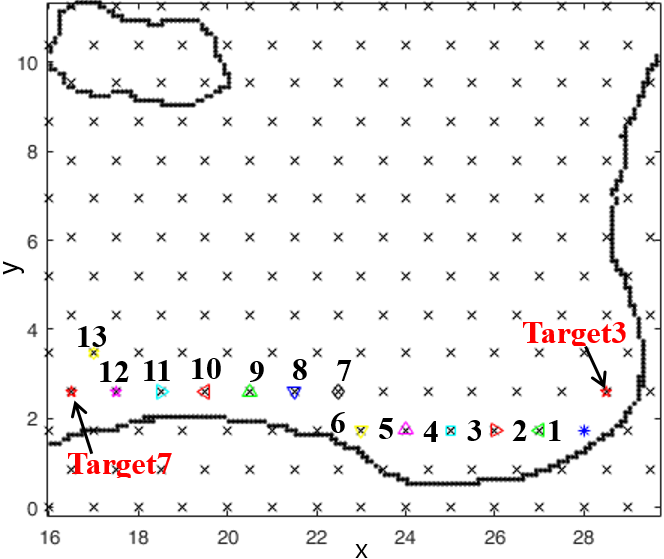}
    \caption[The order of allocation for 13 robots in a curve]{The order of allocation for 13 robots in a curve}
    \label{fig:57line}
\end{figure}

Thirdly, search time of two kinds of allocations for 1 to 13 robots is compared in the area of 900 vertices. Robots are allocated in a circle (Figure \ref{fig:502d2rinitial}) and in a curve (Figure \ref{fig:57line}) respectively. The results are displayed in Figure \ref{fig:55circleline} and \ref{fig:56circleline2} for situation I and II. For Figure \ref{fig:57line}, robots still start from the bottom right corner of the searching area, and every robot is in the obstacle sensing range of at least one robot to ensure that robots can communicate with each other and the operator does not need to move too far to deploy all robots. In both Figure \ref{fig:55circleline} and \ref{fig:56circleline2}, for 1-4 robots, the search time are close since the allocations for 1 to 3 robots are similar in a line near target 3 in both situations. For 5-13 robots, search time for robots in a circle is more than robots in a line. The reason is that robots are easier to block each other when allocated in a circle. Thus, some robots, for example, $rob_{1}$ and $rob_{2}$, need to choose a visited neighbor vertex even in the first step. In the next few steps, a robot may also repeat others’ choices, which is a waste of the resource. The deployment in a line decreases the chance of repeating vertices in the first few steps. It can be predicted that, if the distances between robots can be set larger but less than the communication range, robots will repeat fewer vertices in the first few steps as they could not only communicate to get explored map from others but also have more unvisited vertices around. Thus, robots could spend less time in the search task.


The widths of passages in the area also affect the search time. If narrow passages with widths near $W_{pass}$ are used to connect subareas, robots in one subarea will have a lower chance to move through the passage and leave that subarea. Thus, longer search time is needed compared to those using wide passages in a search area with all other three factors controlled to be the same.

In the proposed algorithm, the number of robots is flexible, and the algorithm is scalable. As the area is unknown, the function of search time may be estimated using the least squares method for simulation results with all possible known information and a long practical distance between robots in a line for initialization. Even for a known area, the optimal number of robots should still be decided by comparing the benefits from the decline of time with the cost of adding more robots.

\subsection{Comparison with Other Algorithms}
\label{s102Dcompare}
The proposed algorithm is also compared with other decentralized random algorithms for search task in \cite{RN101,RN102} in situation I. For \cite{RN101}, there are three methods using the T grid which are the \ac{R} algorithm, the \ac{RU} algorithm and the \ac{RUN} algorithm. However, in RUN, robots may not move straightly along the edges of triangles resulting to problems in path planning and collision avoidance. For example, how to use the repulsive force to plan the path when there are obstacles and other robots in the path and when and where to stop if the time for the movement step is used up without reaching the destination. But \cite{RN101} does not provide solutions for those problems. Therefore, this algorithm is unconvincing and could not be simulated so RUN is not compared with the proposed algorithm. In \cite{RN102}, the \ac{RW} algorithm, the \ac{LF} algorithm and the \ac{LFP} algorithm are discussed which are all compared with the proposed algorithm. In those algorithms, a normal distribution is needed to generate turning angles. The parameters of the normal distribution are set as $\mu=0$ and $\sigma=1$ and the angle range is $(-\pi,\pi]$. The step length of the RW algorithm will also be generated by the normal distribution with $\mu=a/2$, $\sigma=1$ and a maximum step length $a$. The L\'evy distribution is utilized to generate the lengths of movements. Based on \cite{RN107}, it needs two independent random variables, $f$ and $g$, generated by standard normal distribution above for angles to get the medium result $h$ where
\begin{equation}
h=\dfrac{f}{\lvert g\rvert^{1/\alpha}}\label{e15levy1}
\end{equation}
Then the sum of $h$ with an appropriate normalization 
\begin{equation}
z_{n}=\dfrac{1}{n^{1/\alpha}} \sum_{k=1}^{n}h_{k}\label{e16levy2}
\end{equation}
converges to the L\'evy distribution with a large $n$ which is 100 usually so $n$=100 is used. The constant $\alpha$ is a parameter in the L\'evy distribution function ranges from 0 to 2 which is 1.5 in the simulation. When $\alpha$=2, the distribution will become the Gaussian distribution. The repulsive force is needed to disperse robots in LFP which has the following formula according to [40]. 
\begin{equation}
\overrightarrow{F_{rep}}=-\nabla U_{rep}(\boldsymbol{q})=
\begin{cases}
k_{rep}(\dfrac{1}{\rho(\boldsymbol{q})}-\dfrac{1}{\rho_{0}})\dfrac{1}{\rho^{2}(\boldsymbol{q})}\dfrac{\boldsymbol{q}-\boldsymbol{q_{obstacle}}}{\rho(\boldsymbol{q})}\text{, if } \rho(\boldsymbol{q})\leq{\rho_{0}}\\
0 \text{, if } \rho(\boldsymbol{q})>{\rho_{0}}
\end{cases}
\label{e17levy3}
\end{equation}
The repulsive force $F_{rep}$ is used as the weight of the velocity in a certain direction. $U_{rep}(\boldsymbol{q})$ is a differentiable repulsive field function. $k_{rep}$ is the scale factor. Without losing generality, it is chosen as $k_{rep}=1$. $\rho_{0}$ is the maximum distance to apply the force and $\rho(\boldsymbol{q})$ is the minimal distance from a robot to its nearest obstacle. $\boldsymbol{q}=(x,y)$ represents the coordinate of the robot and $\boldsymbol{q_{obstacle}}$ is the coordinate of the obstacle. 

Algorithms R, RU, RW and LF did not provide any solution for collisions between robots. The applied strategy in the simulation is allowing collisions happen and counting the number of collisions. It is assumed that one more loop is needed to finish the search for each collision. In fact, if consecutive collisions happen, more than one extra loops may be required.

\begin{figure}
    \centering
    \includegraphics[width=1\linewidth]{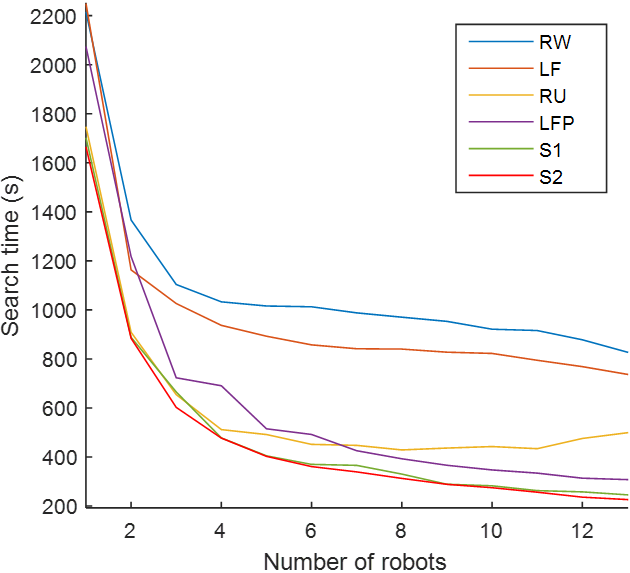}
    \caption[Search time for 1 target using 6 algorithms]{Search time for 1 target using 6 algorithms}
    \label{fig:58compare1tar}
\end{figure}

\begin{figure}
    \centering
    \includegraphics[width=0.7\linewidth]{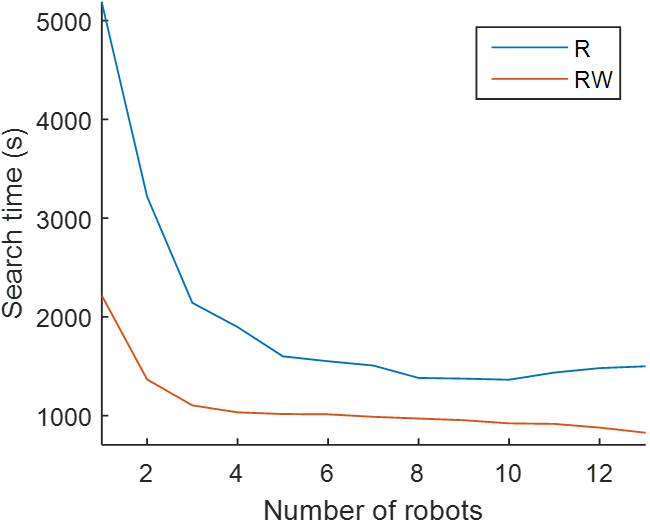}
    \caption[Search time for 1 target using algorithms R and RW]{Search time for 1 target using algorithms R and RW}
    \label{fig:59comparerrw}
\end{figure}

\begin{figure}
    \centering
    \includegraphics[width=1\linewidth]{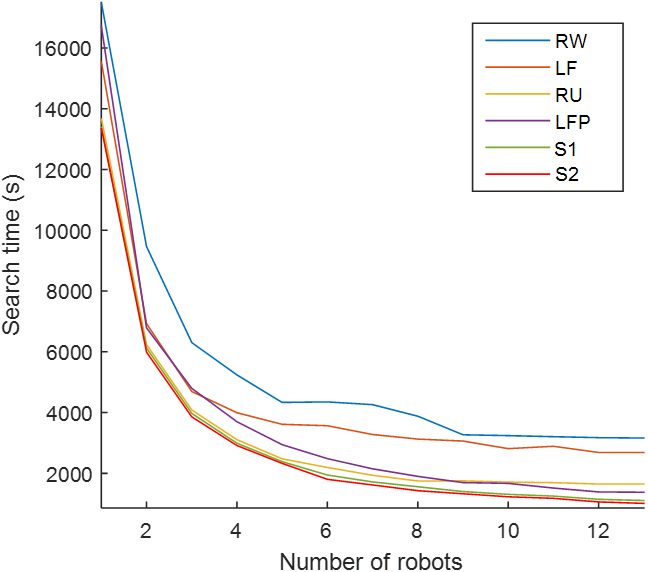}
    \caption[Search time for 9 targets using 6 algorithms]{Search time for 9 targets using 6 algorithms}
    \label{fig:60comparetar9}
\end{figure}

\begin{figure}
    \centering
    \includegraphics[width=0.7\linewidth]{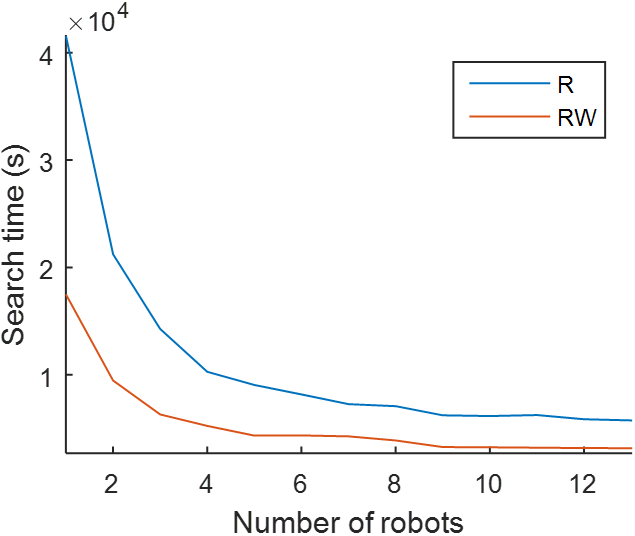}
    \caption[Search time for 9 targets using algorithms R and RW]{Search time for 9 targets using algorithms R and RW}
    \label{fig:61comparetar9rrw}
\end{figure}

To compare all methods in the same environment, all the algorithms are simulated in the area in Figure \ref{fig:712d2initialsetting} with one (Figure \ref{fig:58compare1tar} and \ref{fig:59comparerrw}) and nine targets (Figure \ref{fig:60comparetar9} and \ref{fig:61comparetar9rrw}) separately. The results in the four figures are the average result of at least 100 simulations. Nevertheless, \cite{RN101} had contradicted descriptions in the method of transmitting information of targets and \cite{RN102} used targets as stations to store and transmit information. These are different from the proposed method in this paper. So, to control all the factors to be the same, equation \ref{e92DstopI} for stop judgment, ad-hoc with $r_{c}$ for the communication and broadcasting for transmitting positions of targets are applied in all the algorithms. The algorithm R is much slower than the others. To see the small time difference between other methods, Figure \ref{fig:58compare1tar} and Figure \ref{fig:60comparetar9} compare all algorithms except R for one and nine targets separately. Time for R is only displayed in two additional figures, namely Figure \ref{fig:59comparerrw} with one target and Figure \ref{fig:61comparetar9rrw} with nine targets, with the second slowest algorithm RU to show how slow it is. \cite{RN102} claimed that LF algorithm and LFP algorithm are advantageous when targets are sparsely and randomly distributed. Therefore, the nine evenly separated targets in the area allow those two algorithms to fully unfold their potential, which makes the comparison more convincible.

In the line chart, the first algorithm in Chapter 4 is represented as S1 and the second search algorithm in this chapter is labeled as S2. The four line charts show that the proposed algorithm S2 is always the best under the strict requirement. The general ranking of search time from the graph is S2$<$S1$<$LFP$<$RU$<$LF$<$RW$<$R. For all the algorithms, as the number of robots ascended, the search time descended significantly at the beginning but gently later because more time will be spent on repeating visited vertices and collision avoidance instead of exploring new vertices only. For R and RU, although the number of loops decreases, the number of collisions grows and the time on dealing with collisions accounts for a larger proportion of the total time. This leads to slower descending search time or even ascending time as RU in Figure \ref{fig:58compare1tar} and R in Figure \ref{fig:59comparerrw}. The detail of time and collisions for RU in Figure \ref{fig:58compare1tar} is shown in Table \ref{t122D2RUresult}.  In that table, all values are average values for one simulation. The row `collision' shows the number of collisions and the row `$t_{c}$' is the time to deal with collisions. The `total time' includes $t_{c}$, algorithm time and movement time.  In RW, LF, and LFP, when there are more robots, robots will have higher chance to encounter others. Thus, they will turn more frequently without fully executing the translation part of the algorithm so that the efficiency of the algorithm will decline resulting in a more moderate decrease in the search time. As to the proposed algorithm in Table \ref{t132D2(1)(2)result}, `$t_{a}$' represents the average time used in the algorithm in a simulation, `$t_{a}$/loop' is the $t_{a}$ for each loop. To avoid collisions, the time in waiting will increase with the rise in the number of robots, because the wait time in each loop equals the number of robots multiplies the time for choosing. Therefore, $t_{a}$/loop goes up as the number of robots increases. However, $t_{a}$ only consists a small proportion of the total time even in the simulation with 13 robots (0.5396\%) so the trend of the total time is similar to the trend of the number of loops. However, if there are hundreds of robots in real applications, this proportion will increase to a level that needs to be considered.
\begin{table}
    \centering
    \renewcommand\arraystretch{2}
    \begin{tabular}{|c|c|c|c|c|c|c|c|}
        \hline 
        No.&    1&    2&    3&    4    &5&    6&    7 \\ 
        \hline 
        Loop&303&    143    &110&    81.1    &74.8&    63.6&    56.3 \\ 
        \hline 
        Collisions&0.00&    1.32&    3.74&    7.67&    10.5&    14.8&    21.4 \\ 
        \hline 
        $t_{c}$(s)&0.00&    7.56    &21.5&    44.1&    60.2&    84.9&    123\\ 
        \hline
        Total time (s)&1748&    833&    655&    512&    492&    452    &448 \\ 
        \hline
        No.&8&9&10&11&12&13&\\
        \hline
        Loop&    53.4&    49.1&    48.2&    43.3&    42.9&    42.3&    \\    
        \hline
        Collisions&    21.1&    26.7&    28.7&    32.0&    36.9&    42.1&\\
        \hline
        $t_{c}$(s)&121&    153    &165&    184    &212&    242&\\ 
        \hline
        Total time (s)&429&    437    &443&    434&    476    &499 &\\ 
        \hline
    \end{tabular}  
    \caption[Results of RU for 1-13 Robots in 900 vertices area with 1 target. (RU in Figure \ref{fig:58compare1tar})]{Results of RU for 1-13 Robots in 900 vertices area with 1 target. (RU in Figure \ref{fig:58compare1tar})}
    \label{t122D2RUresult}
\end{table}
\begin{table}
    \centering
    \renewcommand\arraystretch{2}
    \begin{tabular}{|c|c|c|c|c|c|c|c|}
        \hline 
        No.&    1&    2&    3&    4    &5&    6&    7 \\ 
        \hline 
        $t_{a}$(s)&    36.62&    17.71&    12.48&    10.88&    8.00&    6.92&    6.95 \\ 
        \hline 
        $t_{a}$/loop (s)    &0.0161&    0.0171&    0.0182&    0.0192&    0.0207&    0.0220&    0.0231 \\ 
        \hline 
        Total time (s)&    13117&    5989&    3965&    2857&    2279&    1864    &1589 \\ 
        \hline
        No.&8&9&10&11&12&13&\\
        \hline
        $t_{a}$&    6.08&    5.83&    5.50&    5.71&    5.53&    5.59&    \\    
        \hline
        $t_{a}$/loop (s)&0.0243&    0.0270&    0.0270&    0.0284&    0.0302&    0.0312&\\
        \hline
        Total time (s)&1449    &1337&    1204    &1154&    1076&    993&\\ 
        \hline
    \end{tabular}  
    \caption[Results of S2 for 1-13 Robots in 900 vertices area with 9 targets. (\ref{e82Dchoose2a} \& \ref{e82Dchoose2b} in Figure \ref{fig:60comparetar9})]{Results of S2 for 1-13 Robots in 900 vertices area with 9 targets. (\ref{e82Dchoose2a} \& \ref{e82Dchoose2b} in Figure \ref{fig:60comparetar9})}
    \label{t132D2(1)(2)result}
\end{table}

\subsection{Robot Test}
\begin{figure}
    \centering
    \includegraphics[width=1\linewidth]{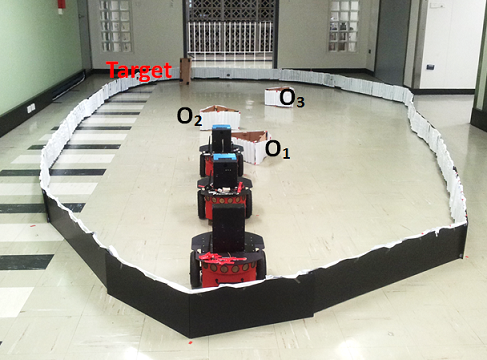}
    \caption[The test area]{The test area}
    \label{fig:62exp2}
\end{figure}

Algorithm \ref{e82Dchoose2a} \& \ref{e82Dchoose2b} in situation I is verified by one to three Pioneer 3-DX robots (Figure \ref{fig:10robot}) in a 429mm*866mm (maximum width*maximum length) area as seen in Figure \ref{fig:62exp2}. The robot has 16 sonars around to cover 360 degrees for objects with a height of 15.5cm-21.5cm and Figure \ref{fig:11sonar} displays half of them. The robot also has a SICK-200 laser to scan 180 degrees in the front (see Figure \ref{fig:103laser}) for objects higher than 21.5cm. Based on the difference of the heights, the test environment is set as Figure \ref{fig:62exp2}. The three large triangular boxes are obstacles. Obstacles and boundaries are set lower enough to be detectable for sonars only. Then a tall brown triangle box at the left side of the end was set as the target and can be detected by both sonars and the laser. Robots use Wi-Fi in ad-hoc mode to communicate with others using \ac{TCP}. 

Parameters of the robot used in the test are in Table \ref{t14exppara}. The safety distance $d_{safe}$ is calculated based on the odometry errors of robots. Based on the measurement, if a robot moves forward straightly for 8m, it may shift 0.08m to the left or right at most. As the moving distance from the bottom to the top of  Figure \ref{fig:62exp2} is less than 8m, $d_{safe}$=0.09m is large enough for collision avoidance even if there are errors. Note that the $r_{so}$ in the test is different from simulations as the used sonars have a minimum target sense range min$(r_{so})$ which is around 0.17m. Objects within this range will be considered as having an infinite distance from the robot. Therefore, the way to calculate $r_{so}$ is changed to the equality condition of the inequality $r_{so}\geq{a+ \text{min}(r_{so})+r_{safe}}$.

\begin{table}
    \centering
    \renewcommand\arraystretch{2}
    \begin{tabular}{|c|c|c|c|c|c|}
        \hline 
        P&$    \underline{r_{rob}\text{ (m)}} $&$    \underline{r_{safe}\text{ (m)}}$ &$    \underline{r_{so}\text{ (m)}} $&$    \underline{r_{st}\text{ (m)}} $&$    a (m)$  \\ 
        \hline 
        R&    0.26m&0.35m&5m&32m    &4.65m\\ 
        \hline 
        S&0.26&0.35&1.44&1.35&1 \\ 
        \hline 
        P&$    \underline{r_{c}\text{ (m)}}$&$    \underline{v_{max}\text{ (m/s)}}$& $    \underline{\omega_{max}\text{ (\SI{}{\degree}/s)}}$&$    t_{move} \text{ (s)}$&$    W_{pass}\text{ (m)} $\\
        \hline
        R&91.4m&1.5m/s&$\SI{300}{\degree}$/s&5.75s&5.35m\\
        \hline
        S&2.09&0.3/s&300&10s&1.7\\
        \hline
    \end{tabular} 
    \caption[Parameters of robots and in experiment]{Parameters of robots and in experiment}
    \label{t14exppara}
\end{table}

\begin{figure}
    \centering
    \includegraphics[width=0.7\linewidth]{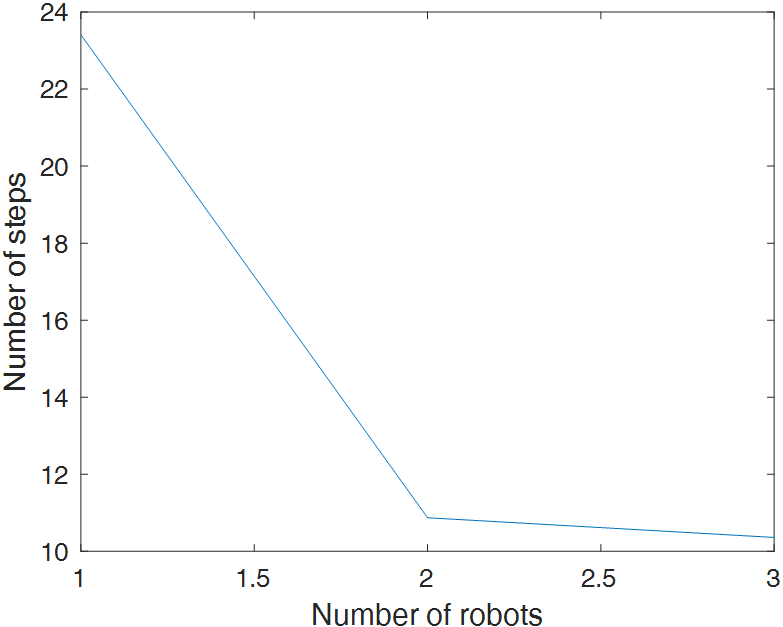}
    \caption[The average result for 10 tests with 1-3 robots]{The average result for 10 tests with 1-3 robots}
    \label{fig:63expresult}
\end{figure}

\begin{figure}
    \centering
    \includegraphics[width=1\linewidth]{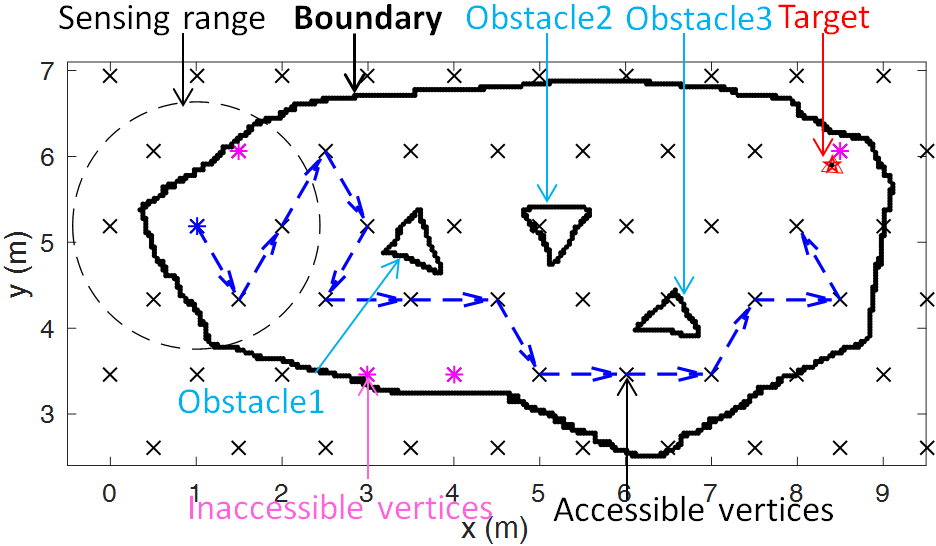}
    \caption[The example route for one robot with legends (13 loops)]{The example route for one robot with legends (13 loops)}
    \label{fig:64expr1}
\end{figure}

\begin{figure}
    \centering
    \includegraphics[width=0.7\linewidth]{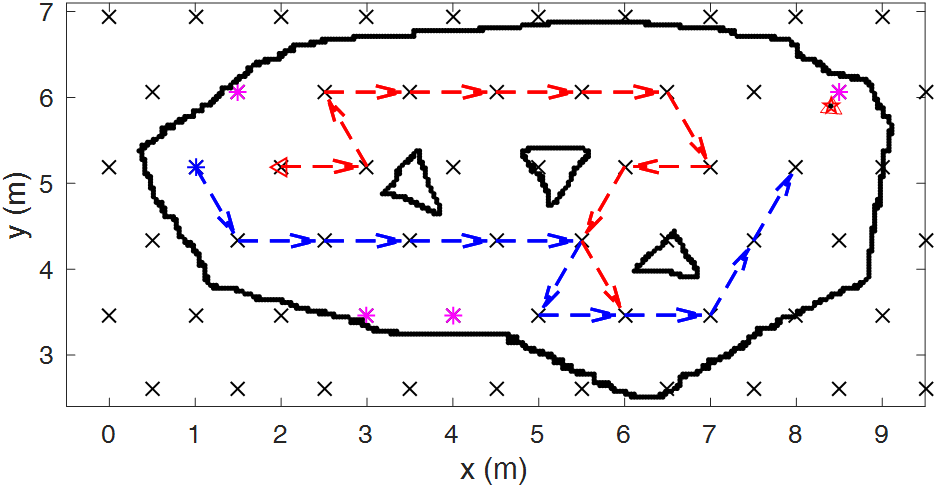}
    \caption[Example routes for two robots (10 loops)]{Example routes for two robots (10 loops)}
    \label{fig:65expr2}
\end{figure}

\begin{figure}
    \centering
    \includegraphics[width=0.7\linewidth]{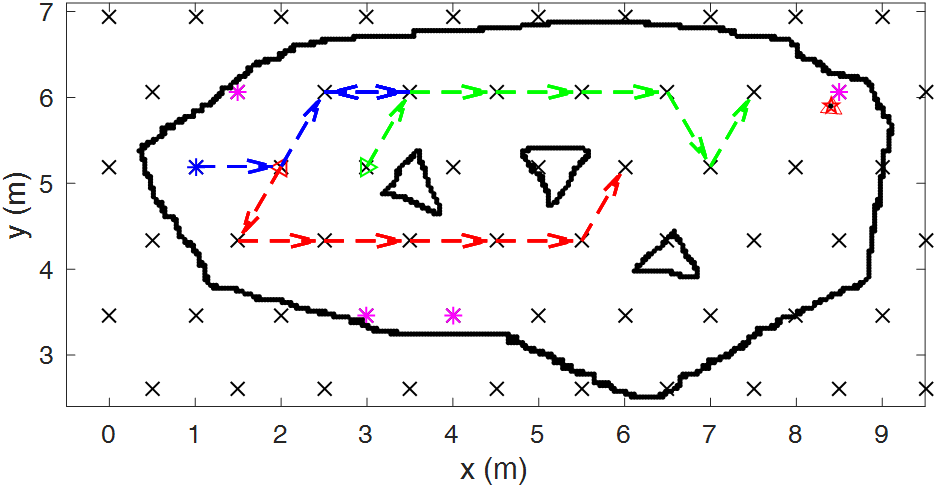}
    \caption[Example routes for three robots (6 loops)]{Example routes for three robots (6 loops)}
    \label{fig:66expr3}
\end{figure}

The test is done for ten times for each number of robots, and the results are illustrated in Figure \ref{fig:63expresult}. The example routes of experiments with one to three robots are displayed in Figure \ref{fig:64expr1}-\ref{fig:66expr3} separately. Figure \ref{fig:64expr1} also has legends labeled on the corresponding items. The search time of the robot test agrees with the result of simulations, namely having a descending trend which is gradually slower. The experiment shows that by applying Algorithms \ref{e82Dchoose2a} \& \ref{e82Dchoose2b} in situation I, the target can be found successfully by 1 to 3 robots without any collision.

\subsection{Factors in Choosing a Grid under the Loose Requirement}
\begin{definition}
    Let $N_{vt}$ represents the number of vertices within $r_{st}$ of one target. Let $P_{t}$ represents the proportion of the number of vertices within $r_{st}$ of all targets to the total number of vertices. 
    \label{d:2$N_{vt}$}
\end{definition}

Under the loose requirement, different grid patterns can be used with various amounts of vertices. The number of vertices is a significant factor for search time. However, fewer vertices do not always result in shorter search time. Three other possible factors for choosing a grid pattern are $P_{t}$, the structure of the grid and $W_{pass}$. Simulations and analyses in this section use situation I and the seventh condition in Table \ref{t2} which satisfies the parameter of robots in Table \ref{t14exppara}. 

\begin{figure}
    \centering
    \subfloat[Ranges and neighbors]{
        \begin{minipage}[t]{0.47\textwidth}
            \centering
            \includegraphics[width=1\linewidth]{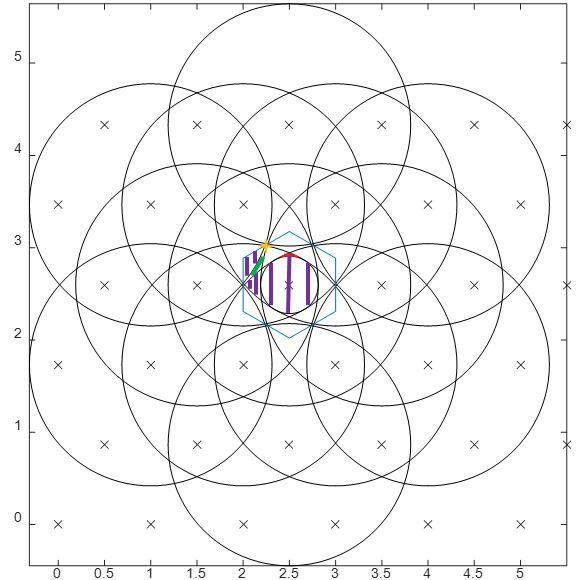}
        \end{minipage}
    }
    \subfloat[Color label]{
        \begin{minipage}[t]{0.47\textwidth}
            \centering
            \includegraphics[width=1\linewidth]{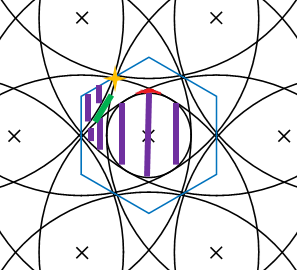}
        \end{minipage}
    }
    
    \caption[$N_{vt}$ for a T grid: star-4 green-5 purple-6 red-7]{$N_{vt}$ for a T grid: star-4 green-5 purple-6 red-7}
    \label{fig:78Tnvt}
\end{figure}

\begin{figure}
    \centering
    \subfloat[Ranges and neighbors]{
        \begin{minipage}[t]{0.47\textwidth}
            \centering
            \includegraphics[width=1\linewidth]{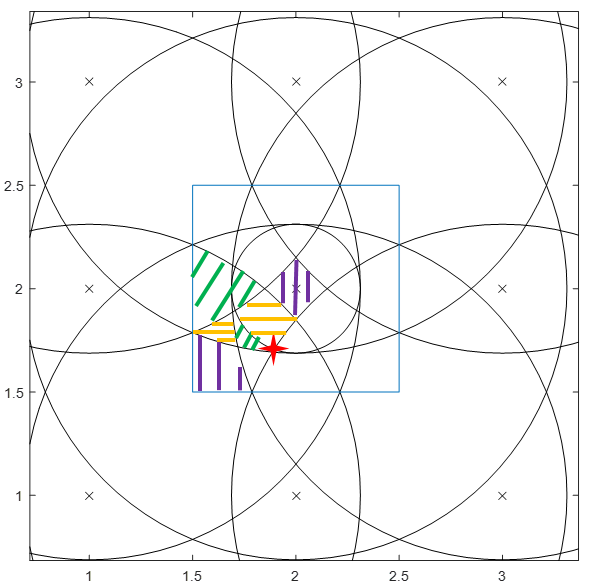}
        \end{minipage}
    }
    \subfloat[Color label]{
        \begin{minipage}[t]{0.47\textwidth}
            \centering
            \includegraphics[width=1\linewidth]{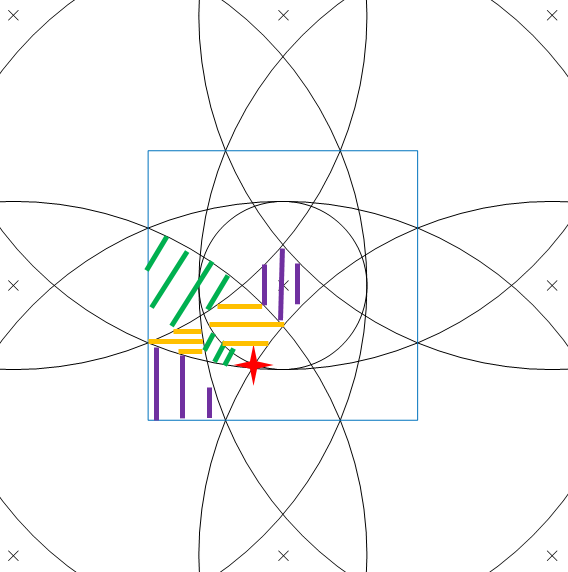}
        \end{minipage}
    }
    
    \caption[$N_{vt}$ for an S grid: purple-4 yellow-5 green-6 star-7]{$N_{vt}$ for an S grid: purple-4 yellow-5 green-6 star-7}
    \label{fig:79Snvt}
\end{figure}

\begin{figure}
    \centering
    \subfloat[Ranges and neighbors]{
        \begin{minipage}[t]{0.47\textwidth}
            \centering
            \includegraphics[width=1\linewidth]{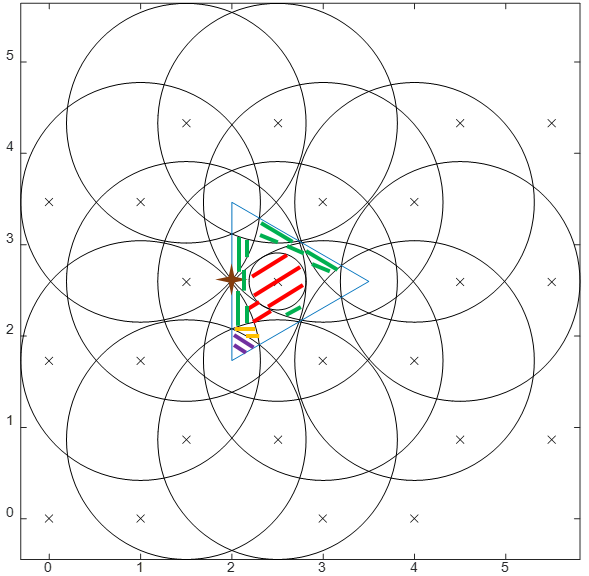}
        \end{minipage}
    }
    \subfloat[Color label]{
        \begin{minipage}[t]{0.47\textwidth}
            \centering
            \includegraphics[width=1\linewidth]{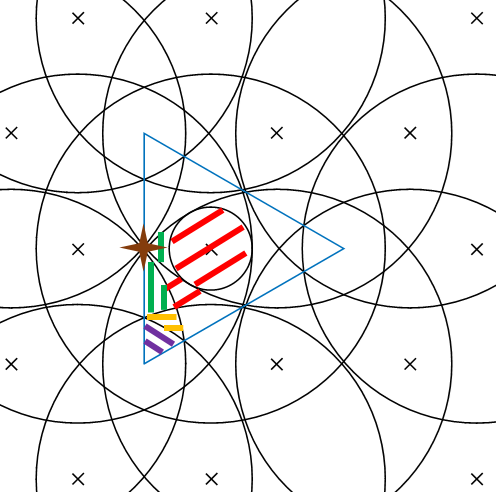}
        \end{minipage}
    }
    
    \caption[$N_{vt}$ for an H grid: star-2 red-3 green-4 yellow-5 purple-6]{$N_{vt}$ for an H grid: star-2 red-3 green-4 yellow-5 purple-6}
    \label{fig:80Hnvt}
\end{figure}

\begin{figure}
    \centering
    \includegraphics[width=0.7\linewidth]{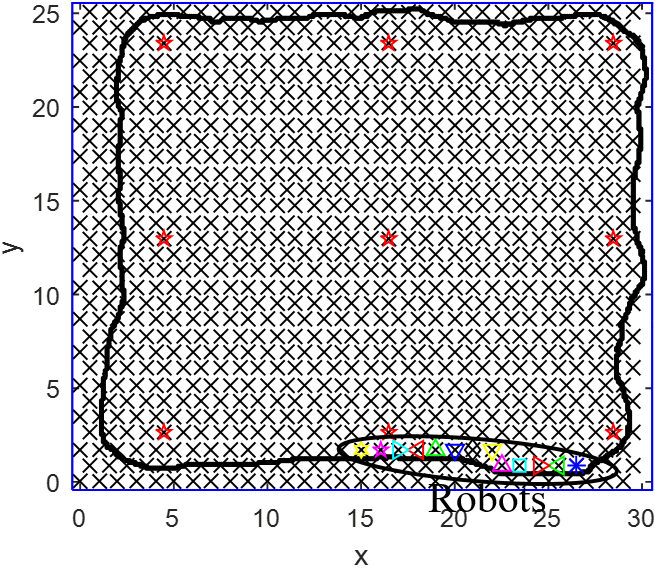}
    \caption[A T grid with no obstacles]{A T grid with no obstacles}
    \label{fig:67noobs}
\end{figure}

\begin{figure}
    \centering
    \includegraphics[width=1\linewidth]{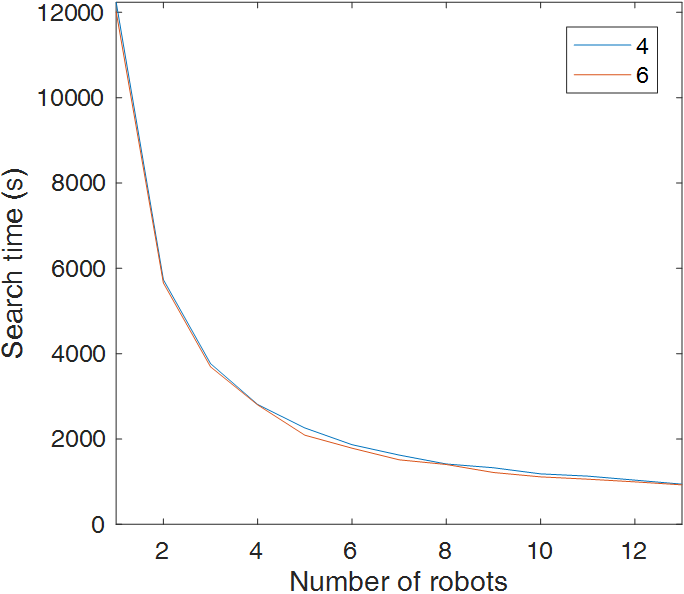}
    \caption[Time for a T grid with $N_{vt}$=4 and $N_{vt}$=6]{Time for a T grid with $N_{vt}$=4 and $N_{vt}$=6}
    \label{fig:68tnvt}
\end{figure}

Example of different $N_{vt}$ are shown in Figure \ref{fig:78Tnvt}, \ref{fig:79Snvt} and \ref{fig:80Hnvt} for a T grid, an S grid and an H grid with different colors. In the examples, $a=1$ and $ r_{st} \geq{a+r_{safe}}$ are set so that a target can be detected from different vertices and the setting could work for all the three gird. The cyan hexagon, square, and triangle show the sections which are occupied by the vertices in the middle of the areas. Circles which have intersections with the cyan areas show all the related $r_{st}$ centered at nearby vertices. Then labels with different colors are used to represent different $N_{vt}$ values. As the graphs are symmetrical, only parts of the area are marked. The effect of $P_{t}$ will be evident when there are many targets in a relatively small area. The area in Figure \ref{fig:67noobs} is used with no obstacles to avoid the effect of $W_{pass}$. The average search time of 500 simulations using 1 to 13 robots via a T grid with $N_{vt}$=4 and $N_{vt}$=6 is presented in Figure \ref{fig:68tnvt}. The corresponding $P_{t}$ values are around 4.5\% and 6.7\% respectively. Figure \ref{fig:68tnvt} shows that when $P_{t}$ is larger, the search time is smaller. To minimize the effect of $P_{t}$ in discussing other factors, this factor of the three kinds of grids should be similar. The ratio between the numbers of vertices of the T grid, the S grid, and the H grid is 1.5:1.3:1. So the corresponding integral $N_{vt}$ values in future simulations are set to be 6, 5 and 4 which lead to the ratio of $P_{t}$ as 1.5:1.25:1. Note that those $N_{vt}$ values are also close to the average values for the three grids in the selected condition, so the setting is practical.

\begin{figure}
    \centering
    \includegraphics[width=0.8\linewidth]{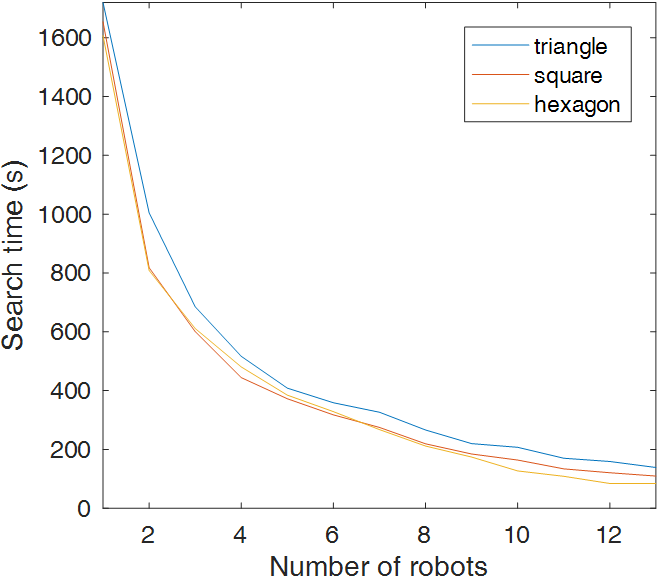}
    \caption[Time for three grids with one target]{Time for three grids with one target}
    \label{fig:693grid1}
\end{figure}

\begin{figure}
    \centering
    \includegraphics[width=0.8\linewidth]{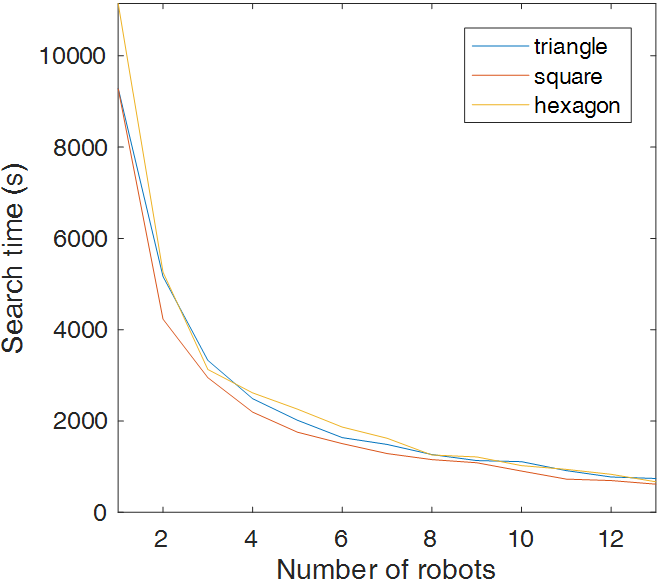}
    \caption[Time for three grids with nine targets]{Time for three grids with nine targets}
    \label{fig:703grid9}
\end{figure}

The structure of the grid affects the search time by affecting the possible routes to a target. This can be understood by looking at the shortest path as an example. In the current situation and condition, all robots use $a_{1}$ as the side length. Then an H grid can be seen as a part of a T grid. Therefore the shortest path via an H grid is equal to or longer than that via the corresponding T grid. However, the average path length which determines the mean time cannot be calculated, so simulations are used to see the effect. Figure \ref{fig:693grid1} and \ref{fig:703grid9} show the average time of 500 simulations for searching one target in the middle or separated nine targets in the area of Figure \ref{fig:67noobs} by one to 13 robots via three grids. There are no obstacles in the area so the effect of $W_{pass}$ can be ignored. Based on Table \ref{t2}, the vertex in the H grid covers the largest area, so the H grid has the least vertices. However, in Figure \ref{fig:693grid1}, the S grid has the least time for three to five robots, and the H grid has the least time for other numbers of robots. In Figure \ref{fig:703grid9}, the S grid has the least time. Therefore, the fewest vertices do not guarantee the least search time, and the structure of the grid can affect the time. 

The third factor $W_{pass}$ affects the search time by affecting the chances to enter or leave an area as discussed in Section \ref{s9simustrict}. For the same $W_{pass}$, the opportunity for each grid is distinct. For example, if all these three grids use $a$ as the side length with a passage $W_{pass}=2a_{1}+2r_{safe}$ which is the largest one in Table \ref{t1}, the width can fit at least three vertices for a T grid and two vertices for both an S grid and an H grid. However, the corresponding ratio of the number of vertices in the three grids is 1.5:1.3:1. Thus, robots using an S grid may have less chance to enter or leave a search section through this passage. So $W_{pass}$ needs to be considered in the simulation design.

From the above discussions, it is clear that when choosing the grid for the loose requirement, the number of vertices is important but not the only determinant. The average $N_{vt}$ needs to be calculated based on the condition in Table \ref{t2}. The given $W_{pass}$ or the minimum $W_{pass}$, which satisfies all the three grids, needs to be found. Then the simulations with these $N_{vt}$ and $W_{pass}$ should be run to compare the search time of the three grids and find the best one.
\subsection{Section Summary}
This section presented a random decentralized algorithm with the repulsive force for multiple robots to search targets in 2D areas through a grid pattern without any collision. The repulsive force could help robots to be away from each other if they met to reduce repetition. The convergence of the algorithm is proved, and simulations showed that the proposed algorithm is the best comparing to other six algorithms. An example of how to estimate the relation between search time, the number of robots and the size of the area was demonstrated for nine targets to help decide the suitable number of robots. A test on Pioneer 3-DX robots also verified the algorithm. When only detecting all the vertices is required, simulations need to be used to find the best grid considering the grid with the least number of vertices, structure of the grid pattern, suitable $W_{pass}$ and $P_{t}$. 

\section{3D Tasks}
This section presents a decentralized random algorithm for multiple robots to search targets in an unknown 3D area with obstacles via the cubic lattice. The algorithm is the same as the one in 2D. It considers the realistic physical constraints and conditions of all parameters as well as collision avoidance so that it can be applied to real robots. Robots choose vertices in sequence and move between vertices along edges of the lattice based on local information. The convergence of the algorithm is proved, and the effectiveness is shown in comparison with three other algorithms.

\subsection{Problem Statement}
The problem is the same as that in Chapter 4. So the definitions and assumptions below are claimed without detailed description. 

In the search task, a three-dimensional area $A\subset\mathbb{R}^{3}$ with a limited number of obstacles $ O_{1} $ to $O_{l} $ is used. $m$ robots labelled as $rob_{1}$ to $rob_{m}$ are employed to search $n$ static targets $t_{1}$ to $t_{n}$ through a cubic grid with a side length $a$. For $rob_{i}$, its coordinate at time $k$ is $p_{i}(k)$. Let $T$ be the set of coordinates of targets and $T_{ki}$ be the set of targets coordinates known by $rob_{i}$. The initial setting is in Figure \ref{fig:413dinitialsetting}.
\begin{assumption}
    Area $A$ is a bounded, connected and Lebesgue measurable set. The obstacles $O_{i}\subset\mathbb{R}^{3}$ are non-overlapping, closed, bounded and linearly connected sets for any $i>0$. They are both static and unknown to robots. \label{a:3D2area}
\end{assumption}
\begin{definition}
    Let $O:=\bigcup{O_{i}}$ for all $i>0$. Then $A_{d}:=A\setminus{O}$ is the area that needs to be detected.\label{d:3D2obs}
\end{definition}
\begin{assumption}
    Initially, all robots have the same grid with the same coordinate system and are allocated at different vertices along the border of the search area. In simulations, robots are manually set at vertices of the C grid\label{a:3D2intialposition}
\end{assumption}
\begin{assumption}
    All sensors and communication equipment that robots carry have spherical ranges. Other radiuses mentioned are also spherical. \label{a:3D2ranges}
\end{assumption}
\begin{definition}
    Robots have a swing radius $r_{rob}$ and the total error is $e$. Then a safety radius to avoid collisions is $ r_{safe}\geq{r_{rob}+e} $ (see Figure \ref{fig:16range3d}).\label{d:3D2rsafe}
\end{definition}
\begin{assumption}
    $W_{pass}\geq{ a\sqrt{3}+2r_{safe}}$ (see Figure \ref{fig:18wpass3D}).\label{a:3D2wpass}
\end{assumption}
\begin{assumption}
    $r_{so}\geq{a+r_{safe}}$ (see Figure \ref{fig:16range3d}).\label{a:3D2rso}
\end{assumption}
\begin{assumption}
    $ r_{c}\geq{2*a+e}$(see Figure \ref{fig:16range3d}).\label{a:3D2rc}
\end{assumption}
\begin{assumption}
    The task area is vast. So, $a$ is much smaller than the length, the width and the height of area $A$. Thus, there are no targets between the boundary and the frontiers of the target sensing range, which means the boundary effect can be ignored. \label{a:3D2boundaryeffect}
\end{assumption}
\begin{assumption}
    $r_{st}\geq{2a/\sqrt{3}+r_{safe}-r_{rob}}$ based on Table \ref{t3}.
\end{assumption}
$V_{a}$ is the set of all the accessible vertices. $N_{s,i}(k)$ represents the sensing neighbors set around $rob_{i}$. In $N_{s,i}(k)$, $V_{v,i}(k)$ and $V_{u,i}(k)$ denote the sets for visited vertices and unvisited vertex and the choice from these two sets are $c_{v}$ and $c_{u}$ respectively. Then a set of new definitions for the distances and vertices in this chapter are stated as follows:
\begin{definition}
    The average position of choices made by robots which chose earlier and current positions of communication neighbors who have not chosen yet is $p_{i,ave}(k)$. Let $d_{c,p_{i,ave}(k)}$ be the set of distances from $p_{i,ave}(k)$ to the possible choices. Then, $ d_{c_{u},p_{i,ave}(k)}$ is the set of distances to unvisited vertices and $ d_{c_{v},p_{i,ave}(k)}$ is the set of distances to visited ones. Then $ max(d_{c_{u},p_{i,ave}(k)})$ and $max(d_{c_{v},p_{i,ave}(k)}) $ are the maximum values in the two sets respectively. The corresponding sets of vertices which result in these maximum distances are represented as $ V_{umax,i}(k)$ and $ V_{vmax,i}(k)$. 
\end{definition}
\subsection{Procedure and Algorithm}    
The general procedure of the algorithm is the same as the 2D algorithm. So the same flow charts in Figure \ref{fig:342dflowsitu1} and \ref{fig:352dflowsitu2} are available for Situation I and II. However, the way to set the sequence is different from the 2D case and readers can refer to Section \ref{s83Dalgo} for detail. Comparing to the first search algorithm, the improvement of this algorithm is on how to use local information to make choice. So, when there are no local neighbors in $r_{so}$, Algorithm \ref{e83Dchoose2a} is used but when there are neighbor robots within $r_{so}$ of a robot, it will apply Algorithm \ref{e83Dchoose2b} to make choice. 
\begin{equation}
p_{i}(k+1)=
\begin{cases}
c_{u} \text{ with prob. } 1/\lvert V_{u,i}(k)\rvert, \text{ if } \lvert V_{u,i}(k)\rvert\neq{0},\\
c_{v} \text{ with prob. } 1/\lvert V_{v,i}(k)\rvert, \text{ if } (\lvert V_{u,i}(k)\rvert=0)\&(\lvert V_{v,i}(k)\rvert\neq{0}),\\
p_{i}(k),  \text{ if } \lvert N_{s,i}(k) \rvert=0
\end{cases}
\label{e83Dchoose2a}
\end{equation}
\begin{equation}
p_{i}(k+1)=
\begin{cases}
c_{u} \text{ with prob. } 1/\lvert V_{umax,i}(k)\rvert, \text{ if } \lvert V_{u,i}(k)\rvert\neq{0},\\
c_{v} \text{ with prob. } 1/\lvert V_{vmax,i}(k)\rvert, \text{ if } (\lvert V_{u,i}(k)\rvert=0)\&(\lvert V_{v,i}(k)\rvert\neq{0}),\\
p_{i}(k),  \text{ if } \lvert N_{s,i}(k) \rvert=0
\end{cases}
\label{e83Dchoose2b}
\end{equation}
Algorithm \ref{e93DstopI} and \ref{e93DstopII} are employed to judge the stop time.
\begin{equation}
\text{In situation I: }p_{i}(k+1)=p_{i}(k)  \text{, if } \lvert T_{ki}\rvert=\lvert T\rvert.
\label{e93DstopI}
\end{equation}
\begin{equation}
\text{In situation II: }p_{i}(k+1)=p_{i}(k)  \text{, if } \forall v_{i}\in V_{a}, \text{ } \lvert V_{u,i}(k)\rvert=0.
\label{e93DstopII}
\end{equation}
\begin{theorem}
    Suppose that all assumptions hold and the proposed decentralized random algorithm, namely Algorithm \ref{e83Dchoose2a} and \ref{e83Dchoose2b} with related judgment strategy \ref{e93DstopI} or \ref{e93DstopII}, are used. Then for any number of robots, with probability 1, there exists such a time $k_{0}>0$ that all targets or all vertices are detected.
    \label{th3D2}
\end{theorem}
\begin{prove}
     The algorithm \ref{e83Dchoose2a} and \ref{e83Dchoose2b} with corresponding judgment methods form an absorbing Markov chain including both transient states and absorbing states. The transient states are all the approachable vertices of the C grid where robots visit but do not stop forever. Absorbing states are the vertices where the robots finally stop. By applying the proposed algorithm, robots start from some initial transient states and move between different transient states with the probability from the algorithm. The robot will go to the unvisited sensing neighbors if there are any with the help of the repulsive force. If all sensing neighbors of a robot are visited, the robot will randomly choose an accessible neighbor with the repulsive force. If no accessible neighbor vertices are available, it stays at current vertices. For \ref{e93DstopI}, robots stop at an absorbing state if all robots know $\lvert T_{ki}\rvert=\lvert T\rvert$. For \ref{e93DstopII}, absorbing states are reached until $\lvert V_{u,i}(k)\rvert=0$ for all accessible vertices. Based on Assumption \ref{a:3D2wpass}, absorbing states can be achieved from any initial states, namely with probability 1. This completes the proof of Theorem \ref{th3D2}.
\end{prove}
\subsection{Simulation Results}
\begin{figure}
    \centering
    \includegraphics[width=0.7\linewidth]{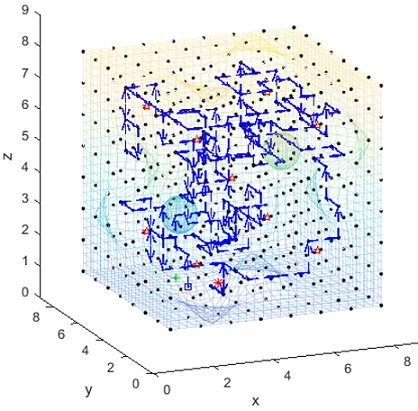}
    \caption[The route of robot 1 in a simulation with 3 robots]{The route of robot 1 in a simulation with 3 robots}
    \label{fig:903d2r1}
\end{figure}

\begin{figure}
    \centering
    \includegraphics[width=0.7\linewidth]{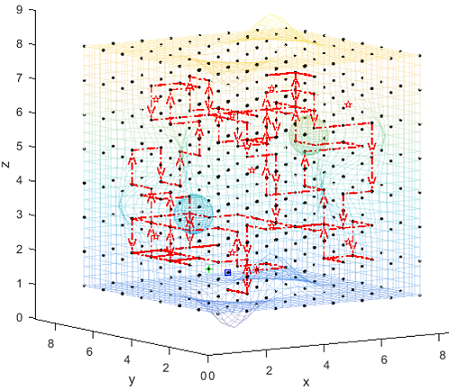}
    \caption[The route of robot 2 in a simulation with 3 robots]{The route of robot 2 in a simulation with 3 robots}
    \label{fig:913d2r2}
\end{figure}

\begin{figure}
    \centering
    \includegraphics[width=0.7\linewidth]{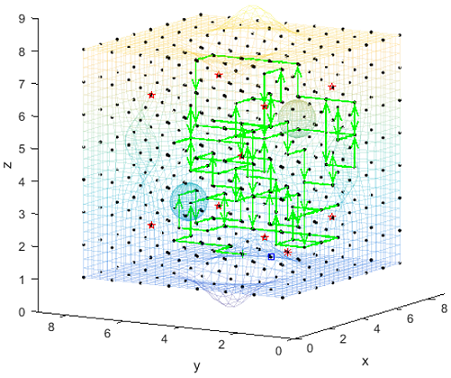}
    \caption[The route of robot 3 in a simulation with 3 robots]{The route of robot 3 in a simulation with 3 robots}
    \label{fig:923d2r3}
\end{figure}

\begin{figure}
    \centering
    \includegraphics[width=1\linewidth]{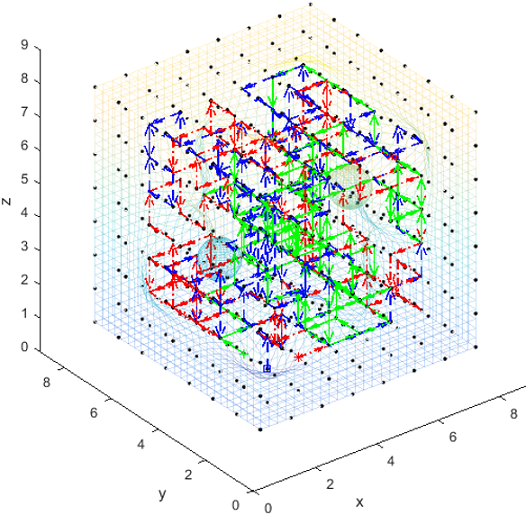}
    \caption[The combined routes in a simulation with 3 robots]{The combined routes in a simulation with 3 robots}
    \label{fig:933d2r123}
\end{figure}

A search task for situation I is simulated in MATLAB2016a. The parameters are set based on Table \ref{t8}. To display the details of the proposed algorithm, a small area which has 8*8*8 vertices with 9 targets is used to be searched by 3 robots using control laws \ref{e83Dchoose2a}, \ref{e83Dchoose2b} \& \ref{e93DstopI} as shown in Figure \ref{fig:903d2r1}-\ref{fig:933d2r123}. The routes are illustrated with arrows to show the directions of movements, and the path of each robot has its color. By analyzing the route of each robot, no collisions were found, and each target was within $r_{st}$ of a visited vertex.

To verify the algorithm further, the search time using different algorithms namely the LFP algorithm in \cite{RN102}, the RU algorithm in \cite{RN2}, rules \ref{e53Dchoose} \& \ref{e63DstopI}(S1 in line charts) in Chapter 4 and rules \ref{e83Dchoose2a}, \ref{e83Dchoose2b} \& \ref{e93DstopI}(S2 in line charts) in this chapter are compared. LFP and RU are chosen as they are better than related algorithms based on the simulation results in 2D. In RU, there are collisions so it is assumed that one collision needs one more movement time to get solved although longer time may be needed if there are consecutive collisions. According to the assumption \ref{a:3D2boundaryeffect}, targets are allocated in the detectable area as in Figure \ref{fig:413dinitialsetting}. Time in movement in the simulation is 5.75s in total including 4.5s for a translation and 1.25s for a rotation which is in Table \ref{t8}.

\begin{figure}
    \centering
    \includegraphics[width=1\linewidth]{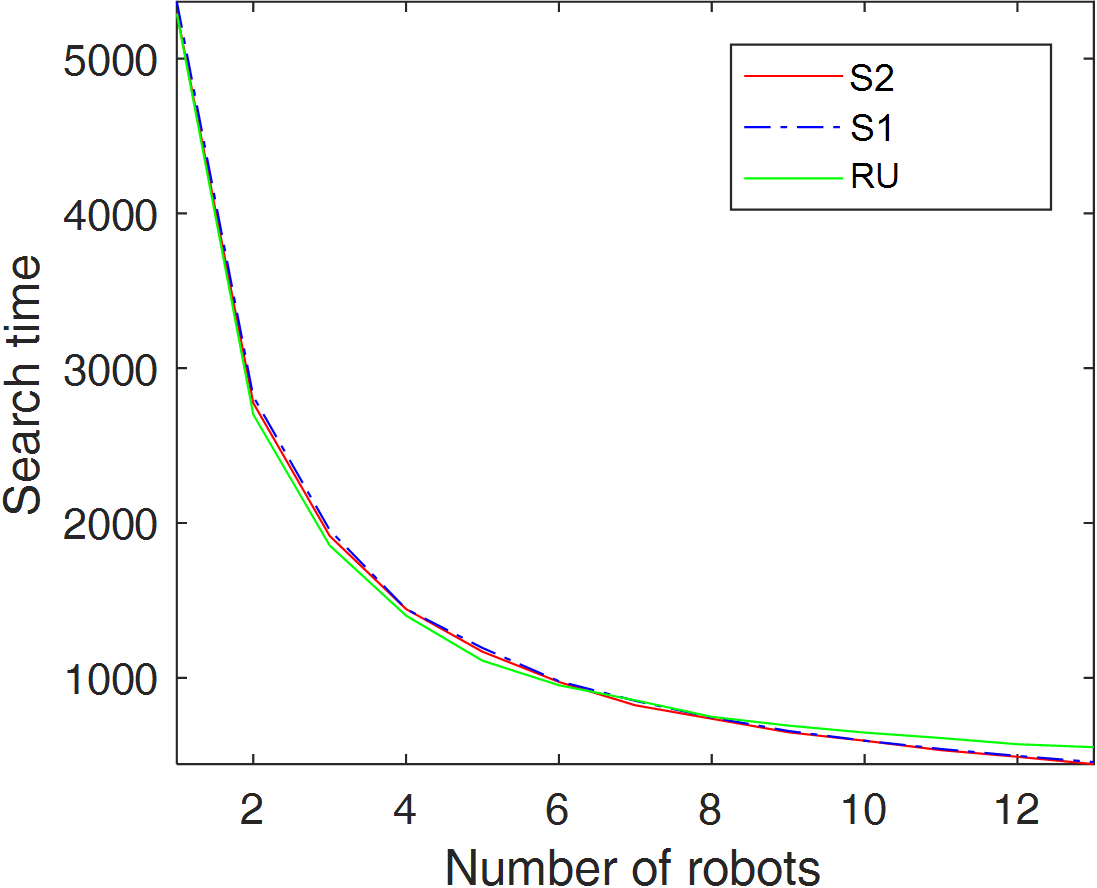}
    \caption[Time for four algorithms with 9 targets]{Time for four algorithms with 9 targets}
    \label{fig:793d2compare9tar}
\end{figure}

The simulation for searching nine separated targets in a larger area (12*12*12 vertices) by 1 to 13 robots has been run for 500 times to get the average search time. Algorithm LFP is always much slower than the others as no previous map will be used. So, Table \ref{t153D2RUresult} is used to show the time for LFP and RU specially and algorithms except LFP are drawn in \ref{fig:793d2compare9tar} to make the graphs clear. The figure demonstrates that time in the three algorithms for one robot is similar as they are the same in that situation. For more robots, time for S2 is always smaller than S1 as robots are separated to different parts to find distributed targets. When there are 1 to 6 robots, algorithm RU is the fastest because robots can move without considering the collisions, which provides more choices in each step for robots to search different areas and it has the simplest rule which saves the calculation time. Also, its disadvantage that it needs extra time to deal with collisions is not severe as the number of collisions is small. When there are more than six robots, algorithms S2 is the fastest as the time saved in collision avoidance is larger than the increased waiting time before choosing in the algorithm. So, the proposed algorithm is the best comparing to those with the collision avoidance scheme. When there are a large number of robots, it is even better than algorithm RU which does not avoid collisions between robots. In all the four algorithms, when the number of robots ascended, the search time descended sharply at the beginning and slowly later. It can be predicted that if the number robots increases, the search time of the proposed algorithm may have a stable period or even an increasing part at the end as the algorithm time in the step `wait' will increase and take up a larger proportion of the total time.

Therefore, the optimal number of robots should be decided based on comparing the benefits of decreased search time and the cost of increased expense on robots. The search time is also affected by unknown factors including the initial positions of robots and the shape and the size of the area. If the size of the area is given, simulations like those in this paper can be used to estimate the proper rough range of the number of robots according to the limitation of the search time in a general situation.
\begin{table}
    \centering
    \renewcommand\arraystretch{2}
    \begin{tabular}{|c|c|c|c|c|c|c|c|}
        \hline 
        No.&    1&    2&    3&    4    &5&    6&    7 \\ 
        \hline 
        RU&5290&    2701&    1855&    1402&    1111&    952&855 \\ 
        \hline 
        LFP&9624&    5063&    3389&    2688&    2004&    1771&1476\\ 

        \hline
        No.&8&9&10&11&12&13&\\
        \hline
        RU&    748    &692&    646    &611&    571&    552&    \\    
        \hline
        LFP&1352&    1193&    1176&    1005&    903&    874&\\
        \hline
    \end{tabular}  
    \caption[Search time for algorithm LFP and RU with 9 targets]{Search time for algorithm LFP and RU with 9 targets}
    \label{t153D2RUresult}
\end{table}
\subsection{Section Summary}
This section proposed a random decentralized search algorithm for multiple robots in a 3D region which is based on the 2D algorithm. All robots move simultaneously with a C lattice to avoid collision. The repulsive force helps robots to have less repeated area. By comparing different algorithms in simulations, the proposed algorithm is always the best in those with collision avoidance and the best in all the algorithms when there are more than a certain number of robots. 

\section{Summary}
This chapter proposed the second search algorithm for a group of mobile robots. The robot moves via the selected grid pattern to select vertices in sequence and avoid collisions in an unknown area. Compared to the first algorithm, the repulsive force is added to disperse robots when they are neighbors in obstacle sensing range. So they tend to leave for different directions to have less repeated area. The algorithm is suitable for both 2D areas and 3D areas with differences in the selection sequence and the dimension of parameters. A rigorous mathematical proof of the convergence of the algorithm was provided with probability 1. In the 2D area, this chapter illustrated how to find the relation between the search time and other parameters. With loose requirement, it suggested using simulations to find the best grid considering different factors. Then, an experiment on three Pioneer 3-DX robots was conducted to verify the algorithm. In both 2D and 3D area, the proposed algorithms were compared with other grid-based algorithm and \acl{LF} related algorithm to show the effectiveness. Conclusively, the algorithm is scalable, robust, reliable and effective.  
\chapter{A Collision-Free Random Search with the Breath-First Search Algorithm}
\chaptermark{A Collision-Free Random Search with the Breath-First Search Algorithm}
The third grid based collision-free search algorithm is proposed in this chapter. The algorithm is decentralized with the help of a grid pattern in a 2D or 3D unknown area. Robots choose their future steps based on random algorithm when there are unvisited neighbors and use the breadth-first search algorithm when they are surrounded by visited vertices. As the algorithm in this chapter is solving the same problem as the task in Chapter 4, the way to state the problem and describe the algorithm will be similar but simpler, and only the different parts will be emphasized. In the two previous search algorithms, there are a considerable number of repeated vertices in the search. So this chapter focuses on how to use the known information to escape the visited area. Thus, a graph search algorithm, namely the breadth-first search algorithm is added to find one of the nearest unvisited vertices and the shortest way to it. The shortest way in this chapter means the route with the fewest number of steps from the current vertex to the target vertex. To check the effect of this algorithm, simulations of the algorithm will be carried out comparing to 7 other algorithms to show the effectiveness. Then the advantages and disadvantages of the three proposed algorithms will be discussed. 

Similar algorithms with a method to decrease the amount of repeated search include \cite{RN233} and \cite{RN118,RN119}. In \cite{RN233}, the grid pattern is also used for robots to move through. Robots used the explored map to find the nearest vertices and `the nearest' means the minimum Euclidean distance which can be represented by $d_{e}$. However, with unknown obstacles in the area, a robot may need to move a longer actual step $d_{s}$ to reach the target point, especially when the robot and the target vertex is at the two sides of the narrow part of a long obstacle. With this method, robots may walk without passing the edge of the grid, and a dynamic collision avoidance method must be provided. However, \cite{RN233} did not provide the detail of collision avoidance method and only claimed it used the embedded repulsive force method from Pioneer 3-DX robots. It also did not provide what a robot should do if it is not on a vertex, but its algorithm with assumptions on ranges was based on the implied assumption that robots are on the vertices. As the maximum speed is set, the step with different lengths will use different time. If all robots waited for the longest movement time which is estimated and designed for the one that leaves the visited area to move further, the normal robots which only move $a$ can be thought as wasting their time in waiting. If robots are not synchronized, the communication time points in the algorithm are not synchronized either. Suppose that the blocking mode communication was used to solve the synchronization problem, robots needed to wait for each other in the step for communication which might also be thought as a waste of time. In the experiment of \cite{RN233}, the result did not agree with the algorithm as one robot moved five steps with one unit step length in each step but another robot moved six steps with one step having two unit step lengths. With blocking mode communication, these two robots should both go the sixth vertex as they chose at the same time. Without blocking mode, the one with five steps should go to the last vertex as it chose earlier. However, neither of these agrees with the result of the provided experiment. So the experiment may be a fake one. Also, the experiment did not show the situation to clarify its ambiguous collision avoidance algorithm. In \cite{RN118,RN119}, a \acl{LF} algorithm is used to have less repetition than the \ac{RW} algorithm. However, the levy flight algorithm of a robot only ensures this robot has less repetition than using the RW algorithm, but different robots may still repeat the visited point of other robots. The reason is that robots are independent and did not communicate the information about the visited area as the choice was not based on that. Another disadvantage of \cite{RN118,RN119} is that they did not consider any collision avoidance and the task area has no obstacles. So that algorithm is not practical and could not be used on real robots.  

The explored area of the robot is a known subarea of the task area. Therefore, graph search algorithms are considered in this chapter. Except for the breadth-first search algorithm, there are other related graph search algorithms such are A* algorithm \cite{RN224,RN179}, and greedy best first search algorithm. However, these two algorithms are suitable to the situation for finding the path to a known target position. In this situation, the greedy best first algorithm estimates the distance to the target and only needs to explore a small area to find a path while the breadth-first search calculates the distance to the start point in each direction evenly which results in a large explored area but the shortest path. A* algorithm estimates the sum of the estimated values of the two algorithms above, so it has the advantage of both algorithms, namely smaller explored area, and shortest path. However, in the problem discussed in this chapter, there may be many unvisited vertices around the visited area, and the target vertex should be the one with the fewest steps ($d_{s}$) to reach which is unknown. If the A* algorithm is used, it needs to be executed for each unvisited vertex at the boundary of the explored area which may have repeated explored parts. Then all paths need to be compared to find the one with the smallest $d_{s}$. The breadth-first search algorithm could find the paths to all locations without repeated exploration and will explore different directions equally and generate the route after exploration. So it could have less calculation and should be used to ensure the found path has the smallest $d_{s}$.  
\section{2D Tasks}
This part proposes the third decentralized random search algorithm. It adds the breadth-first search on the 1st search algorithm to help robots to jump out of the detected zone. The algorithm is compared with five algorithms from other authors and two search algorithms in previous chapters to show the advantage of this proposed algorithm. The convergence of the algorithm is also proved.
\subsection{Problem Statement}
In this section uses a two-dimensional area $A$ with a limited number of static obstacles $ O_{1} $, $ O_{2} $,$\ldots $, $O_{l} $. Here, the strict requirement is considered, and the T grid is used. There are $m$ robots labelled as $rob_{1}$ to $rob_{m}$ to search static targets $t_{1}$ to $t_{n}$. An arbitrary known T grid pattern with a side length of $a$ is used for robots to move in each step with the following definitions and assumptions. All these initial settings can be seen in Figure \ref{fig:712d2initialsetting}. The repulsive force is no longer needed, so the related definitions are not needed either. Further explanation of the definitions and assumptions below can be seen in Chapter 4. 
\begin{assumption}
    The area $A$ is a bounded, connected and Lebesgue measurable set. The obstacles $O_{i}$ are non-overlapping, closed, bounded and linearly connected sets for any $i\geq{0}$. Robots have no knowledge about them before exploration. \label{a:2area3}
\end{assumption}
\begin{definition}
    Let $O:=\bigcup{O_{i}}$ for all $i>0$. Then $A_{d}:=A\setminus{O}$ represents the area that needs to be detected.\label{d:2obs3}
\end{definition}
\begin{assumption}
    The initial condition is that all robots use the same T grid pattern and the same coordinate system. They start from different vertices near the boundary which can be seen as the only known entrance of the area.\label{a:2Initial_P3}
\end{assumption}
\begin{assumption}
    All the ranges and radius in 2D search problems are circular. \label{a:2circular3}
\end{assumption}
\begin{definition}
    To avoid collisions with other things before move, a safe distance $ r_{safe} $ should include both $ r_{rob} $ and $e$ (see Figure \ref{fig:5r_ob}). So $ r_{safe}\geq{r_{rob}+e} $.\label{d:2rsafe3}
\end{definition}
\begin{assumption}
    All passages $W_{pass}$ between obstacles or between the obstacle and the boundary are wider than $a+2*r_{safe}$. Other $W_{pass}$ for other grid patterns can be seen in Table \ref{t1}.\label{a:2wpass3}
\end{assumption}
\begin{assumption}
    The sensing radius for obstacles is $r_{so}$ where $r_{so}\geq{a+r_{safe}}$.\label{a:2$r_{so}3$}
\end{assumption}
\begin{assumption}
    $ r_{st} \geq{a+r_{safe}}$.\label{a:2rst3}
\end{assumption}
\begin{assumption}
    $ r_{c}\geq{2*a+e}$ (see Figure \ref{fig:6rangeorder}). \label{a:2rc3}
\end{assumption}
\begin{assumption}
    The curvature of the concave part of the boundary should be smaller than or equal to $1/ r_{st} $.\label{a:2D3curvature}
\end{assumption} 
$V_{a}$ is the set of all the accessible vertices. $N_{s,i}(k)$ represents the set of sensing neighbors around $rob_{i}$. In $N_{s,i}(k)$, $V_{v,i}(k)$ and $V_{u,i}(k)$ denote the two sets for visited vertices and unvisited vertices and the choice from these two sets are $c_{v}$ and $c_{u}$ respectively.

For the breadth-first search algorithm, the shortest route to the nearest unvisited vertex is generated. 
\begin{definition}
    The route is represented as a set of vertices $V_{r,i}$ for $rob_{i}$. Then $\lvert V_{r,i}(k)\rvert $ is the total number of steps in that route. Specially, if the route has not been fully generated, $\lvert V_{r,i}(k)\rvert=0$. Let $v_{r,i}(j)$ be the $j$th step in the route and $c_{r,i}$ be the choice from the route $V_{r,i}(k)$. The number of steps left in the route is then $\lvert V_{r,i}(k)\rvert-j$. \label{d2D3vri}
\end{definition}
\subsection{Procedure and Algorithm }
\label{s102D3}
The proposed algorithm drives robots to go through the area through a grid pattern. If the breadth-first  rule is not applied, the general progress in this search algorithm is the same as the first search algorithm and follow the flow chart in Figure \ref{fig:342dflowsitu1} and \ref{fig:352dflowsitu2}. However, there are also some differences.

The way to use the breadth-first search for the choice step of $rob_{i}$ is described as follows: 
\begin{enumerate}[fullwidth,itemindent=0em]
    \item Update map: In the turn of $rob_{i} $, it updates its map based on received information which include the choice from others if $rob_{i} $ is not the first to choose in this loop.
    \item check obstacles: $rob_{i}$ check obstacles in the $r_{so}$ and calculates $\lvert V_{u,i}(k)\rvert$.
    \item check $\lvert V_{u,i}(k)\rvert$: Is there any vacant unvisited neighbor? 
    \begin{enumerate}[label=3.\arabic*]
        \item {\color{red}$\lvert V_{u,i}(k)\rvert\neq0$ and select $p_{i}(k+1)$:} If $\lvert V_{u,i}(k)\rvert$ is not 0, there are unvisited vertices so $rob_{i}$ choose one from $V_{u,i}(k)$ randomly.
        \item $\lvert V_{u,i}(k)\rvert$=0: check $\lvert V_{v,i}(k)\rvert$
        \begin{enumerate}[label=3.2.\arabic*]
            \item {\color{red}$\lvert V_{v,i}(k)\rvert=0$ and stay:} stay at current position $p_{i}(k)$ as all neighbor vertices are blocked.
            \item $\lvert V_{v,i}(k)\rvert\neq0$ and set flag: There are no unvisited neighbors and $rob_{i} $ is surrounded by visited neighbors. Thus, the flag bit for whether the route map is generated or not is set. In the next step, $rob_{i}$ will have higher priority than others with the flag bit being zero. If multiple robots within $ r_{c}$ set their flag bits, they will judge the priority between them according to the original method which is base on relative positions.
        \end{enumerate}
    \end{enumerate}
    \item Check the old route: $rob_{i}$ checks whether there is any previously generated route map to follow. 
    \begin{enumerate}[label=4.\arabic*]
        \item No routes: $rob_{i}$ will find a path using the breadth-first search algorithm
        \item A route is being calculated: $rob_{i}$ will continue calculation use the updated map if it is updated.
        \item A full route found: $rob_{i}$ will check whether the selected vertex to go is still unvisited or not based on the explored area.
        \begin{enumerate}[label=4.2.\arabic*]
            \item {\color{red}Unvisited}: If it is not visited, keep using the route and choose the next vertex $c_{r,i}$ in the route map.
            \item Visited: $rob_{i}$ will redesign a route based on the explored area.        
        \end{enumerate} 
    \end{enumerate}
    
    \item The loop for checking visitation states: $rob_{i}$ checks the visitation states of the sensing neighbors of current sensing neighbors. Each checked vertex is saved in a list with its father node information and the number of steps from $p_{i}(k)$ to it. 
    \begin{enumerate}[label=5.\arabic*]
        \item No unvisited vertices: continue the loop and check one step length further until an unvisited vertex is found. If a vertex is found more than once, only the first will be recorded as the corresponding path has fewer steps from $p_{i}(k)$. 
        \item Found an unvisited vertex: break the check loop. Then following the father vertices back, the shortest route $V_{r,i}(k)$ from $p_{i}(k)$ to the unvisited vertex is generated.
    \end{enumerate} 
    \item Check the time for the selection: As the map may be large, the search may cost a long time. The time for the selection step needs to be checked
    \begin{enumerate}[label=6.\arabic*]
        \item {\color{red}$\leq{\text{time for normal selection}}$:} $rob_{i}$ chooses the next step $c_{r,i}$ based on the route $V_{r,i}(k)$.
        \item {\color{red}$>$time for normal selection:} $rob_{i}$ stays at current position $p_{i}(k)$. In the time slot for the movement, continue its calculation. In the next loop, stop calculation and do normal things based on the flow chart until it is the turn for $rob_{i}$.
    \end{enumerate}
\end{enumerate}

The steps above in red will generate a choice. Mathematically, this can be concluded as:
\begin{equation}
p_{i}(k+1)=
\begin{cases}
c_{u} \text{ with prob. } 1/\lvert V_{u,i}(k)\rvert, \text{ if } \lvert V_{u,i}(k)\rvert\neq{0},\\
c_{r,i}, \text{ if } (\lvert V_{u,i}(k)\rvert=0)\&((\lvert V_{v,i}(k)\rvert\neq0)\&(\lvert V_{r,i}(k)\rvert\neq{0})),\\
p_{i}(k),  \text{ if } (\lvert N_{s,i}(k) \rvert=0)\mid(\lvert V_{r,i}(k)\rvert={0}).
\end{cases}
\label{e182Dchoose3}
\end{equation}

The stop strategy for this choosing rule is the same as Chapter 4.
\begin{equation}
\text{In situation I: }p_{i}(k+1)=p_{i}(k)  \text{, if } \lvert T_{ki}\rvert=\lvert T\rvert.
\label{e192DstopI3}
\end{equation}

\begin{equation}
\text{In situation II: }p_{i}(k+1)=p_{i}(k)  \text{, if } \forall v_{i}\in V_{a}, \text{ } \lvert V_{u,i}(k)\rvert=0.
\label{e192DstopII3}
\end{equation}
\begin{theorem}
    Suppose that all the above assumptions hold and the search algorithm \ref{e182Dchoose3} and a judgment strategy \ref{e192DstopI3} or \ref{e192DstopII3} are used for situation I or situation II. Then, for any number of robots, there is such a time $k_{0}>0$ that all targets are detected with probability 1.
    \label{th2D3}
\end{theorem}
\begin{prove}
    The algorithm \ref{e182Dchoose3} with a judgment method forms an absorbing Markov chain. It includes both transient states and absorbing states. The transient states are all the approachable vertices of the grid that robots visit but do not stop forever. Absorbing states are the vertices that the robots stop forever. Applying Algorithm \ref{e182Dchoose3}, robots select the unvisited neighbor vertices to go. If all sensing neighbors are visited before finding the targets, robots will look for the shortest route to the nearest unvisited vertex and only stop when all targets are found, or all vertices are detected. For \ref{e192DstopI3}, the absorbing state will be reached when robots know $\lvert T_{ki}\rvert=\lvert T\rvert$. For \ref{e192DstopII3}, absorbing states are reached until $\lvert V_{u,i}(k)\rvert=0$ for any vertex. The assumption about $W_{pass}$ guarantees that all the accessible vertices can be visited so absorbing states can be achieved from any initial states, namely with probability 1. This completes the proof of Theorem \ref{th2D3}
\end{prove}
\subsection{Simulations with the Strict Requirement}
\label{s9simustrict3}
Situation I of the proposed algorithm is simulated to verify the algorithm. For situation I, one to nine targets which are averagely allocated in the area are tested, and the initial setting is the same as Figure \ref{fig:712d2initialsetting}. 

\begin{figure}
    \centering
    \includegraphics[width=0.8\linewidth]{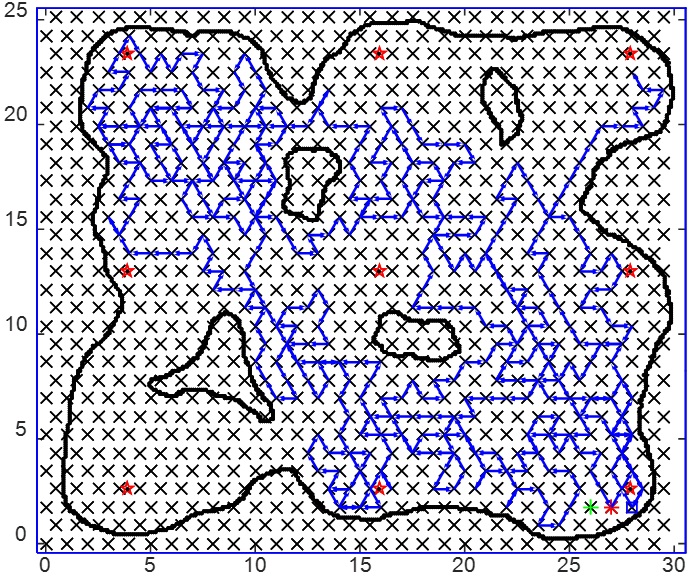}
    \caption[The route of robot 1 in a simulation with 3 robots]{The route of robot 1 in a simulation with 3 robots}
    \label{fig:822d3r1}
\end{figure}

\begin{figure}
    \centering
    \includegraphics[width=0.8\linewidth]{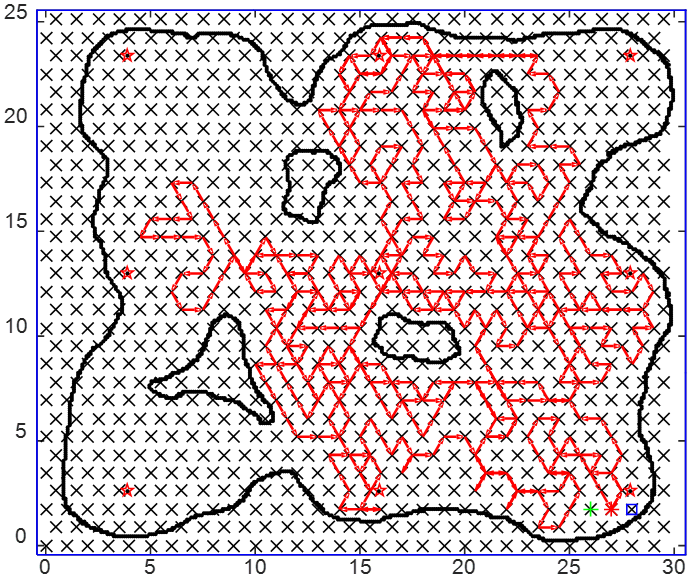}
    \caption[The route of robot 2 in a simulation with 3 robots]{The route of robot 2 in a simulation with 3 robots}
    \label{fig:832d3r2}
\end{figure}

\begin{figure}
    \centering
    \includegraphics[width=0.8\linewidth]{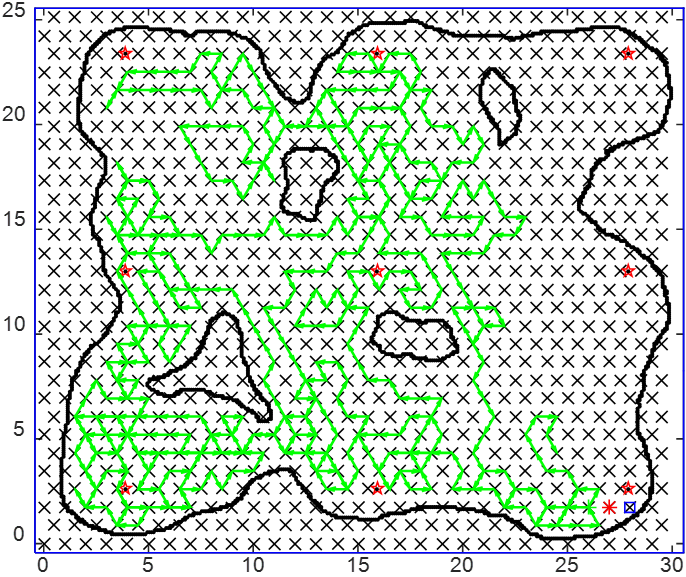}
    \caption[The route of robot 3 in a simulation with 3 robots]{The route of robot 3 in a simulation with 3 robots}
    \label{fig:842d3r3}
\end{figure}

\begin{figure}
    \centering
    \includegraphics[width=1\linewidth]{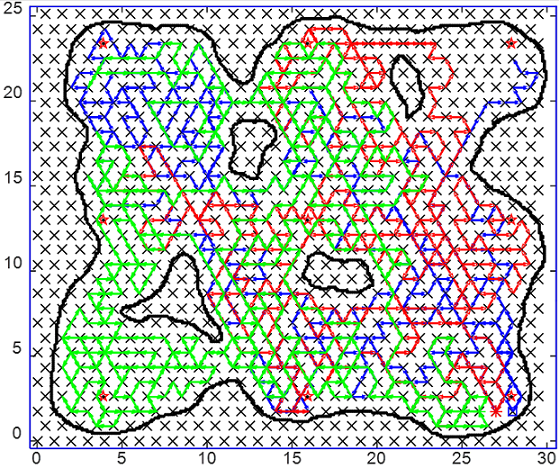}
    \caption[The combined routes in a simulation with 3 robots]{The combined routes in a simulation with 3 robots}
    \label{fig:852d3r123}
\end{figure}

In situation I, an example of the routes of three robots in searching 9 targets is shown in Figure \ref{fig:822d3r1}-\ref{fig:852d3r123}. There are 30*30 vertices in total in the graph. The three robots are at the bottom right corner which is the only known entrance to allocate the robots, and the nine targets are labeled as red stars. The boundaries and obstacles are shown by the bold black lines of discrete dots which obey the Assumption \ref{a:2wpass3}. Parameters used in the simulation are from Table \ref{t5} based on the parameters of the Pioneer 3-DX robot. The figures demonstrate that each target is within $r_{st}$ of a visited vertex, so the search task is finished successfully. Based on the recorded trace (Figure \ref{fig:822d3r1}-\ref{fig:842d3r3}), robots have no collisions with each other so it is a collision-free algorithm. Also, the routes in Figure \ref{fig:852d3r123} have less repetition compared to the Figure \ref{fig:492d2r4} in Chapter 5 and Figure \ref{fig:39route123} in Chapter 4. The route of each robot occupied a larger percentage of the area than before which is the benefit from the breadth-first search algorithm.
 

\subsection{Comparison with Other Algorithms}
In this section, the third search algorithm is compared with the previous two search algorithms. It is also compared to the R and RU algorithms in \cite{RN101} which used the grid pattern but did not consider collision avoidance clearly.  Algorithms in \cite{RN102} including RW, LF, and LFP are also compared because they are also random algorithms in unknown areas with collision avoidance rules. The RUN algorithm in \cite{RN101} is not compared as discussed in the previous chapter. Algorithm LF and LFP use L\'evy distribution to generate step length and uniform distribution to generate the moving direction. The way to generate L\'evy distribution can be seen in Section \ref{s102Dcompare} of Chapter 5

As there is no collision avoidance scheme in R, RU, RW, and LF for collision between robots. In the simulation, the total number of collisions is counted, and one more loop is added to the total number of loops for each collision. In reality, one step may not be enough considering one vertex may be chosen by robots from all sensing neighbors. So this is a relatively reasonable way to calculate the total search time. 

\begin{figure}
    \centering
    \includegraphics[width=1\linewidth]{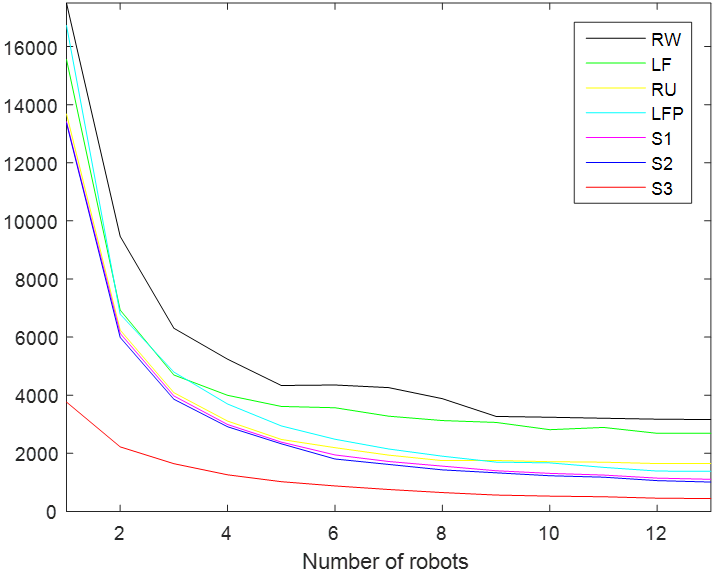}
    \caption[All algorithms except R for 1-13 robots]{All algorithms except R for 1-13 robots}
    \label{fig:872d3all}
\end{figure}

\begin{figure}
    \centering
    \includegraphics[width=0.8\linewidth]{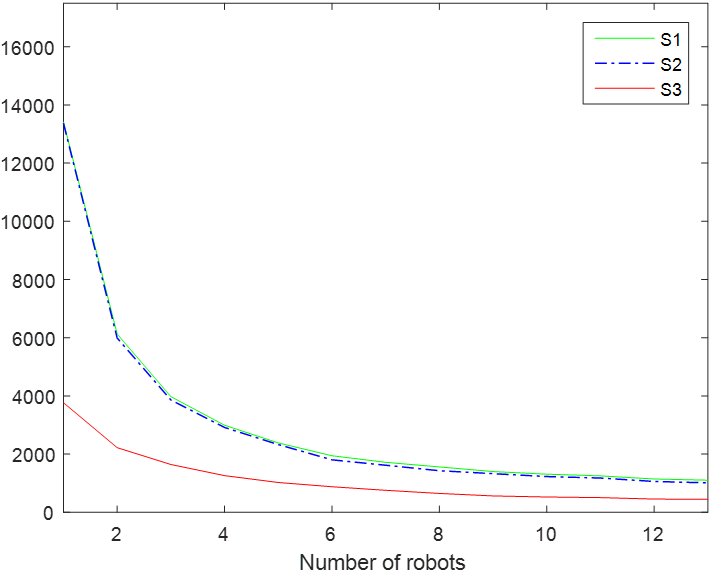}
    \caption[Time for 3 search algorithms]{Time for 3 search algorithms}
    \label{fig:862d3123}
\end{figure}

To compare all methods in the same environment, all the eight algorithms are simulated in the 900 vertices area with nine targets. The results in the figures are the average result of at least 100 simulations. Different communication methods are used in \cite{RN101} and \cite{RN102}, but all the factors need to be the same. So equation \ref{e192DstopI3} is used as the stop judgment, the ad-hoc mode with $r_{c}$ is used for communication and the broadcasting mode is chosen for sending positions of found targets in all the algorithms. \cite{RN102} claimed that LF and LFP algorithm would perform better when targets are sparsely and randomly distributed. Therefore, the nine evenly separated targets are used. Thus, the proposed algorithm is compared to the optimal performance of these two algorithms, and the effectiveness of the proposed algorithm will be more convincible. Two slowest algorithm R and RW were picked up to another graph in Figure \ref{fig:61comparetar9rrw} in Chapter 5. So here we draw the other seven algorithms in Figure \ref{fig:872d3all} to see the effect of the proposed algorithm and compare the three search algorithm in Figure \ref{fig:862d3123} to show the small difference. 

The Figure \ref{fig:872d3all} and \ref{fig:61comparetar9rrw} illustrate that the proposed algorithm \ref{e182Dchoose3} in situation I with judgment rule \ref{e192DstopI3} is always the best under the strict requirement. If the algorithms in Chapter 4, 5 and 6 are labeled as S1, S2, and S3. The general ranking of search time of the eight algorithms is S3$<$S2$<$S1$<$
LFP$<$RU$<$LF$<$RW$<$R. Generally speaking, as the number of robots increased, the search time in all algorithms decreased drastically at the beginning but slowly later as more time was spent on other things instead of exploring new vertices only. Specifically, in Algorithm S2 and S1, robots need to spend time on visiting repeated vertices. For the RU and RW algorithms, robots are set to deal with extra steps caused by collision avoidance. Algorithm S3 requires robots to spend time on finding the route to leave the area of repeated vertices. In the LF, the RW, and the LFP algorithm, robots will meet each other more frequently which will break the generated route and create a new move to avoid collisions.

\begin{figure}
    \centering
    \includegraphics[width=0.7\linewidth]{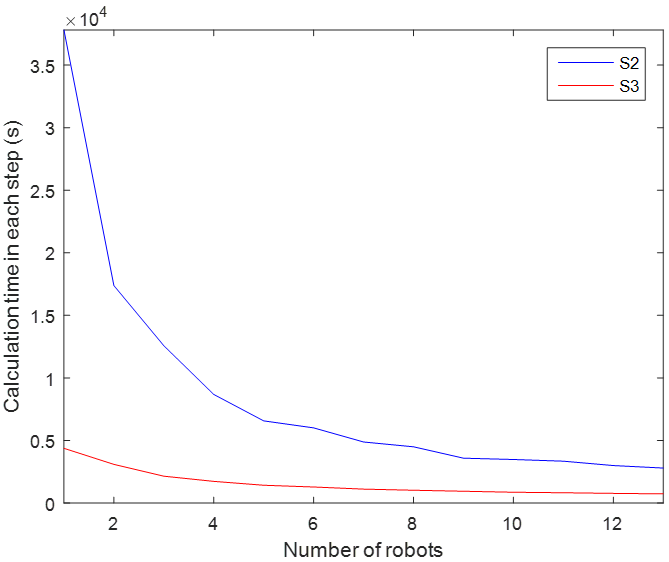}
    \caption[Search time for search algorithm 2 and 3]{Search time for search algorithm 2 and 3}
    \label{fig:89s2s3situii}
\end{figure}

Figure \ref{fig:862d3123} demonstrate that the second search algorithm only improved the efficiency in a very limited range because robots would not meet each other frequently especially in a larger area, so the potential field algorithm did not have enough chance to show its effect. However, the third proposed algorithm largely decreased the search time to less than half of the second search algorithm due to the breadth-first search algorithm in reducing repeated visitation. So the proposed algorithm is effective in Situation I. In Situation II, only the second search algorithm is compared with the third search algorithm in Figure \ref{fig:89s2s3situii} as Algorithm S2 is the best in the seven algorithms to compare. It is clear that Algorithm S3 decreased the search time significantly and it is the best search algorithm. 

In the three search algorithms, the first one is the simplest with calculation like set operations and generating random numbers. The robot only needs to keep a map of visited area, so this method is suitable for robots with limited calculation speed and memory. However, it will spend the longest time. The second algorithm will add distance calculation on the first one. Nevertheless, it only adds some calculation for the Euclidean distance when using the repulsive force. The repulsive force only works when robots are close to each other, so the improvement is small. The third algorithm needs a considerable amount of calculation especially when the known map is huge. However, a time-consuming calculation will not affect the normal calculation of other robots so it will not change the waiting time of Algorithm S1. However, a relatively larger waiting time will allow more robots to finish path planning and move earlier. This algorithm outweighs all other algorithms and needs to be used on robots with high computation ability and large memory.
\subsection{Section Summary}
This section presented a decentralized random search algorithm with the breadth-first search algorithm in unknown 2D areas. The movements of robots are synchronized and are based on a grid pattern. When robots have unvisited sensing neighbors to visit, they choose one from them randomly like the first search algorithm. However, if the surrounding vertices are all searched, the robot will use the breadth-first search algorithm to find the shortest route to the nearest unvisited area using its explored map of the area. It is proved that the algorithm converges with probability 1, and the proposed algorithm was compared with seven other related algorithms in simulations. The result showed that the proposed algorithm is always the best and it can decrease the search time greatly. 

\section{3D Tasks}
This section presents a random search algorithm in 3D with the breadth-first search algorithm. The control algorithm is decentralized only based on local information and robots search the area with obstacles via the cubic lattice. The algorithm is the same as the one in 2D. It considers the collision avoidance as well as real physical constraints and conditions of all parameters. Therefore, it is a practical algorithm for real robots. Robots choose vertices in sequence and move between vertices along edges of the lattice based on local information. The convergence of the algorithm is proved, and the effectiveness is shown in the comparison with other fours algorithms in both situation I and II.

\subsection{Problem Statement}
The problem is the same as that in Chapter 4. So the definitions and assumptions below are claimed without detailed description. 

In the search task, a three-dimensional area $A\subset\mathbb{R}^{3}$ with a limited number of obstacles $ O_{1} $ to $O_{l} $ is used. $m$ robots labeled as $rob_{1}$ to $rob_{m}$ are used to search $n$ static targets $t_{1}$ to $t_{n}$ through a cubic grid with a side length $a$. $p_{i}(k)$ is the coordinate of $rob_{i}$ at time $k$. Let $T$ be the set of coordinates of targets and $T_{ki}$ be the set of coordinates of targets which are known by $rob_{i}$. The initial setting is in Figure \ref{fig:413dinitialsetting}.
\begin{assumption}
    Area $A$ is a bounded, connected and Lebesgue measurable set. The obstacles $O_{i}\subset\mathbb{R}^{3}$ are non-overlapping, closed, bounded and linearly connected sets for any $i>0$. They are both static and unknown to robots. \label{a:3D3area}
\end{assumption}
\begin{definition}
    Let $O:=\bigcup{O_{i}}$ for all $i>0$. Then $A_{d}:=A\setminus{O}$ is the area that needs to be detected.\label{d:3D3obs}
\end{definition}
\begin{assumption}
    Initially, all robots have the same grid with the same coordinate system and are allocated at different vertices along the border of the search area. In the simulation, robots are manually set at some vertices of the C grid.\label{a:3D3intialposition}
\end{assumption}
\begin{assumption}
    All sensors and communication equipment that robots carry have spherical ranges. Other radiuses mentioned are also spherical. \label{a:3D3ranges}
\end{assumption}
\begin{definition}
    Robots have a swing radius $r_{rob}$ and a total error $e$. Then a safety radius to avoid collision is $ r_{safe}\geq{r_{rob}+e} $ (see Figure \ref{fig:16range3d}).\label{d:3D3rsafe}
\end{definition}
\begin{assumption}
    $W_{pass}\geq{ a\sqrt{3}+2r_{safe}}$ (see Figure \ref{fig:18wpass3D}).\label{a:3D3wpass}
\end{assumption}
\begin{assumption}
    $r_{so}\geq{a+r_{safe}}$ (see Figure \ref{fig:16range3d}).\label{a:3D3rso}
\end{assumption}
\begin{assumption}
    $ r_{c}\geq{2*a+e}$ (see Figure \ref{fig:16range3d}).\label{a:3D3rc}
\end{assumption}
\begin{assumption}
    $r_{st}\geq{2a/\sqrt{3}+r_{safe}-r_{rob}}$ based on Table \ref{t3}.
\end{assumption}
\begin{assumption}
    The task area is vast. So, $a$ is much smaller than the length, the width and the height of area $A$. Thus, there are no targets between the boundary and the frontiers of the target sensing range which means the boundary effect can be ignored. \label{a:3D3boundaryeffect}
\end{assumption}
$V_{a}$ is the set of all the accessible vertices. $N_{s,i}(k)$ represents the set of sensing neighbors around $rob_{i}$. In $N_{s,i}(k)$, $V_{v,i}(k)$ and $V_{u,i}(k)$ denote the two sets for visited vertices and unvisited vertex and the choices from these two sets are $c_{v}$ and $c_{u}$ respectively.

For the breadth-first search algorithm, the following definitions are used for generating the shortest route to the nearest unvisited vertex. 
\begin{definition}
    The route is represented as a set of vertices $V_{r,i}$ for $rob_{i}$. Then $\lvert V_{r,i}(k)\rvert $ is the total number of steps in that route. Specifically, if the route has not been fully generated, $\lvert V_{r,i}(k)\rvert=0$. Let $v_{r,i}(j)$ be the $j$th step in the route and $c_{r,i}$ be the choice from the route $V_{r,i}(k)$. The number of steps left in the route is then $\lvert V_{r,i}(k)\rvert-j$. \label{d3D3vri}
\end{definition}
\subsection{Procedure and Algorithm}    
The general procedure of the algorithm is the same as the 2D algorithm. The difference is on the dimension of parameters and the way to decide the sequence of choice which is in Section \ref{s83Dalgo}. So the same flow charts \ref{fig:342dflowsitu1} and \ref{fig:352dflowsitu2} are available for situation I and II. Compared to the first search algorithm, this algorithm helps robots to escape from a visited area by looking for the route with minimum steps to the nearest unvisited vertices. The way to use the breadth-first search algorithm can be seen in Section \ref{s102D3}. The random algorithm with the breadth-first search algorithm can be summaries as

\begin{equation}
p_{i}(k+1)=
\begin{cases}
c_{u} \text{ with prob. } 1/\lvert V_{u,i}(k)\rvert, \text{ if } \lvert V_{u,i}(k)\rvert\neq{0},\\
c_{r,i}, \text{ if } (\lvert V_{u,i}(k)\rvert=0)\&((\lvert V_{v,i}(k)\rvert\neq0)\&(\lvert V_{r,i}(k)\rvert\neq{0})),\\
p_{i}(k),  \text{ if } (\lvert N_{s,i}(k) \rvert=0)\mid(\lvert V_{r,i}(k)\rvert={0})
\end{cases}
\label{e183Dchoose3}
\end{equation}
The stop strategy for this choosing rule is the same as that in Chapter 4.
\begin{equation}
\text{In situation I: }p_{i}(k+1)=p_{i}(k)  \text{, if } \lvert T_{ki}\rvert=\lvert T\rvert.
\label{e193DstopI3}
\end{equation}
\begin{equation}
\text{In situation II: }p_{i}(k+1)=p_{i}(k)  \text{, if } \forall v_{i}\in V_{a}, \text{ } \lvert V_{u,i}(k)\rvert=0.
\label{e193DstopII3}
\end{equation}
\begin{theorem}
    Suppose that all above assumptions hold and the search algorithms \ref{e182Dchoose3} and a judgment strategy \ref{e193DstopI3} or \ref{e193DstopII3} are used for situation I or situation II. Then, for any number of robots, there is such a time $k_{0}>0$ that all targets are detected with probability 1.
    \label{th3D3}
\end{theorem}
\begin{prove}
    The algorithm \ref{e183Dchoose3} with a stop strategy forms an absorbing Markov chain. It includes both transient states and absorbing states. The transient states are all the accessible vertices of the grid that robots visit but do not stop. Absorbing states are the vertices that the robots stop. Applying algorithms \ref{e183Dchoose3}, robots select the unvisited neighbor vertices to go. If all sensing neighbors are visited before finding the targets, robots will look for the shortest route to the nearest unvisited vertex and only stop when all targets are found, or all vertices are detected. For \ref{e193DstopI3}, absorbing states will be reached until all robots know $\lvert T_{ki}\rvert=\lvert T\rvert$. For \ref{e193DstopII3}, absorbing states are reached when $\lvert V_{u,i}(k)\rvert=0$ for any vertex. The assumption about $W_{pass}$ guarantees that all the accessible vertices can be visited so absorbing states can be achieved from any initial states, namely with probability 1. This completes the proof of Theorem \ref{th3D3}
\end{prove}
\subsection{Simulation Results}
\begin{figure}
    \centering
    \includegraphics[width=0.75\linewidth]{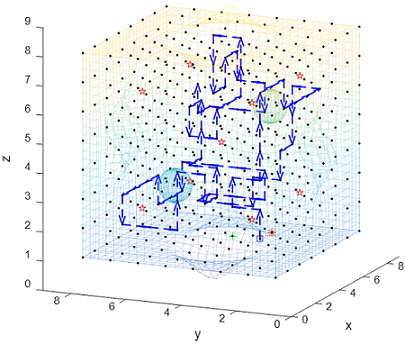}
    \caption[The route of robot 1 in a simulation with 3 robots]{The route of robot 1 in a simulation with 3 robots}
    \label{fig:943d3r1}
\end{figure}

\begin{figure}
    \centering
    \includegraphics[width=0.75\linewidth]{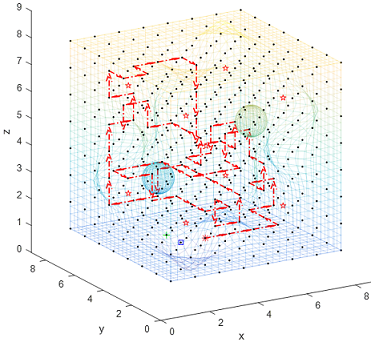}
    \caption[The route of robot 2 in a simulation with 3 robots]{The route of robot 2 in a simulation with 3 robots}
    \label{fig:953d3r2}
\end{figure}

\begin{figure}
    \centering
    \includegraphics[width=0.75\linewidth]{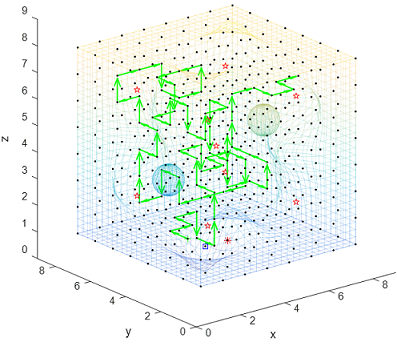}
    \caption[The route of robot 3 in a simulation with 3 robots]{The route of robot 3 in a simulation with 3 robots}
    \label{fig:963d3r3}
\end{figure}

\begin{figure}
    \centering
    \includegraphics[width=1\linewidth]{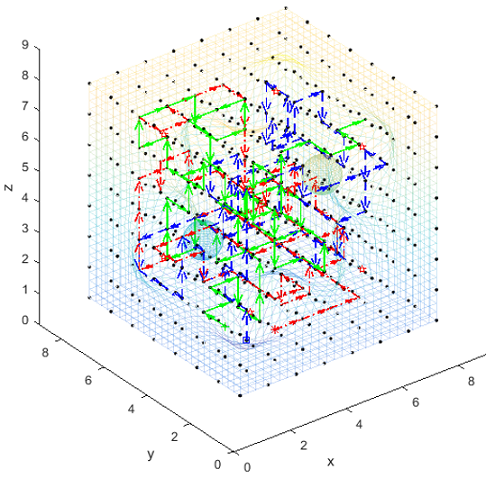}
    \caption[The combined routes in a simulation with 3 robots]{The combined routes in a simulation with 3 robots}
    \label{fig:973d3r123}
\end{figure}

A search task for situation I is simulated in MATLAB2016a. According to the assumption \ref{a:3D3boundaryeffect}, targets are allocated within $r_{st}$ from the accessible vertices as in Figure \ref{fig:413dinitialsetting}. Parameters in Table \ref{t8} are used, so the time in the movement part of the simulation is set as 5.75s in total including 4.5s for the translation and 1.25s for rotation. To check the performance of escaping the searched area from routes, a small area which has 8*8*8 vertices with nine targets is searched by three robots using the control law, Algorithm \ref{e183Dchoose3}\, and the stop strategy, Algorithm \ref{e193DstopI3}, as shown in Figure \ref{fig:943d3r1} to \ref{fig:973d3r123}. All routes use arrows to show the directions of movements, and the route of each robot has a different color. The figures for individual robot show that there was no route with many repeated parts and in the combined routes for the three robots, they tended to search their area without too much overlapping. This effect is more evident if these fours figures are compared with Figure \ref{fig:903d2r1}-\ref{fig:933d2r123} in Chapter 5. By analyzing the route of each robot, no collisions were found, and each target was within $r_{st}$ of a visited vertex. So the search task is finished with less repetition than the second search algorithm.

In situation I, the search algorithm in chapter 4 and 5, namely, Algorithm \ref{e53Dchoose} and \ref{e83Dchoose2a} \& \ref{e83Dchoose2b} will be compared. Other algorithms include the best algorithm in \cite{RN102} which is the LFP algorithm and the RU algorithm in \cite{RN2}. Collisions avoidance is not stated clearly in RU, so it is assumed that it accepts collision and one collision needs one more movement to solve although longer time may be needed if there are consecutive collisions or if multiple robots choose the same vertex from different directions. 

\begin{figure}
    \centering
    \includegraphics[width=0.7\linewidth]{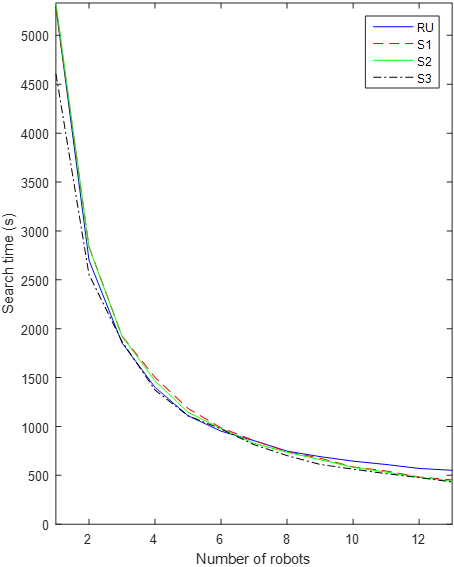}
    \caption[Compare 4 algorithms in situation I]{Compare 4 algorithms in situation I}
    \label{fig:983d3compare4}
\end{figure}

Simulations for searching nine separated targets in a larger area (12*12*12 vertices) by 1 to 13 robots are run for 500 times to get the average search time. Based on the simulation in a 3D area in Chapter 5, the search time for LFP algorithm is always much larger than the others because no map can be used to assist the search. As the proposed algorithm should be faster than other algorithms, LFP is compared with RU only in Table \ref{t153D2RUresult} as a reference. So other three algorithms are compared with the third search algorithm in Figure \ref{fig:983d3compare4} to make the line graph clear. Given by the figure, the proposed algorithm was always the best in all these five algorithms. Although the algorithm TU outperformed the other two search algorithm for 1 to 6 robots because of its simplicity, the wide range of choice and few collisions happened, it still had repeated routes, so it was slower than the proposed algorithm in this chapter. However, comparing to the simulation results in a 2D area, the advantage of the proposed algorithm was not that clear. The possible reason is that, although the 3D area has a larger number of vertices than the 2D area (12*12*12 VS. 30*30). The fewest number of steps needed to go across the body diagonal of the cube is (12-1)*3=33 steps. Nevertheless, for the 2D area using a T grid, the number of steps for going through the diagonal is (30-1)+(30-1)/2=43.5 steps, which is larger than that of the C grid. For the breadth-first search algorithm, it means the total breadth of the 3D grid is smaller, so the effect of going out of a large visited area is less significant.

\begin{figure}
    \centering
    \includegraphics[width=0.7\linewidth]{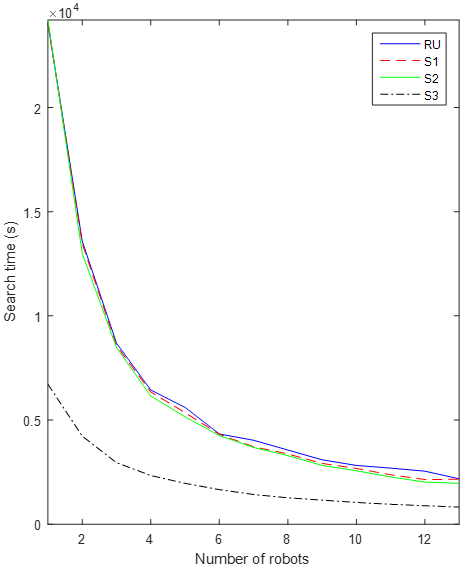}
    \caption[Compare 4 algorithms in situation II]{Compare 4 algorithms in situation II}
    \label{fig:993d3comp4}
\end{figure}

For situation II, robots need to have a map to know when all parts of the area are searched. Therefore, algorithms RW, LF, and LFP which do not need map cannot be used. So algorithm R and RU are compared with the three proposed search algorithms in this report. In the result, the search time for algorithm R is too long, so the two slow algorithms are illustrated in Table \ref{t203D3RRU}. Algorithms except algorithm R are displayed in Figure \ref{fig:993d3comp4}. The table and figure display that the third search algorithm was the best, and its search time was less than half of the second best search algorithm. Unlike situation I, situation II needed more steps. Thus, there were more chances to utilize the breadth-first search algorithm to boost the search speed.


\begin{table}
    \centering
    \renewcommand\arraystretch{2}
    \begin{tabular}{|c|c|c|c|c|c|c|c|}
        \hline 
        No.&    1&    2&    3&    4    &5&    6&    7 \\ 
        \hline 
        R&79447    &47067&    32892&    24373&    19840&    16840&    13227 \\ 
        \hline 
        RU&24036&    13584&    8689&    6445&    5606&    4334&    4032\\ 
        
        \hline
        No.&8&9&10&11&12&13&\\
        \hline
        R&    12459    &11321&    10957&    10748&    8492&    8791&    \\    
        \hline
        RU&3562    &3094&    2825&    2694&    2547&    2182&\\
        \hline
    \end{tabular}  
    \caption[Search time for algorithm R and RU in situation II]{Search time for algorithm R and RU in situation II}
    \label{t203D3RRU}
\end{table}
The general trend of all the algorithms is that the search time had a dramatic drop followed by a gentle descent as the number of robots increased. It can be predicted that if the number of robots increases further, the search time of the proposed algorithm may have a stable period or even a moderate rise because the algorithm time in the `wait' step will increase which may take up a larger proportion of total time. Also, the number of steps will decrease slower. Therefore, when choosing the suitable number of robots in applications, the balance between the merits of decreased search time and the increased expense on robots should be considered. The search time is also affected by unknown factors including the initial positions of robots and the shape and the size of the area. If the size of the area is given, simulations like those in this paper can be used to estimate the proper rough range of the number of robots in a general situation.
\subsection{Section Summary}
This section proposed a grid-based decentralized algorithm for search task in 3D area without collision. The algorithm uses a combination of random choice and the breadth-first search algorithm to decrease the amount of repeated visitation. The convergence of the algorithm is proved, and it is compared to all the related algorithms in both situation I and situation II. The result showed that the proposed algorithm is always the best in different situations.
\section{Summary}
This chapter proposed the third search algorithm for mobile robots which is suitable for both 2D and 3D areas with small differences in the dimension of parameters and the setting of the selection sequence. When a robot has unvisited neighbors to visit, it uses the random algorithm to select an unvisited one like the first and second search algorithms. When the robot is in a visited section, the breadth-first search algorithm is employed to find the nearest unvisited vertex referring to the number of steps. Then it plans a path for the robot to go to the unvisited vertice thus reducing the repetition and search time. Robots move via a grid pattern to select vertices in sequence and avoid collisions in the unknown area. In both 2D and 3D area, the proposed algorithm was compared with all other related algorithms and showed that the proposed algorithm in this chapter is the best in different situations.  
\chapter{Conclusion}
\chaptermark{Conclusion}
This report studied the problems of complete coverage and search by a team of robots with collision avoidance. The task area in these two problems is arbitrary and unknown with obstacles inside. One optimal decentralized complete coverage algorithm and three decentralized random based search algorithms were proposed. All algorithms are suitable for both 2D and 3D areas and utilized a grid pattern to simplify the representation of the area and calculation. In each step, the robot moves between vertices along edges of the grid with limitations on sensing ranges and the communication range. Robots are always within the task area and do not need to go through the boundary. All robots are synchronized so the robot failure can be detected. With the sequence, the grid pattern and communication, collisions are avoided while robots are making choices. Chapter 2 illustrated how to select the grid with the minimum number of vertices to cover an area based on parameters of robots. Tables with selection criteria are provided with examples of real robots. Using the selected grid, multiple mobile robotic sensors can employ the optimal self-deployment algorithm in Chapter 3 to have a complete coverage of a static or expanding area. If the robots are fed into certain points of the area continuously when the vertices are vacant, the algorithm which drives robots to the furthest available vertices will ensure the least number of feeding steps are needed. Then, Chapter 4, 5 and 6 proposed three decentralized random algorithms for search tasks. In the search problem, the number of targets can be known or unknown, and a sequence of selection is given based on the relative positions of robots. The first search algorithm chooses a random unvisited neighbor to go if there is any. Otherwise, it randomly selects a visited vertex. If no accessible neighbors are available, it keeps stable in that step. In the second algorithm, the repulsive force is used on the robots which are close to each other to disperse robots to move to different areas. Thus, there will be less repeated vertices. In the third algorithm, the repetition is further descended by adding the breadth-first search algorithm when a robot is surrounded by visited vertices. The shortest route is planned to the nearest unvisited vertices based on the explored map of the robot.

In Chapter 5, the second search algorithm was further discussed with an example of generating the formula for search time based on the size of the area and the number of robots. For the coverage with the loose requirement, factors related to the search time were discussed based on simulations and analyses. The convergence of each algorithm was proved with probability one, and they are simulated in MATLAB to discover their features and compared to other related algorithms to show the effectiveness. Experiments on Pioneer 3-DX robots were used to test the second search algorithm which verified the algorithm.       
\section{Future Work}
\begin{enumerate}[itemindent=0em]
	\item In the initialization of the algorithms, robots need to be manually set to vertices of grids or robots have a reference point to build its coordinate system and grid. Based on some consensus problems without collision avoidance \cite{RN1,RN2}, future works should consider finding an autonomous collision-free method for multiple robots to form the same coordinate system.
	\item The breadth-first search algorithm in Chapter 6 uses a considerable amount of time, memory and energy. As robots need to be cheap with limited resource, other graph search algorithms should be developed.
	\item The obstacle in this report is static which limits the usage of the proposed algorithms. So another research direction could be developing a collision avoidance search algorithm with moving or deforming obstacles \cite{RN98,RN96,RN166,RN339,RN340,RN95,RN193}. Thus, the algorithm will not be limited to the `wait and choose' scheme.
	\item This report is restricted to the case of static targets. However, when the targets are moving, the proposed algorithms may not be the best. So search problems for moving targets like those in \cite{RN118,RN119} need to be considered.  
	\item The 3D algorithms in this report are not tested on real robots. Therefore, experiments on UUVs or UAVs are required to verify the algorithms.
	\item Formation control of multiple robots is also a related interesting area. A possible topic can be adding collision avoidance to the formation control in 3D areas based on \cite{RN11,RN90,RN311}. 
	\item With formation control algorithms, the sweep coverage or patrolling in a region with changing width could be considered as in \cite{RN5,RN228,RN163}.
	\item The odometry error and other errors should be discussed further. As the simulation and experiment environments were not large enough and the odometry error is related to the distance traveled, the current error did not affect the result seriously. However, in applications in a vast area, that may be a problem. A possible direction may be adding some buoyage for some UUV applications.
	\item In this report, all robots have the same ability. However, in real search and rescue tasks, a robot team consists of robots with different functions. Thus, cooperation between heterogeneous robots will be a realistic research topic \cite{RN170,RN183,RN225,RN115}.     
	\item The communication in the proposed algorithms is ideal, however, in real environments such as a battlefield, there may be interference and limitations to the bandwidth. Thus communication delay or data loss could be considered with estimation of certain information \cite{RN318,RN221,RN222}.
	\item The movement of the robots in the algorithm is based on the ability of the Pioneer 3-DX robot. However, some robots may not have the capability to turn a particular angle while staying at the same position. Therefore, the non-holonomic model in \cite{RN319,RN349,RN135} and the dynamics of vehicles in a 3D area (see, e.g., \cite{RN206}) should be researched.
\end{enumerate}
\bibliographystyle{myplain}
\bibliography{a}
\addcontentsline{toc}{chapter}{Appendix}
\appendix
\chapter{Acronyms}
\label{chapter:chap_Acronyms}


\begin{acronym}[XXXXXXXX]
	\acro{2D}{two-dimensional}	
	\acro{3D}{three-dimensional}
	\acro{BPAUV}{Battlespace Preparation Autonomous Underwater Vehicle}
	\acro{C}{cubic}	
	\acro{DOF}{Degrees of Freedom}	
	\acro{DR}{decenalized random}
    \acro{GPS}{Global Positioning System}
    \acro{H}{regular hexagonal}
    \acro{HP}{hexagonal prismatic}
    \acro{LF}{L\'evy flight}	
    \acro{LFP}{L\'evy flight with potential field}	 
    \acro{R}{random choice}
    \acro{RD}{rhombic dodecahedral}	
    \acro{RU}{random choice with unvisited vertices first}
    \acro{RUN}{random choice with path to the nearest unvisited vertex}
    \acro{RW}{random walk}	
    \acro{S}{square}	
    \acro{SLAM}{Simultaneous Localization And Mapping}	
	\acro{SNR}{Signal to Noise Ratio}
	\acro{T}{equilateral triangular}
	\acro{TCP}{Transmission Control Protocol}	
	\acro{TO}{truncated octahedral}
	\acro{TPR}{True Positive Rate}	
	\acro{UAV}{Unmanned Aerial Vehicle}
	\acro{UDP}{User Datagram Protocol}
	\acro{UGV}{Unmanned Ground Vehicle}
	\acro{UUV}{Unmanned Underwater Vehicle}
	\acro{WMSN}{Wireless Mobile Sensor Network}	
	\acro{WSN}{Wireless Sensor Network}	
\end{acronym}
\end{document}